%% file: LACE_arxiv_version.tex
\documentclass[letterpaper]{article} 
\usepackage{aaai2027}  

\usepackage[hyphens]{url}   
\usepackage{graphicx}  
\usepackage{natbib}    
\usepackage{caption}   
\usepackage{tikz}
\usetikzlibrary{arrows.meta}

\usepackage{cuted}
\usepackage{algorithm}
\usepackage{algorithmic}
\usepackage{amsmath}
\usepackage{booktabs}
\usepackage{listings}
\usepackage{array}
\newcommand{\sysname}{LACE}
\newcommand{\suppmat}{the supplementary material}
\newcommand{\supppseudocode}{Section~A of \suppmat}
\newcommand{\suppevaldetails}{Section~B of \suppmat}
\newcommand{\supptasks}{Section~C of \suppmat}
\newcommand{\suppbaselines}{Section~D of \suppmat}

\newcommand{\suppablations}{Section~F of \suppmat}
\newcommand{\suppscreening}{Section~G of \suppmat}
\newcommand{\suppcomposition}{Section~H of \suppmat}
\newcommand{\suppstats}{Section~I of \suppmat}
\newcommand{\suppsize}{Section~J of \suppmat}
\newcommand{\suppzeroshot}{Section~K of \suppmat}
\newcommand{\suppdynamics}{Section~L of \suppmat}
\newcommand{\suppprompts}{Section~M of \suppmat}
\newcommand{\supptabpfn}{Section~N of \suppmat}
\newcommand{\suppruntime}{Section~O of \suppmat}

\usepackage{placeins}
\usepackage{dblfloatfix}


\title{Evolving Executable Pipeline Programs for AutoML with Language Models}

\author{
    Sofoklis Kitharidis\textsuperscript{1},
    Cor J. Veenman\textsuperscript{1,2},
    Jan N. van Rijn\textsuperscript{1},
    Thomas Bäck\textsuperscript{1},
    Niki van Stein\textsuperscript{1}
}

\affiliations{
    \textsuperscript{1}Leiden Institute of Advanced Computer Science (LIACS),
    Leiden University, The Netherlands\\
    \textsuperscript{2}Netherlands Organization for Applied Scientific Research (TNO), The Netherlands
}

\begin{document}

\maketitle

\begin{abstract}
Automated machine learning (AutoML) systems search for pipelines within a space of preprocessing operators, learners, and hyper-parameters specified in advance: they can select and tune known components, but cannot produce structure outside that space. We present \sysname{}, an AutoML framework that instead searches over complete executable pipeline programs: an evolutionary loop maintains a population of scikit-learn-compatible Python classes, and a large language model acts as the variation operator. To our knowledge, \sysname{} is the first to formulate general tabular pipeline AutoML this way, evaluated on standardized OpenML tasks under a leakage-controlled protocol that withholds dataset identity from the generator. Because every candidate is ordinary Python, the returned pipeline and the search that produced it can be inspected and edited directly, rather than only through a framework's model objects. On 68 OpenML classification tasks, \sysname{} with GPT-5.4-mini significantly outperforms auto-sklearn, H2O, and a fixed XGBoost baseline, with no detectable difference against AutoGluon, the strongest search-based system evaluated, while covering the full benchmark. Newer tabular foundation models are more accurate on the subset of tasks they support, but apply a fixed pretrained predictor rather than returning an editable task-specific program. \sysname{}'s contribution is therefore not raw accuracy but a search space defined by code: complete coverage, pipelines practitioners can reuse directly, and a component set extended by editing the prompt rather than the framework.
\end{abstract}

% % Optional links block. Use only if links do not de-anonymize the submission.
% \begin{links}
%     \link{Code and data}{https://doi.org/10.5281/zenodo.21886859}
%     % \link{Datasets}{https://anonymous-link-to-datasets}
% \end{links}

\section{Introduction}

Automated machine learning (AutoML) automates pipeline construction, but most established systems search within a fixed operator and hyperparameter space. Systems such as AutoGluon-Tabular, H2O AutoML, and auto-sklearn optimize curated sets of preprocessors, learners, and hyper-parameters \citep{erickson2020autogluon,ledell2020h2o,feurer2015autosklearn}, while evolutionary systems such as TPOT vary structures within a fixed operator inventory~\citep{olson2016tpot}. Tabular foundation models instead avoid per-task search by applying a pretrained predictor \citep{hollmann2025tabpfn}. Because preprocessing, model, and hyperparameter choices interact, jointly optimizing them is difficult \citep{baratchi2024automl}. Classical AutoML systems typically return fitted objects accessed through framework APIs, whereas tabular foundation models apply pretrained predictors. These outputs can be inspected, but are not ordinarily returned as self-contained, editable task-specific programs.

Large language models suggest a different search representation. Code-generation systems produce executable Python, while execution feedback separates valid from invalid programs~\citep{chen2021codex,li2022alphacode}. Existing LLM-based ML systems generate features, configure tools, or revise broader experiment workflows \citep{hollmann2023caafe,trirat2025automlagent,nam2025mlestar}, but none, to our knowledge, formulate general tabular AutoML as population-based evolution of constrained pipeline programs. We therefore introduce \sysname{} (\emph{LLM-Automated Code Evolution}), which optimizes LLM-generated scikit-learn-compatible pipeline classes through population-based evolution. Generated code determines preprocessing, learner choice, ensembling, and exposed hyperparameters. Candidates are evaluated on held-out inner-validation splits, and their scores or errors guide subsequent generations. The search space is thus defined by the programs the model can produce under the interface and package constraints, rather than by a fixed operator grammar.

Public AutoML benchmarks may occur in LLM pretraining data, potentially confounding pipeline design with dataset recognition \citep{xu2024contamination}. To reduce this risk, \sysname{} withholds dataset identity and other identifying information from the generator; the complete information boundary is specified in Section~\ref{sec:problem-setting}. Across 68 classification tasks and three LLM backbones, the strongest \sysname{} variant is competitive with AutoGluon, outperforms auto-sklearn and H2O, covers the full benchmark, and returns readable classical pipelines that can be inspected and reused.

\paragraph{Contributions.}
\begin{enumerate}
\item \textbf{Evolutionary AutoML over executable programs.} We introduce, to our knowledge, the first LLM-driven population-based AutoML framework that evolves complete executable pipeline programs.
\item \textbf{Open code instead of a fixed grammar.} We replace a predefined pipeline grammar with generated code that chooses preprocessing, learners, ensembles, and exposed hyperparameters.
\item \textbf{Inspectable and editable task-specific programs.} We audit all final programs and show that \sysname{} exposes task-specific modelling decisions as ordinary code that can be inspected, modified, and reused outside the search framework.
\end{enumerate}

\section{Related Work}
\paragraph{AutoML and pipeline search.}
Classical AutoML often formulates model construction as the combined algorithm selection and hyperparameter optimization problem: choosing among predefined learners and preprocessors while tuning their hyperparameters, as introduced by Auto-WEKA and commonly optimized with engines such as SMAC \citep{thornton2013autoweka,hutter2011smac,lindauer2022smac3}. Auto-sklearn adds meta-learning and ensemble construction \citep{feurer2015autosklearn}; AutoGluon and H2O AutoML use stacked ensembles \citep{erickson2020autogluon,ledell2020h2o}; and FLAML performs cost-frugal search~\citep{wang2021flaml}. Evolutionary systems such as TPOT, GAMA, and RECIPE evolve pipelines by composing predefined preprocessing and learning operators \citep{olson2016tpot,gijsbers2019gama,desa2017recipe}, whereas AutoML-Zero constructs learning algorithms from low-level scalar, vector, and matrix operations~\citep{real2020automlzero}. This provides broad expressivity, but useful learners must be assembled from fine-grained operations rather than selected as complete library components. TabPFN represents a distinct alternative rather than classical AutoML: it applies a pretrained predictor without per-task pipeline search \citep{hollmann2023tabpfn,hollmann2025tabpfn}. Existing approaches therefore either search within predefined component libraries, construct algorithms from low-level operations, or replace per-task search with pretraining.

\paragraph{LLM-based AutoML and ML-engineering agents.}
Recent LLM systems construct ML solutions directly as code. CAAFE generates features and LLM-FE evolves feature-transformation programs, while AutoML-GPT and MLCopilot configure existing tools or retrieve prior solutions \citep{hollmann2023caafe,abhyankar2025llmfe,zhang2023automlgpt,zhang2024mlcopilot}. NNGPT generates complete neural architectures and training specifications through one-shot generation, retrieval, and continual fine-tuning of the generator itself \citep{kochnev2026nngpt}. End-to-end agents iteratively revise code and experiment plans, with some using branching or tree search and others coordinating multiple agents \citep{guo2024dsagent,hong2025datainterpreter,grosnit2024agentk, jiang2025aide,chi2024sela,liang2025imcts,nam2025mlestar, li2024autokaggle,trirat2025automlagent,fang2025mlzero,yang2025rdagent}. Many target open-ended ML-engineering or competition settings in which the agent inspects the data and runs arbitrary experiments \citep{chan2025mlebench,huang2024mlagentbench}. \sysname{} instead evolves one constrained pipeline program from a coarse task summary and forbids nested AutoML, so that measured improvements are attributable to the search rather than to dataset recognition or a hidden inner optimizer. We therefore treat these agents as related approaches rather than baselines: they search a different object (whole scripts or trajectories) under a different information regime (direct dataset access and arbitrary experimentation), so a head-to-head comparison would confound the search method with the information available to it.

\paragraph{LLM-guided evolutionary program search.}
LLM-guided program search generates and mutates executable programs that are scored automatically. FunSearch evolves short functions within a fixed skeleton \citep{romeraparedes2024funsearch}, EoH and ReEvo evolve optimization heuristics \citep{liu2024eoh,ye2024reevo}, and AlphaEvolve extends the pattern to larger codebases \citep{novikov2025alphaevolve}. LLaMEA formalizes the language model as an evolutionary variation operator, with follow-up work adding external numerical tuning and benchmarking infrastructure \citep{vanstein2025llamea,vanstein2025llameahpo,vanstein2025blade}. Most closely, Guided Evolution has an LLM perform mutation and crossover over the code of a seed neural network within a multi-objective loop \citep{morris2024llmguidedevolution}, though it refines one architecture on a single vision dataset. In these settings, the evolved program is itself a solution procedure. \sysname{} instead evolves a learner that is fitted on training data and scored on held-out data, making generalization, assessment leakage, benchmark contamination \citep{bordt2024elephants,ronval2025tabcontam, silvestri2025latentknowledge}, and delegation to nested AutoML central concerns. The constrained interface, coarse task summary, and no-nested-AutoML rule address these differences.

% \paragraph{Summary}
% Search-based and evolutionary AutoML are bounded in advance by a configuration space, an operator inventory, or a primitive instruction set, and tabular foundation models replace per-task search with a single pretrained predictor. LLM agents lift these bounds but optimize entire workflows with direct access to the data. LLM-guided program search evolves code under execution feedback, but its programs are solution procedures scored by execution, not learners scored by generalization. \sysname{} brings these ingredients together: a population-based evolutionary search whose variation operator is an LLM and whose individuals are complete, executable pipeline programs, generated from a coarse task summary and evaluated under a leakage-safe protocol.

\section{Method}
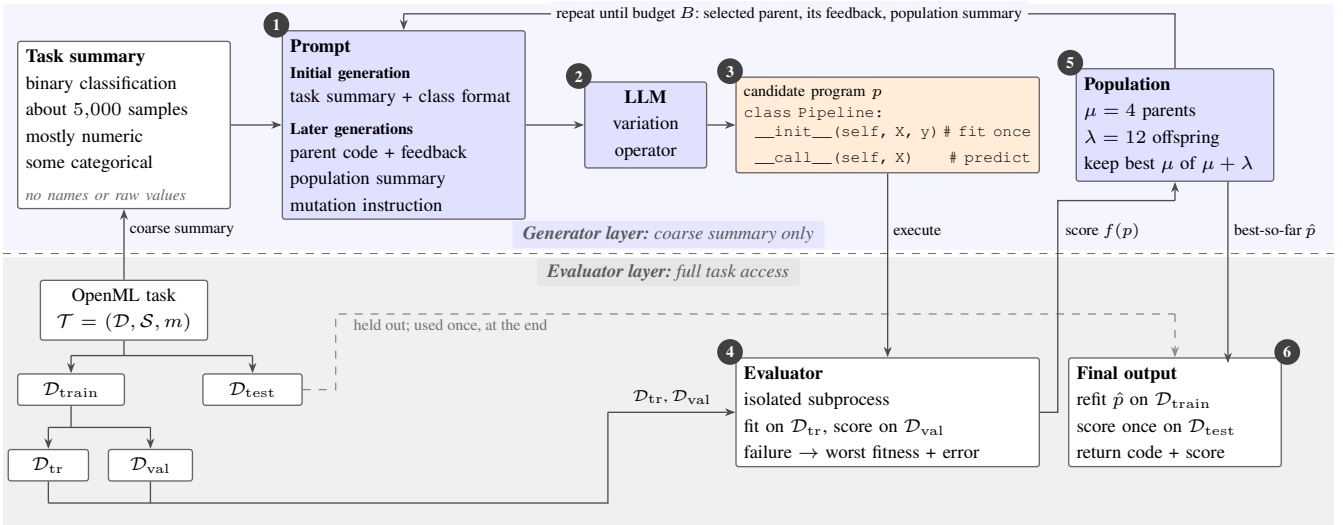
\begin{figure*}[t]
\centering
\resizebox{\textwidth}{!}{%
  \input{Figures/method_figure/lace_overview}
}
\caption{Overview of \sysname{}. The generator receives a coarse,
non-identifying task summary and proposes executable pipelines; the isolated evaluator has full task access and returns inner-validation scores or errors. After the budget is exhausted, the selected pipeline is refit and evaluated once per official test fold.}
\label{fig:lace-overview}
\end{figure*}

Figure~\ref{fig:lace-overview} summarizes the information flow in \sysname{}, from the coarse task description provided to the language model to the leakage-controlled evaluation and evolutionary selection of executable pipeline programs.

\subsection{Problem Setting}
\label{sec:problem-setting}

We consider supervised classification on tabular data. A task $\mathcal{T} = (\mathcal{D}, \mathcal{S}, m)$ consists of a dataset $\mathcal{D}$, pre-defined train/test splits
$\mathcal{S}=\{(\mathcal{D}_{\mathrm{train}}^{(k)},
\mathcal{D}_{\mathrm{test}}^{(k)})\}_{k=1}^{K}$, and an evaluation metric $m$. A \emph{candidate} is an executable program $p$ that implements a fixed interface: instantiating $p$ fits a model on training data, and calling the fitted object returns predictions on held-out data. We write $\mathcal{P}$ for the space of such candidates. Unlike classical AutoML, where $\mathcal{P}$ is an enumerable space of components and hyperparameter ranges specified in advance, here $\mathcal{P}$ is defined only implicitly, as the set of programs the generator can emit that compile and satisfy the interface.

To obtain search feedback without consulting the official test data, we split each pre-defined training partition
$\mathcal{D}_{\mathrm{train}}^{(k)}$ into an inner-training set
$\mathcal{D}_{\mathrm{tr}}^{(k)}$ and an inner-validation set
$\mathcal{D}_{\mathrm{val}}^{(k)}$. A candidate is fit on $\mathcal{D}_{\mathrm{tr}}$ and scored on $\mathcal{D}_{\mathrm{val}}$, and this score is its fitness:
\begin{equation}
f(p) \;=\; \frac{1}{K}\sum_{k=1}^{K}
m\!\left(p;\,
\mathcal{D}_{\mathrm{tr}}^{(k)},
\mathcal{D}_{\mathrm{val}}^{(k)}\right),
\label{eq:fitness}
\end{equation}
with the search maximizing $f(p)$ over $p \in \mathcal{P}$. Search therefore optimises an empirical estimate on a finite validation split and is subject to selection overfitting; Section 5 reports official-test performance, which is never consulted during search.

The protocol enforces a strict separation between search and final assessment. During evolution, candidates are scored only through Eq.~\eqref{eq:fitness} on inner-validation data; the official test partitions $\{\mathcal{D}_{\mathrm{test}}^{(k)}\}_{k=1}^{K}$ are never read. After the search budget is exhausted, the single selected candidate $\hat{p}$ is evaluated under the benchmark's prescribed ten-fold protocol: refit on each fold's training partition and scored once on the held-out test partition, and we report the mean over folds. Throughout, the generator receives only a coarse task summary (task type, approximate sample count, and the number of boolean, integer, and real-valued features), which omits the dataset name, task identifier, feature names, target semantics, class distribution, and evaluation metric, while the evaluator retains full access to the task, its splits, and its metric.

\subsection{Pipeline Representation, Evaluation, and Feedback}
\label{sec:representation}

Candidates follow a fixed fit-on-construction, predict-on-call interface: \texttt{\_\_init\_\_(X, y, ...)} fits exactly once and \texttt{\_\_call\_\_(X)} returns predictions. Alongside the class code, the language model returns a short description and exposes tunable constructor arguments with default values. Figure~\ref{fig:lace-overview} shows the interface in context, and \suppprompts{} gives the complete prompt and example.

The prompt restricts candidates to scikit-learn, NumPy, SciPy, and pandas; a single estimator or an ensemble of at most three learners; a single shared preprocessing stage applied once before all base learners (rather than a separate transformer per learner); and no internal cross-validation, model selection, or hyperparameter search. Tunable values must instead be exposed as constructor arguments. Crucially, these constraints are requested in the prompt and are not enforced by the execution harness; the harness attempts to execute every program the generator returns. We therefore verify compliance after the fact through the static audit of Section~\ref{sec:audit}, which reports how often each constraint was in fact respected.

Before execution, the evaluator applies split-safe ordinal encoding, boolean casting, and missing-value imputation to provide a stable numeric interface; each transform is fit only on the corresponding training portion. Candidates run in isolated subprocesses under a wall-clock timeout. Successful evaluations return inner-validation fitness, while failures receive the worst fitness together with the first captured error and source line (\suppevaldetails{}).

\subsection{Evolutionary optimization}
\label{sec:evo}

\sysname{} uses a \((\mu+\lambda)\) evolutionary loop \citep{back1996evolutionary} in which the language model is the variation operator. The initial population of \(\mu\) pipelines is generated from the task summary and interface alone. Each later offspring is produced by selecting a parent uniformly at random and providing its code, evaluation feedback, a scored population summary, and one randomly chosen mutation instruction (\suppprompts{}).

After evaluating \(\lambda\) offspring, elitist truncation retains the best \(\mu\) members of the combined parent and offspring population. A best-so-far pipeline is tracked until the candidate budget, including invalid or timed-out candidates, is exhausted; it is then refit and evaluated on the official test folds. Formal pseudocode appears in \supppseudocode{}.

\section{Experimental Setup}
\label{sec:experimental-setup}

\subsection{OpenML tasks, protocol, and metrics}
\label{sec:openml-protocol}
We evaluate all 68 classification tasks in the OpenML \texttt{amlb-classification-all} benchmark suite~\citep{gijsbers2024amlb}. The suite spans a wide range of tabular settings, from 100 to 4,898,431 instances and from 4 to 10,000 features, covering small-data regimes through large, high-dimensional tasks. The full task list is given in \supptasks.

All methods are evaluated under the OpenML task protocol. Each task provides pre-defined split instances through its OpenML definition (here, one repeat of ten folds), and we retrieve the pre-defined train/test indices directly rather than constructing our own outer folds. For \sysname{}, as described in Section~\ref{sec:problem-setting}, during search each pre-defined training fold is further split into inner-training and inner-validation portions (an inner-validation fraction of $0.2$), candidate selection uses only the inner-validation scores aggregated across the ten folds, and each held-out test fold is scored once per repetition, after candidate selection, for the final selected pipeline. The evaluation metric is task-specific: classification accuracy for 67 of the 68 tasks and ROC--AUC for the remaining task (\suppbaselines{}). All methods are evaluated using the same metric and pre-defined partitions for a given task. The OpenML definition provides a single repeat of ten folds, which fixes the fold structure within one run. On top of this, we perform five independent experimental repetitions per task, with seeds \(0,\dots,4\), to capture method stochasticity and inner-split variation; each repetition uses the same ten pre-defined folds but a different seed. Reported results summarize these five repetitions unless stated otherwise.

Each run uses a budget of \(B=100\) generated candidates, counting invalid and timed-out ones; with \(\mu=4\) and \(\lambda=12\), this comprises four initial candidates followed by eight generations of twelve. Per-candidate timeouts are four hours for heavy tasks and one hour otherwise. A task is heavy if it has at least 400{,}000 samples, at least 1{,}000 features, or an instance--feature product of at least 10{,}000{,}000; 20 of 68 tasks qualify. The longer limit for heavy tasks reflects the known scaling of fit-and-evaluate time and failure rates with dataset size and dimensionality \citep{guyon2019automlchallenge,bischl2021benchmarkingsuites}: without it, large tasks would accrue failed or timed-out evaluations for reasons unrelated to pipeline quality. The tier changes only the timeout, not the candidate budget, so every task receives the same number of generated candidates.

For cross-task aggregation, each method's per-task score is converted to a normalized deviation by linearly rescaling the valid scores within each task so that the best score maps to \(0\) and the worst to \(1\); figures report the complementary closeness, \(100\,(1-\text{deviation})\).

\subsection{Baselines}
\label{sec:baselines}
We compare \sysname{} with three mature AutoML systems, AutoGluon-Tabular~\citep{erickson2020autogluon}, auto-sklearn~\citep{feurer2015autosklearn}, and H2O AutoML~\citep{ledell2020h2o}; three pretrained tabular models, TabPFN~v2.5, TabPFN~v3~\citep{hollmann2025tabpfn}, and TabICL~\citep{tabicl}; and a fixed XGBoost pipeline~\citep{chen2016xgboost}. The AutoML systems represent classical search-based pipeline construction, the pretrained models provide strong but architecturally distinct references, and XGBoost provides an untuned non-AutoML baseline.

To isolate differences in pipeline search rather than access to neural architectures, \sysname{} and the three search-based AutoML systems are restricted to classical, non-neural tabular learners, matching the component families available in the \sysname{} execution environment. TabPFN and TabICL are evaluated separately as pretrained neural predictors, representing a distinct pretraining-based approach rather than components of the shared search space. To control for runtime, every baseline was subject to an external wall-clock limit equal to the time consumed by the corresponding \sysname{} task repetition. All methods use the same pre-defined OpenML splits and task-specific metric as \sysname{} (Section~\ref{sec:openml-protocol}). Exact versions, model families, budget allocation, and unsupported-task handling are reported in \suppbaselines{}.

% \paragraph{Why not ML-engineering agents?}
% We do not compare against open-ended ML-engineering agents such as AIDE, SELA, or AutoML-Agent. These systems search a different object (whole scripts or experiment trajectories, often with access to dataset identity and arbitrary internal tuning), whereas \sysname{} optimizes a single constrained pipeline class under a fixed interface, a coarse non-identifying task summary, and no nested AutoML. Evaluating them faithfully would mean relaxing exactly the constraints that define our controlled setting; imposing our constraints on them would handicap systems built for a broader one. Our baselines are instead systems that run on the same OpenML tasks under the same protocol.

\subsection{Implementation details}
\label{sec:implementation}

\sysname{} is implemented in Python using scikit-learn 1.8.0 for pipeline execution and OpenML 0.15.1 for task and split retrieval. Experiments were run using a modular open-source benchmarking toolbox for LLM-driven algorithm design \citep{vanstein2025blade}. These components, the analysis scripts, and the run-level result tables are provided in the accompanying Zenodo artifact. Each pipeline exposes its tunable values as constructor arguments with default settings (a median of 12 per pipeline); all reported scores use these generator-chosen defaults. The search uses $\mu=4$ parents, $\lambda=12$ offspring, a budget of 100 generated candidates, elitist truncation, uniform-random parent selection, and crossover disabled, at a sampling temperature of $0.7$. These settings were selected on a three-task development subset; the ablations behind them are reported in \suppablations.

Three language models serve as the variation operator of Figure~\ref{fig:lace-overview} under the same configuration: two through provider APIs, \texttt{gpt-5.4-mini} and \texttt{deepseek-v4-flash}, and one served locally, \texttt{qwen3-coder-next} (a 4-bit quantized build). All runs use a shared, heterogeneous compute cluster, so the time-matched budgets (Section~\ref{sec:baselines}) are measured runtimes rather than hardware-normalized figures; we therefore treat it as a comparison under common infrastructure, not a strict compute-parity benchmark. 

\section{Main Results}
\label{sec:results}
\begin{figure*}[t]
\centering
\includegraphics[width=0.935\textwidth]%
{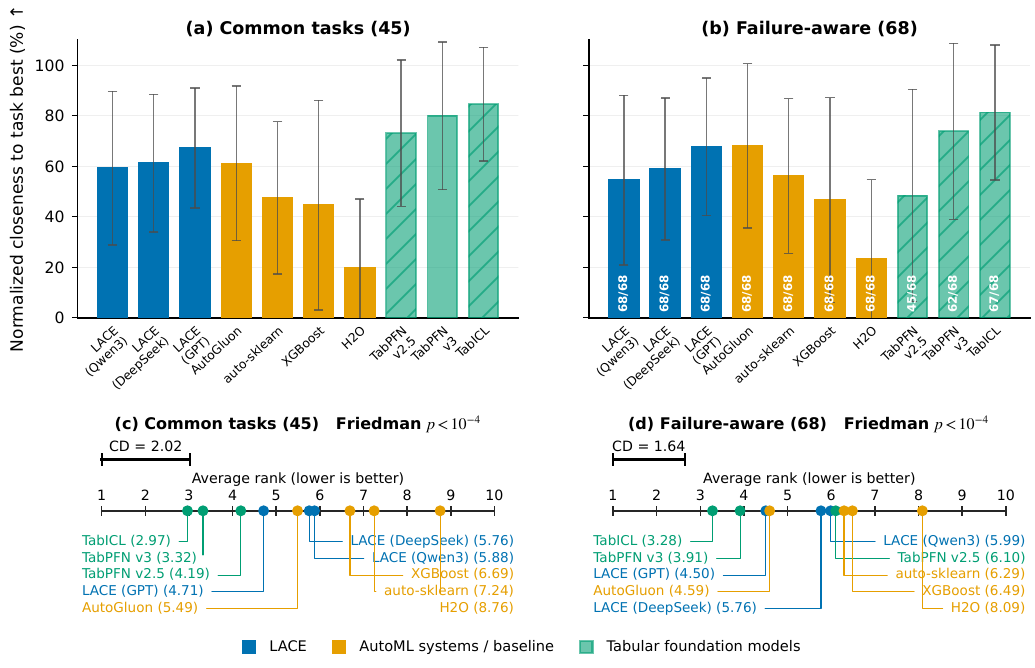}
\captionsetup{font=small}
\caption{Benchmark comparison, coloured by method family. Top: normalized closeness to the task-best score (higher is better; mean and one standard deviation) on (a)~45 common tasks and (b)~all 68 with failure-aware penalties; bar labels show coverage. Bottom: Demšar-style mean task-level ranks (lower is better), with the Nemenyi critical difference at \(\alpha=0.05\), on (c)~45 common tasks and (d)~all 68. Pairwise tests are in \suppstats{}.}
\label{fig:rank-deviation}
\end{figure*}

We report two complementary views. The primary comparison uses the 45 tasks for which every method returns a valid score, separating predictive performance from benchmark coverage; a second, failure-aware view uses all 68 tasks. Both summarize each method by its mean task-level rank and its normalized closeness to the task-best score, computed within each task before aggregation (Figure~\ref{fig:rank-deviation}).

\subsection{Aggregate Performance on Common Tasks}
\label{sec:results-common}
Figure~\ref{fig:rank-deviation}a compares methods on the 45 tasks completed by all systems. TabICL and TabPFN~v3 rank first and second, followed by TabPFN~v2.5 and \sysname{} with GPT-5.4-mini, which obtains a mean rank of \(4.71\) and normalized deviation of \(32.8\%\), ahead of AutoGluon, XGBoost, auto-sklearn, and H2O. DeepSeek and Qwen3-Coder-Next are weaker (mean ranks \(5.76\) and \(5.88\)). Large standard deviations indicate substantial task-level heterogeneity, so the ordering is not uniform dominance.

Paired Wilcoxon signed-rank tests at \(\alpha=0.05\), applied on the tasks shared by each pair after averaging repetitions and corrected with Holm's procedure, sharpen this picture. LACE (GPT-5.4-mini) significantly outperforms auto-sklearn, XGBoost, and H2O (\(p_{\mathrm{Holm}}<0.01\) in each case). The comparison with AutoGluon is not significant: LACE(GPT-5.4-mini) wins 33 and
loses 35 of the 68 shared tasks, with a median difference of \(-0.001\) points and \(p_{\mathrm{Holm}}=0.66\). The comparison with TabPFN~v2.5 is likewise not significant (\(p_{\mathrm{Holm}}=0.199\)). TabPFN~v3 and TabICL significantly outperform LACE (GPT-5.4-mini) on their shared tasks, although the median margins are small (\(0.34\) and \(0.26\) percentage points in the corresponding task metric); against TabICL the mean difference nevertheless favors LACE (GPT-5.4-mini) (\(+0.22\) points), reflecting a small number of tasks with large margins in its favor. The non-significant comparisons do not establish equivalence. Full effect sizes and win--tie--loss counts are reported in \suppstats.

\subsection{Coverage-Aware Performance}
\label{sec:results-coverage}
The common-task comparison excludes 23 tasks because at least one method returns no valid result. All \sysname{} variants, AutoGluon, auto-sklearn, XGBoost, and H2O cover all 68 tasks, whereas TabPFN~v2.5, TabPFN~v3, and TabICL cover 45, 62, and 67 tasks, respectively. Figure~\ref{fig:rank-deviation} summarizes both comparisons; full per-method rank, deviation, and fractional-win values are reported in \suppstats{}.

We therefore conduct a failure-aware sensitivity analysis in which a missing method receives the tied average of the remaining worst ranks and normalized deviation \(1\) (Figure~\ref{fig:rank-deviation}b). The top of the ordering is unchanged, with TabICL (\(3.28\), penalized on a single task) ahead of TabPFN~v3 (\(3.91\)), while coverage reshapes the middle of the field: LACE (GPT-5.4-mini) improves from \(4.71\) to \(4.50\) and AutoGluon from \(5.49\) to \(4.59\), leaving the two essentially tied, and TabPFN~v2.5, missing 23 tasks, falls from \(4.19\) to \(6.10\), the strongest coverage effect in the comparison; \suppsize{} decomposes this effect by task size.

No single method dominates, and a consistent distinction between method families emerges. TabPFN~v3 and TabICL provide the strongest predictive performance on the tasks they support, but do not cover the complete benchmark. Classical AutoML systems cover it in full with generally weaker aggregate performance. On the full benchmark, \sysname{} with GPT-5.4-mini combines complete coverage with no significant difference from AutoGluon, significantly stronger than auto-sklearn, H2O, and the fixed XGBoost baseline, and behind only the newest pretrained models, and then only on the tasks they cover.

\subsection{Effect of the LLM Backbone}
\label{sec:results-backbones}

The choice of language-model backbone affects the quality of the resulting pipelines (Figure~\ref{fig:rank-deviation}). GPT-5.4-mini is the strongest \sysname{} configuration, with a common-task mean rank of \(4.71\), compared with \(5.76\) for DeepSeek and \(5.88\) for Qwen3-Coder-Next, and paired Wilcoxon tests favor it over both alternatives after Holm correction (\suppstats). All three variants complete the full set of 68 tasks, and each ranks ahead of auto-sklearn, XGBoost, and H2O on the common-task comparison, while only LACE (GPT-5.4-mini) ranks ahead of AutoGluon. The framework is thus not tied to a single model, though the backbone matters: LACE (GPT-5.4-mini) leads both alternatives by small but consistent median differences (\(0.06\) and \(0.18\) points).

\section{Analysis of Generated Pipelines}
\label{sec:analysis}
The benchmark comparison establishes where \sysname{} ranks; this section examines how evolutionary feedback changes generated programs and what the open representation makes visible.

\subsection{Evolutionary Search Improves Zero-Shot Pipelines}
\label{sec:zero-shot-improvement}

The first candidate of each run is generated before any execution feedback and therefore provides a zero-shot reference. Search improves on it in two ways. It restores executability: first-shot validity ranges from \(37.6\%\) for LACE(Qwen3) through \(66.5\%\) for LACE(DeepSeek) to \(79.7\%\) for LACE(GPT-5.4-mini), whereas every final pipeline is valid, so each run beginning from an invalid program recovered within the budget. It also improves programs that already ran, and this improvement holds on held-out data: after refitting each valid first program on the official training folds and evaluating it on the corresponding test folds under the same protocol as the returned pipeline, the returned pipeline is better on 190 of 193 backbone--task units, with a median gain of \(2.99\) percentage points in the corresponding task (bootstrap 95\% CI \([2.42, 4.17]\); Wilcoxon \(p<10^{-32}\)); under LACE(GPT-5.4-mini) all 66 usable units improve, with a median gain of \(4.53\) points. The comparison conditions on a valid first program and assigns no score to runs whose first program failed. On the inner-validation objective the returned pipeline also exceeds the best valid member of the four-candidate initial population by \(1.58\) points, so the gain is not a best-of-four effect. Full distributions and validity accounting are reported in \suppzeroshot{}.

\subsection{Pipeline composition and structural diversity}
\label{sec:pipeline-composition}

The generated programs did not collapse onto a small set of templates. Among the \(1{,}020\) final pipelines, \(99.9\%\) added preprocessing beyond the harness and \(73.3\%\) combined at least two learners. Gradient boosting appeared in \(69.4\%\) of finals, forests or extra trees in \(52.5\%\), and linear models in \(55.8\%\). We characterize structure by the ordered preprocessing steps, learner families, ensemble type, and exposed hyperparameter names, ignoring numerical values. Across the \(70{,}998\) valid candidates, this yields \(65{,}298\) (92\%) unique structural signatures, with a median of 68 distinct structures within a run. Part of this uniqueness reflects hyperparameter naming, but the coarser family statistics show the same pattern.

Every returned pipeline is an executable Python program whose preprocessing, learners, and hyperparameter values can be inspected and modified. Beyond this general editability, 89 of the \(1{,}020\) finals (\(8.7\%\), spanning 53 tasks) implemented prediction-combination logic directly in NumPy or ordinary Python rather than relying solely on a library ensemble wrapper. In one illustrative example selected for clarity, from LACE(GPT-5.4-mini) on task 2073, the final program combines fitted SVC and random-forest class probabilities with explicit weights \(0.60\) and \(0.40\). Because that choice is ordinary Python, a practitioner can change either coefficient, replace a member, or reuse the block outside the search framework. Inspecting or modifying a fitted AutoML ensemble typically requires interacting with the originating framework's model objects, while a pretrained predictor uses fixed pretrained weights and exposes no comparable task-specific pipeline program. The selection rule, source excerpt, and lineage are reported in \suppcomposition{}.

\subsection{Auditing pipelines as code}
\label{sec:audit}

Because the Section~\ref{sec:representation} constraints are prompted rather than enforced, we audited all \(1{,}020\) final programs with static detectors and manually reviewed every flag. No final pipeline triggered the detectors for internal hyperparameter search, nested AutoML, or hyperparameter-driven estimator switching. Of 24 flagged programs, 22 were confirmed violations: five candidate-authored out-of-fold constructions, thirteen four-learner ensembles, and four pipelines that refitted preprocessing during prediction. Only two confirmed violations occurred under LACE(GPT-5.4-mini), the backbone used for the primary baseline comparison. The audit is a targeted analysis rather than a proof of absence; detector definitions and complete classifications are reported in \suppcomposition{}. All flagged programs remain in the primary analysis, and excluding the confirmed violations leaves every coverage figure, mean-rank ordering, and Holm-corrected conclusion unchanged (\suppcomposition{}).

\subsection{Search dynamics and program evolution}
\label{sec:search-dynamics}

\begin{figure}[t]
\centering
\includegraphics[width=0.93\columnwidth]{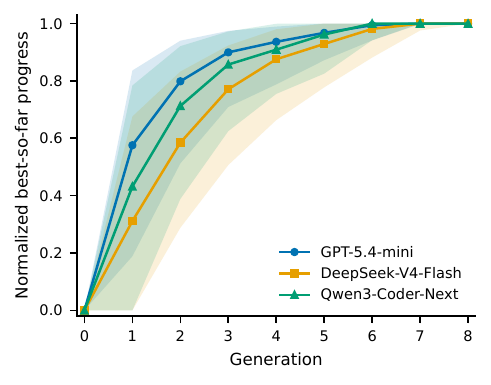}
\caption{Normalized best-so-far progress by generation, shown as the median and interquartile range across runs. Runs without a valid initial candidate or improvement are excluded.}
\label{fig:dynamics}
\end{figure}

Evolution continued throughout: the median run made its final improvement in generation 7 of 8 and had already realized \(91\%\) of its eventual gain by candidate 50 (Figure~\ref{fig:dynamics}). Selected pipelines had a median creation generation of 7 and a median ancestry depth of four mutation edges from their nearest generation-0 ancestor; \(52.6\%\) of finals appeared in the last two generations, and only \(1.3\%\) were unmodified initial candidates. Across valid parent–offspring comparisons, a median of 21.9\% of offspring improved on their parent, usually through small steps with a median gain of 0.23 percentage points on the accuracy-scored runs. Executability, exclusions, and complete lineage statistics are reported in \suppdynamics{}.

\section{Discussion and Limitations}
\label{sec:discussion}
\sysname{} shows that evolutionary search over executable pipeline programs is a viable route to AutoML: it is competitive with AutoGluon, outperforms the other search-based baselines, and covers all 68 tasks (Section~\ref{sec:results-coverage}). Relative to program-evolution AutoML that searches low-level primitives, such as AutoML-Zero, \sysname{} uses a coarser library-based representation that produced high candidate validity in our experiments (Section~\ref{sec:zero-shot-improvement}), though its search space remains bounded by the available libraries. Two properties of this result deserve emphasis. First, every \sysname{} score was obtained at the hyperparameter values chosen by the generator, with no external tuning; the comparison therefore measures the search over generated programs alone. Coupling the exposed parameter spaces to an external tuner is a natural extension. Second, a five-task ablation showed that the generated search space depends on what the prompt announces: an installed but unmentioned TabPFN component was never used, whereas explicitly announcing it led the generator to incorporate it, usually within ensembles (\supptabpfn{}). The ablation shows that the effective search space is shaped by announced components, not package availability alone.

The method inherits the capabilities and computational cost of its variation operator. The three backbones were retained from a seven-model screening that included both API-accessed and locally served models (\suppscreening{}), and LACE(GPT-5.4-mini) significantly outperformed the other two under the final configuration; the margins are small (median differences of \(0.06\) and \(0.18\) points). The campaign is computationally intensive by design because each generated program must be fitted and evaluated: it comprises \(102{,}000\) planned candidate evaluations, each subject to a one- or four-hour wall-clock limit. This cost did not advantage \sysname{} in the comparison, however: every baseline received the same per-task wall-clock budget \sysname{} consumed, so each had the same time available to produce a solution; the pretrained models require no candidate search but were nevertheless subject to the same external budget, and they must condition a transformer on the training data at every prediction, whereas a returned \sysname{} pipeline is a fitted classical model with negligible inference cost. These limits are caps rather than typical realized runtimes, and the heterogeneous cluster precludes a strict runtime comparison with prior work. Per-task wall-clock times are reported in \suppruntime{}. The search dynamics nevertheless indicate substantial scope for reducing cost: the median run had realized \(91\%\) of its eventual improvement by candidate 50, motivating shorter, adaptive, or early-stopped budgets.

Several limitations bound these conclusions. The evaluation covers a single benchmark suite of tabular classification tasks with five repetitions per configuration, on shared heterogeneous infrastructure that supports a common-budget rather than a strict compute-parity comparison. The reported results characterize \sysname{} at generator-chosen defaults and leave the interaction between generated structure and external numerical tuning unmeasured. The compliance audit is targeted static analysis with manual classification, not a proof of absence. Finally, the coarse task summary reduces, but cannot eliminate, the possibility that pretraining exposure to public benchmarks inflates the results.

\section{Conclusion}
\label{sec:conclusion}

We introduced \sysname{}, an AutoML framework that evolves complete executable pipeline programs using a language model as the variation operator. Across 68 OpenML tasks, the strongest \sysname{} variant is competitive with AutoGluon, outperforms the remaining search-based baselines, and covers tasks unsupported by several pretrained tabular models. The generated pipelines are diverse, improve through revision, and expose task-specific choices as editable code. The results also show that this effective program space is shaped by the components announced to the generator, rather than by package availability alone. These results establish executable programs as a promising search representation for AutoML. External tuning, adaptive budgets, and stronger backbones remain future work.

\section*{Code Availability}
The implementation of LACE, together with a notebook reproducing a minimal example on an OpenML task, is available at
\url{https://doi.org/10.5281/zenodo.21886859}.

\clearpage
% \section*{Ethical Statement}
% This work uses publicly available OpenML benchmark datasets under their published licenses and involves no human subjects or newly collected personal data. Generated candidate programs were executed in isolated subprocesses with resource limits, and all generated pipelines were retained for audit. The experimental campaign has a substantial computational footprint, which Section~\ref{sec:discussion} reports; the released artifacts allow reuse of its results without re-execution.  We do not foresee negative societal impacts specific to this method beyond those general to AutoML systems.

% \section*{Acknowledgments}
\bibliography{references}
\clearpage
\appendix
\section*{Supplementary Material}

This supplementary material provides additional implementation details,
configuration studies, analyses, and per-task results supporting the main paper.
The accompanying code artifact, available at the Zenodo
repository cited in the main paper, contains the source code, scripts, and representative artifacts needed to inspect the reproducibility of the reported results.

\paragraph{Guide.}
Sections A--D cover the search procedure, candidate evaluation, the task set, and baseline configurations. Sections E--G then document the exposed hyperparameter spaces and the small-scale studies behind the search configuration. The analyses follow: pipeline composition and the compliance audit in Section H, pairwise tests and task-size decomposition in Sections I--J, and zero-shot and search-dynamics results in Sections K--L. Section M reproduces the prompts, and Sections N--P report the foundation-model ablation, per-task wall-clock times, and per-task score distributions.

\paragraph{Experimental setting at a glance.}
The main study evaluates \sysname{} on 68 OpenML classification tasks with three language-model backbones (GPT-5.4-mini, DeepSeek-V4-Flash, and Qwen3-Coder-Next) and five repetitions per task and backbone. Each run uses a budget of $B=100$ candidates, $\mu=4$ parents, $\lambda=12$ offspring, no crossover, and a 20\% inner-validation split within each official training fold. Search uses only inner-validation scores; the official test fold is evaluated only after the final pipeline has been selected.

\section{Evolutionary Search Pseudocode}
\label{sec:algorithm-details}

Algorithm~\ref{alg:lace} gives the complete population-based search procedure summarized visually in the overview figure of the main paper. Candidate fitness is computed exclusively from inner-validation data during search. The official test folds are used only after the algorithm returns the selected pipeline.

\begin{algorithm}[!ht]
\caption{\sysname{} evolutionary search}
\label{alg:lace}

\begin{algorithmic}[1]
\STATE \textbf{Input:} budget $B$, parents $\mu$, offspring $\lambda$, mutation instructions $\mathcal{M}$
\STATE $P \gets \emptyset$
\FOR{$i = 1$ \TO $\mu$}
    \STATE $x \gets \textsc{LLM}(\text{task} + \text{interface} + \text{format})$
    \STATE $f(x) \gets \textsc{Evaluate}(x)$ \COMMENT{record exposed space; defaults used in this study}
    \STATE $P \gets P \cup \{x\}$
\ENDFOR

\STATE $x^\star \gets \arg\max_{x \in P} f(x)$

\WHILE{evaluated candidates $< B$}
    \STATE $O \gets \emptyset$
    \FOR{$j = 1$ \TO $\lambda$}
        \STATE $p \gets$ uniform random parent from $P$
        \STATE $m \gets$ uniform random instruction from $\mathcal{M}$
        \STATE $x \gets \textsc{LLM}(\text{task} + \mathrm{summary}(P) + p + \mathrm{feedback}(p) + m + \text{format})$
        \STATE $f(x) \gets \textsc{Evaluate}(x)$
        \STATE $O \gets O \cup \{x\}$
    \ENDFOR

    \STATE $P \gets$ best $\mu$ of $P \cup O$ \COMMENT{elitist truncation}

    \IF{$\max_{x \in P} f(x) > f(x^\star)$}
        \STATE $x^\star \gets \arg\max_{x \in P} f(x)$
    \ENDIF
\ENDWHILE

\STATE \textbf{return} $x^\star$; refit and evaluate once per official test fold

\end{algorithmic}
\end{algorithm}

\section{Candidate Evaluation Details}

\label{sec:eval-details}
Each candidate runs in a separate process with its own environment, under a wall-clock timeout covering the full evaluation: one hour for most tasks and four hours for the designated heavy tasks. A candidate evaluation is considered successful only when every inner-validation fold returns a finite score. If any fold fails, times out, or produces invalid predictions or a non-finite score, the entire candidate receives the worst fitness. Failure modes include compilation and import errors, a missing or malformed class, constructor and interface errors, invalid or wrongly shaped predictions, scoring errors, fold-level failures, and timeouts. The first captured error and its source line are returned to the generator, as illustrated in the main paper's pipeline-representation section.

A successful evaluation therefore always reports \texttt{failed\_splits=0}, and returns a summary of the form:
\begin{lstlisting}[basicstyle=\ttfamily\footnotesize,frame=none,columns=fixed,%
breaklines=false,aboveskip=3pt,belowskip=3pt]
SEARCH[inner_val] metric = mean +/- std |
    splits: folds=F | failed_splits=0
\end{lstlisting}

A failed evaluation reports the worst fitness together with the first captured error and its source line, for example:
\begin{lstlisting}[basicstyle=\ttfamily\footnotesize,frame=none,columns=fixed,%
 breaklines=false,aboveskip=3pt,belowskip=3pt]
SEARCH[inner_val] predictive_accuracy =
    -inf +/- nan |
    splits: folds=10 | failed_splits=10
First failed split: line 75, in __call__:
    AttributeError: 'RidgeClassifier'
    object has no attribute 'predict_proba'
\end{lstlisting}

\section{OpenML Task Set and Timeout Policy}
\label{sec:openml-task-appendix}
The experiments use the 68 tasks of the OpenML \texttt{amlb-classification-all} suite:

\begin{lstlisting}[basicstyle=\ttfamily\footnotesize,frame=none,columns=fixed,breaklines=true,aboveskip=3pt,belowskip=3pt]
2073, 3945, 7593, 10090, 146818, 146820, 167120, 168350, 168757, 168784, 168868, 168909, 168910, 168911, 189354, 189355, 189356, 189922, 190137, 190146, 190392, 190410, 190411, 190412, 211979, 211986, 359953, 359954, 359955, 359956, 359957, 359958, 359959, 359960, 359961, 359962, 359963, 359964, 359965, 359966, 359967, 359968, 359969, 359970, 359971, 359972, 359973, 359974, 359975, 359976, 359977, 359979, 359980, 359981, 359982, 359983, 359984, 359985, 359987, 359989, 359990, 359991, 359992, 359993, 359994, 360112, 360113, 360114.
\end{lstlisting}
Dataset sizes range from 100 instances (task 190412, with 10,000 features) to 4,898,431 instances (task 360112, 41 features); the smallest feature count is 4 (task 359955). The following 20 tasks meet the heavy-tier criterion \(n \geq 400{,}000\), \(d \geq 1{,}000\), or \(nd \geq 10{,}000{,}000\), and receive a four-hour per-candidate timeout; the remaining 48 receive one hour:
\begin{quote}\small\ttfamily
3945, 7593, 10090, 168868, 168909, 189354, 189355, 189356, 190412, 359953, 359966, 359967, 359973, 359976, 359985, 359989, 359994, 360112, 360113, 360114.
\end{quote}
The tier changes only the per-candidate wall-clock limit, not the search budget, number of repetitions, inner-validation fraction, or metric. If any split fails or times out, the evaluator records the entire candidate evaluation with the worst fitness and returns the first captured error as feedback. Per-split outcomes are retained for diagnostics, but candidate fitness is averaged over the predefined folds only when every fold returns a finite score. A final selected pipeline that fails at official-test evaluation is likewise recorded with the worst fitness rather than silently skipped.

\section{Baseline Configurations}
\label{sec:baseline-config}
Table~\ref{tab:baseline-configs} summarizes the baseline configurations. All baselines were evaluated on the 68 OpenML tasks with five repetitions per task, using the official OpenML split instances and the task-specific evaluation metric: \texttt{predictive\_accuracy} for 67 tasks and \texttt{area\_under\_roc\_curve} for task 360114. Each baseline received, per task and repetition, the same total wall-clock budget \sysname{} used on that task, divided dynamically across the remaining pre-defined splits.

\begin{table*}[t]
\centering
\small
\begin{tabular}{@{}>{\raggedright\arraybackslash}p{0.14\textwidth}%
>{\raggedright\arraybackslash}p{0.10\textwidth}%
>{\raggedright\arraybackslash}p{0.68\textwidth}@{}}
\toprule
Baseline & Version & Main configuration \\
\midrule
AutoGluon-Tabular & 1.4.0 &
\texttt{high\_quality} preset, \texttt{num\_gpus=0},
\texttt{dynamic\_stacking=False}, and \texttt{num\_stack\_levels=1};
task budget divided across remaining pre-defined splits. Search space restricted to GBM, CAT, XGB, RF, XT, KNN, and LR; neural and
foundation-model families (\texttt{NN\_TORCH}, \texttt{FASTAI},
\texttt{REALMLP}, \texttt{TABM}, \texttt{TABPFNV2}, \texttt{TABICL},
\texttt{MITRA}) excluded. Each split records the fitted model names and the run fails if any excluded family appears. \\
auto-sklearn & 0.15.0 &
\texttt{time\_left\_for\_this\_task} set to the per-split budget; per-model time limit set to half the per-split budget (minimum 30 seconds). \\
H2O AutoML & 3.46.0.10 &
Allowed algorithms GLM, DRF, GBM, and StackedEnsemble, with DeepLearning excluded; a 20\% inner-validation split (seeded by repetition) within each official training fold is used for model selection. \\
TabPFN v2.5 & \texttt{tabpfn} 6.0.6 &
\texttt{create\_default\_for\_version(V2\_5)} on GPU; full official folds with no row or feature subsampling; \texttt{ignore\_pretraining\_limits=False}; external wall-clock timeout. \\
TabPFN v3 & \texttt{tabpfn} 8.0.3 &
OSS default classifier with \texttt{n\_estimators=8}, \texttt{device=cuda}, \texttt{fit\_mode=fit\_preprocessors}, \texttt{memory\_saving\_mode=auto}, \texttt{ignore\_pretraining\_limits=False}; full official folds, no external HPO. \\
TabICL & \texttt{tabicl} 2.1.1 &
\texttt{TabICLClassifier} with
\texttt{n\_estimators=8},
\texttt{device=cuda},
\texttt{batch\_size=1},
\texttt{kv\_cache=False},
\texttt{offload\_mode=auto},
and \texttt{random\_state} set to the repetition seed;
full official OpenML folds with no manual row or feature subsampling and no
external HPO; external wall-clock timeout matched to the corresponding
\sysname{} task repetition. Remaining constructor options use the package
defaults. \\
XGBoost & 3.2.0 &
Fixed scikit-learn pipeline: median imputation (numeric), most-frequent imputation and one-hot encoding (categorical), then \texttt{XGBClassifier}; no HPO or internal CV. \\
\bottomrule
\end{tabular}
\caption{Baseline configurations used in the OpenML experiments.}
\label{tab:baseline-configs}
\end{table*}

TabPFN~v2.5, TabPFN~v3, and TabICL were evaluated on the full official training folds without manual row or feature subsampling or external hyperparameter optimization. Each was subject to the external wall-clock limit matched to the corresponding \sysname{} task repetition. For AutoGluon, dynamic stack-level selection was disabled before the full benchmark and the depth was fixed to one stack level. DyStack selected the same depth in the preliminary smoke test. Fixing this choice avoids an additional diagnostic sub-fit within every official split while retaining stacked ensembling. The fixed XGBoost baseline uses \texttt{n\_estimators=500}, \texttt{max\_depth=6}, \texttt{learning\_rate=0.05}, \texttt{subsample=0.8}, \texttt{colsample\_bytree=0.8}, and \texttt{tree\_method=hist}, with four threads. For the search-based AutoML baselines, the evaluation environments did not include deep-learning libraries such as PyTorch, FastAI, TensorFlow, or Keras; TabPFN was evaluated in its own environment. TabPFN v2.5 and v3 are run on the full official folds without disabling their pretraining-size limits (roughly 50{,}000 rows and 2{,}000 features for v2.5, one million rows and 2{,}000 features for v3); tasks exceeding these limits are not truncated and instead appear as failed splits, runner failures, or timeouts.

\section{Exposed Hyperparameter Spaces}
\label{sec:hpo-outcome}
The main paper notes that each final pipeline exposes a median of 12 tunable constructor arguments. This section details how those spaces were produced. The initialization and mutation prompts request a hyperparameter-space block alongside each pipeline, and the generators complied: at least 97.5\% of candidates emitted a nonempty \texttt{\# Space:} block. In the experiments reported here, all pipelines are evaluated at their generator-chosen default arguments, and these exposed spaces are not tuned. The emitted spaces remain part of the returned artifacts and can be tuned outside the framework as a direct extension of the approach.

\section{Configuration Ablations}
\label{sec:ablations}
We conducted small-scale screening experiments before the full benchmark runs to examine crossover, offspring count, parent selection, mutation instructions, and backbone choice. The backbone screening used tasks 190146, 359970, and 359974, which are also contained in the 68-task benchmark. We therefore report these experiments only to document the design choices and do not treat them as an independent held-out evaluation. We varied one aspect at a time.

\paragraph{Crossover.}
We compared crossover rates of $0$, $0.25$, and $0.75$ with $\mu=4$ parents and $\lambda=4$ offspring. No setting showed a consistent advantage: a rate of $0.75$ was nominally best, but the difference from disabling crossover was within run-to-run variation on the development tasks. We therefore disable crossover and rely on mutation alone, rather than adding a mechanism without a demonstrated benefit.

\paragraph{Offspring count and parent selection.}
Increasing the offspring count from $4$ to $12$ (at $\mu=4$) improved development-set performance, so we adopt $\lambda=12$. Replacing uniform-random parent selection with tournament selection (size $2$) did not improve over random selection on these tasks, so we keep random selection.

\paragraph{Mutation instructions.}
We compared the three-instruction set used in the main runs (change the estimator family, refine the preprocessing, adjust a small ensemble) against two alternatives: removing the model-family instruction, and a variant emphasizing preprocessing and local hyperparameter changes over larger structural edits. The three-instruction set performed best, winning on all three development tasks; removing the model-family instruction was clearly worse, and the preprocessing-focused variant recovered part of that drop but remained below the full set.

\paragraph{Backbone selection.}
We screened seven language models as the variation operator and retained three for the main study based on screening-task performance and candidate validity, with inference cost and deployment mode used as practical selection criteria. We selected \texttt{gpt-5.4-mini} as the strongest cost-effective commercial model, \texttt{deepseek-v4-flash} as a competitive lower-cost API model, and \texttt{qwen3-coder-next} as the strongest locally served model. The screening also included \texttt{qwen3.5-27b}, \texttt{nemotron-3-super}, \texttt{nemotron-3-nano-30b}, and \texttt{gemini-2.5-flash}. Task-wise screening results and the selection rationale are reported in Appendix~\ref{sec:backbone-selection}.

\section{Backbone Screening}
\label{sec:backbone-selection}

\begin{figure*}[tb]
\centering
\includegraphics[width=\textwidth]
{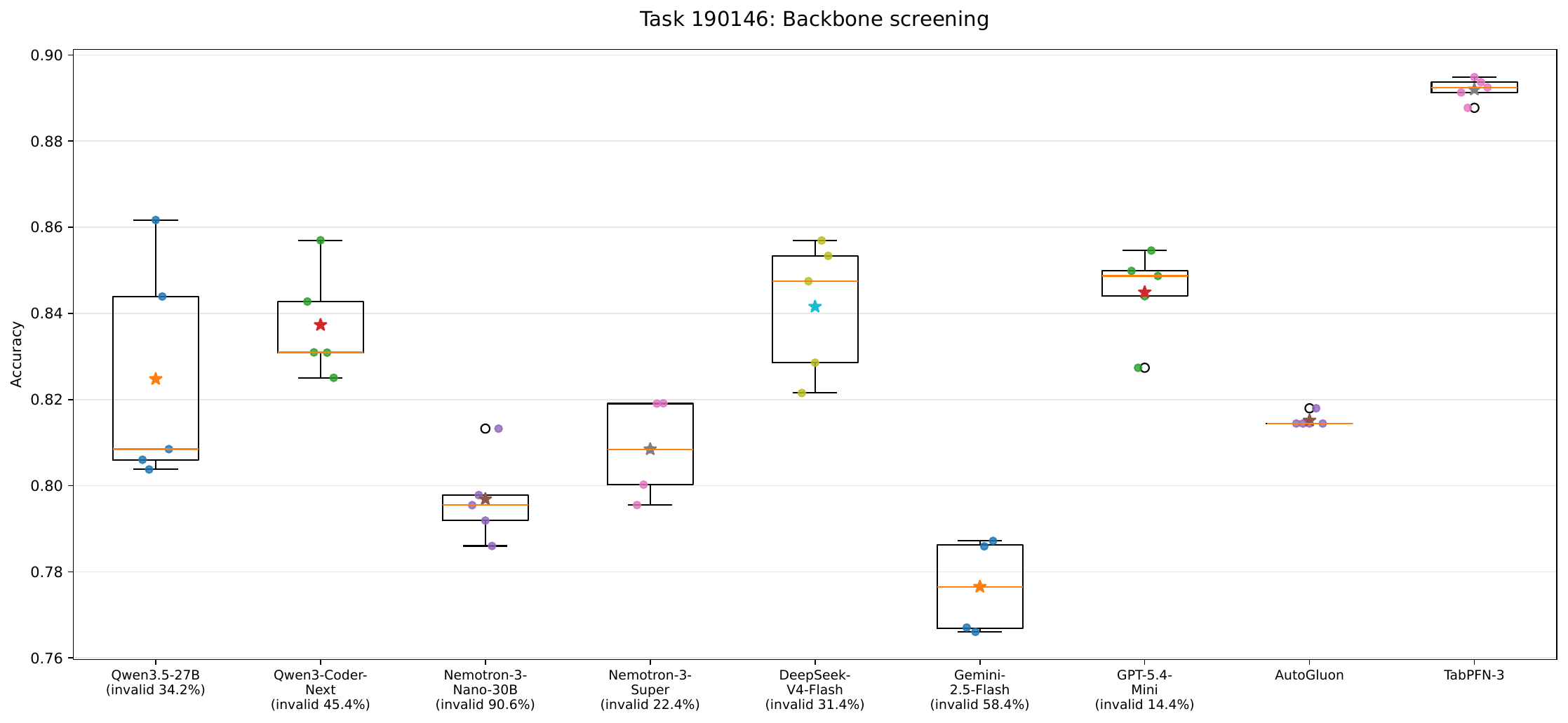}
\caption{Backbone screening on OpenML task 190146. Boxplots and individual
points summarize the best inner-validation fitness obtained over five
repetitions. Percentages below the candidate language-model names report the
fraction of generated candidates assigned non-finite fitness. An earlier harness configuration, AutoGluon, and TabPFN are shown only as contextual
references and were not candidates in the backbone selection.}
\label{fig:backbone-screening-190146}
\end{figure*}

\begin{figure*}[tb]
\centering
\includegraphics[width=\textwidth]
{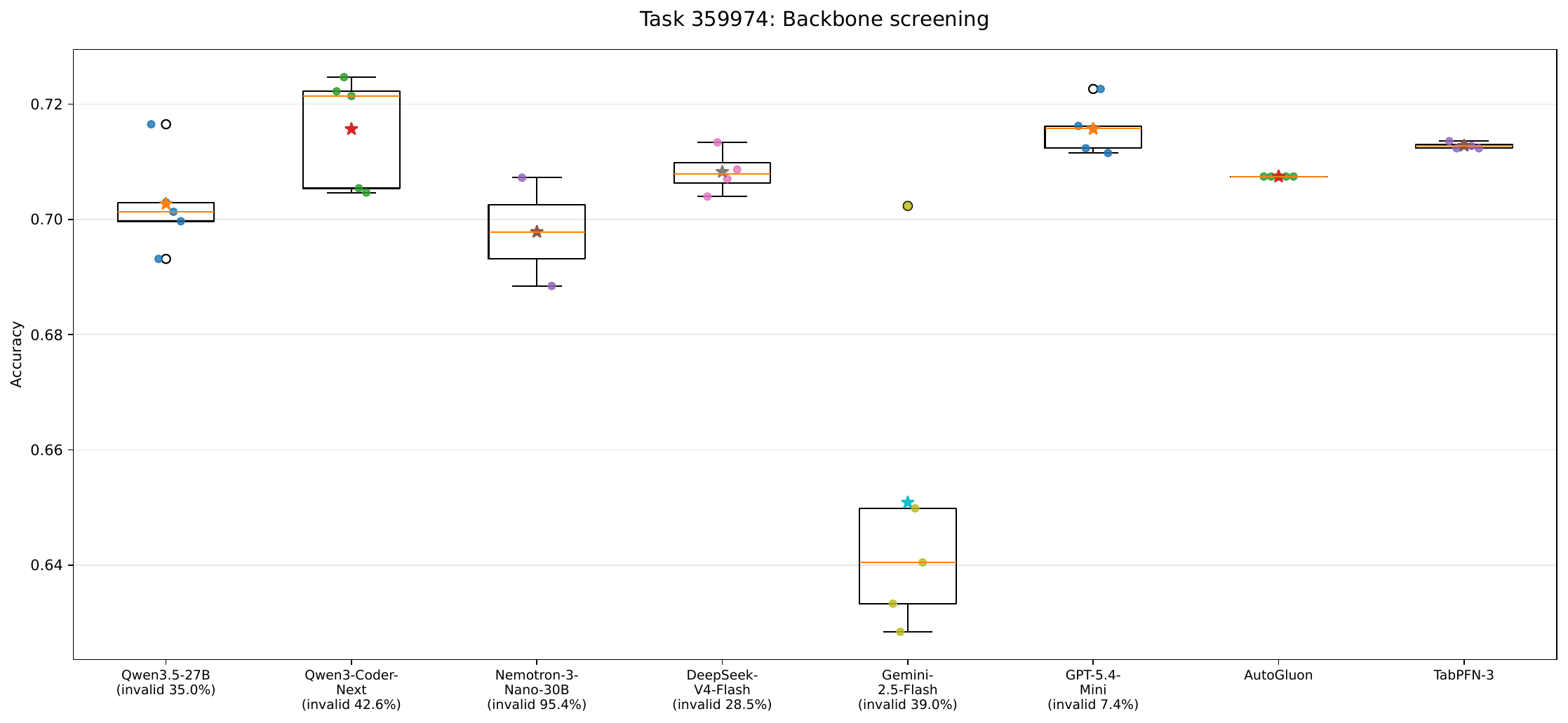}
\caption{Backbone screening on OpenML task 359974, using the same protocol and
notation as Figure~\ref{fig:backbone-screening-190146}.}
\label{fig:backbone-screening-359974}
\end{figure*}

Before the full benchmark runs, we screened seven candidate language models as the variation operator on OpenML tasks 190146, 359970, and 359974. Each model was evaluated over five repetitions under the screening configuration. Figures~\ref{fig:backbone-screening-190146}--% 
\ref{fig:backbone-screening-359974} show the best inner-validation fitness reached in each repetition. The percentages shown below the model names report the fraction of generated candidates assigned non-finite fitness, and therefore summarize candidate validity separately from the quality of the best valid pipeline.

An earlier harness configuration, AutoGluon, and TabPFN are included in the figures only as contextual reference points; they were not candidates in the language-model selection. Because the three screening tasks are also contained in the main benchmark, these results are used only to explain the backbone choice and not as independent evidence of generalization.

Across the three tasks, \texttt{gpt-5.4-mini} achieved the strongest or near-strongest screening performance among the candidate language models and also generated the smallest fraction of invalid candidates. Among the locally served models, \texttt{qwen3-coder-next} consistently achieved the strongest best-pipeline fitness, although with a higher invalid-candidate rate than \texttt{qwen3.5-27b}. \texttt{deepseek-v4-flash} provided competitive performance as a lower-cost API-accessed alternative. We retained one strong commercial model, one lower-cost API model, and the strongest locally served model. The selection therefore reflects performance, candidate validity, cost, and deployment mode rather than the ranking on any single task.

\begin{figure*}[tb]
\centering
\includegraphics[width=\textwidth]
{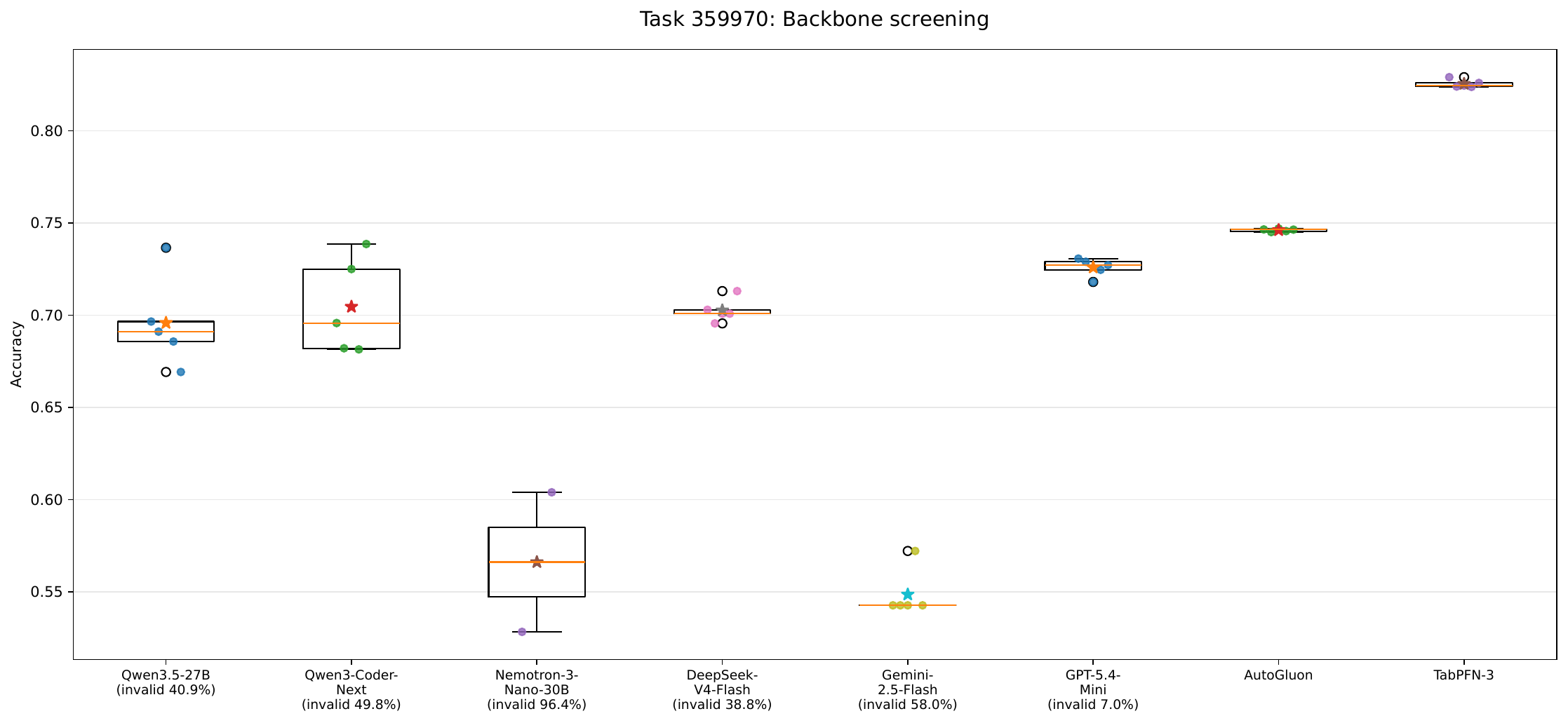}
\caption{Backbone screening on OpenML task 359970, using the same protocol and
notation as Figure~\ref{fig:backbone-screening-190146}.}
\label{fig:backbone-screening-359970}
\end{figure*}

\section{Pipeline Analysis Details}
\label{sec:composition-appendix}
This section documents the static analysis behind the composition and audit results reported in the main paper.
 
\paragraph{Coverage and conventions.}
The experiment comprises \(68\) tasks, five repetitions per task, and three language-model backbones, yielding \(68 \times 5 \times 3 = 1{,}020\) completed runs: 340 each for DeepSeek, GPT-5.4-mini, and Qwen3. With a budget of 100 candidates per run, these runs correspond to 102{,}000 planned candidate evaluations. The logs contain 101{,}959 candidate records because 17 runs recorded between 95 and 99 of their 100 candidates. Of the observed records, 70{,}998 (\(69.6\%\)) received finite fitness and are treated as valid candidates. Relative to the 34{,}000 planned candidate slots per backbone, this comprises 25{,}396 valid candidates for DeepSeek (\(74.7\%\)), 29{,}691 for GPT-5.4-mini (\(87.3\%\)), and 15{,}911 for Qwen3 (\(46.8\%\)). The AST-based analyzer parsed 101{,}674 programs (\(99.7\%\)); a regex fallback handled the remaining 285, all of which were invalid candidates, and no program failed both parsers. Finally, the source hash of every traced final pipeline matched its corresponding candidate record.

Table~\ref{tab:composition-overview-appendix} gives an overview of the final pipelines before the more detailed family, preprocessing, ensemble, and audit breakdowns below.

\begin{table}[t]
\centering
\small
\begin{tabular}{lrrrrr}
\toprule
Backbone & Finals & Preproc. & Ensemble & Custom & Flags \\
\midrule
DeepSeek
    & 340 & 100.0\% & 72.6\% & 3.2\% & 7 \\
GPT-5.4-mini
    & 340 & 100.0\% & 70.2\% & 2.9\% & 2 \\
Qwen3
    & 340 & 99.7\% & 77.1\% & 15.9\% & 15 \\
\midrule
Overall
    & 1{,}020 & 99.9\% & 73.3\% & 7.4\% & 24 \\
\bottomrule
\end{tabular}
\caption{Overview of the \(1{,}020\) traceable final pipelines. Preproc. is the share with at least one detected preprocessing construction beyond the evaluator's fixed preprocessing; Ensemble is the share combining at least two resolved base learners; Custom is the share whose estimator logic is hand-written rather than assigned to a recognized library family; Flags counts programs marked by at least one audit detector before manual classification.} \label{tab:composition-overview-appendix}
\end{table}

\paragraph{Sensitivity to confirmed constraint violations.}
The primary analysis retains every selected final program. To test whether the 22 confirmed violations affect the aggregate conclusions, we marked the corresponding repetition-level official-test scores as missing without assigning an artificial worst value. On top of that, we recomputed each method–task result under the same rule as the main analysis, requiring at least three valid repetitions of five. The exclusions are concentrated in the weaker backbones: 2 of 340 repetitions for GPT-5.4-mini, 7 for DeepSeek, and 13 for Qwen3. Twenty method–task pairs were affected, eighteen losing one repetition and two losing two; every pair retained at least three valid repetitions, so no method lost coverage on any task. Aggregate positions move by at most 0.02 in mean rank and 0.2 percentage points in normalized deviation. No coverage figure, fractional-win count, or mean-rank ordering changes on either the common-task or the full-benchmark analysis; the only ordering change is that DeepSeek and AutoGluon exchange positions under fixed-45 normalized deviation, a pair that bears on no reported claim. All nine Holm-corrected pairwise comparisons keep the same verdict: \sysname{} with GPT-5.4-mini shows no detectable difference from AutoGluon or TabPFN~v2.5, significantly outperforms auto-sklearn, XGBoost, and H2O, is significantly outperformed by TabPFN~v3 and TabICL, and significantly outperforms the other two backbones. Only two of the confirmed violations occur under GPT-5.4-mini, the backbone used in the primary comparison, and excluding all confirmed violations leaves the reported statistical conclusions unchanged.

\paragraph{Detectors.}
Each program is parsed and scanned for the six prompt constraints, with helper-method and dataflow reachability so that constructs inside methods reachable from \texttt{\_\_init\_\_} or \texttt{\_\_call\_\_} are attributed to fitting or prediction respectively. Internal cross-validation flags calls to model-selection and evaluation utilities (\texttt{cross\_val\_score}, \texttt{cross\_val\_predict}, \texttt{cross\_validate}, \texttt{KFold}-family splitters, and estimators with built-in cross-validated selection such as \texttt{RidgeCV}). Internal hyperparameter search flags \texttt{GridSearchCV}-style classes, references to \texttt{optuna}, \texttt{hyperopt}, or \texttt{skopt}, and hand-written sweep loops that iterate over literal collections while fitting models and tracking scores; sweeps expressed through \texttt{range} or \texttt{numpy} spacing functions are outside this detector's scope. For nested AutoML we look for references to auto-sklearn, AutoGluon, H2O, TPOT, or FLAML. Estimator switching flags conditionals on constructor arguments that instantiate different estimator families in their branches, though dictionary-dispatch switching falls outside this detector's scope. Ensemble size resolves the literal length of the estimator list passed to \texttt{VotingClassifier} or \texttt{StackingClassifier}; 74 of 759 ensembles used a non-literal list and are recorded as unresolved, and twelve records resolving to sizes of zero or one were inspected as detector edge cases rather than interpreted literally. To detect fitting at prediction time, we flag any \texttt{fit} or \texttt{fit\_transform} call reachable from \texttt{\_\_call\_\_}. Cross-fitting internal to permitted components, \texttt{StackingClassifier} and \texttt{CalibratedClassifierCV} without \texttt{cv="prefit"}, is recorded in separate descriptive fields and is not counted as a violation. Every final carrying any flag was exported with its source and classified manually. The audit is a targeted static analysis of these stated detectors, not a proof of absence.
 
\paragraph{Estimator families and preprocessing.}
Families are counted per pipeline, so an ensemble contributes to several. Table~\ref{tab:family-appendix} gives the full distribution for valid candidates and final pipelines, and Table~\ref{tab:preproc-appendix} the added-preprocessing categories. The class-imbalance column is a keyword heuristic (\texttt{class\_weight} or resampling constructs) and should be read as indicative. Final pipelines contained a median of three added preprocessing constructions (interquartile range 2--4) and a median of 12 tunable constructor arguments (DeepSeek 12, GPT-5.4-mini 19, Qwen3 9).
 
\begin{table*}[t]
\centering
\small
\caption{Estimator-family occurrence among valid candidates (Cand.) and final pipelines (Final), in percent. Denominators: DeepSeek 25{,}396 and 340; GPT-5.4-mini 29{,}691 and 340; Qwen3 15{,}911 and 340; overall 70{,}998 and 1{,}020.}
\label{tab:family-appendix}
\begin{tabular}{lrrrrrrrr}
\toprule
& \multicolumn{2}{c}{DeepSeek} & \multicolumn{2}{c}{GPT-5.4-mini}
& \multicolumn{2}{c}{Qwen3} & \multicolumn{2}{c}{Overall} \\
Family & Cand. & Final & Cand. & Final & Cand. & Final & Cand. & Final \\
\midrule
Gradient boosting     & 70.8 & 75.3 & 51.7 & 66.7 & 56.3 & 66.2 & 59.6 & 69.4 \\
Forest / extra trees  & 35.5 & 37.1 & 46.0 & 64.3 & 49.9 & 56.2 & 43.1 & 52.5 \\
Linear                & 58.5 & 55.3 & 60.4 & 44.8 & 74.4 & 67.4 & 62.8 & 55.8 \\
SVM                   & 21.1 & 19.1 & 21.7 & 18.0 & 16.6 & 15.9 & 20.3 & 17.7 \\
k-NN                  &  4.3 &  4.1 &  2.1 &  2.9 &  4.9 &  5.9 &  3.5 &  4.3 \\
Naive Bayes           &  0.8 &  0.9 &  0.5 &  0.3 &  4.4 &  2.9 &  1.5 &  1.4 \\
Decision tree         &  1.3 &  2.1 &  1.2 &  1.2 &  0.6 &  0.3 &  1.1 &  1.2 \\
Other scikit-learn    &  4.2 &  5.9 &  0.5 &  0.6 &  2.3 &  1.5 &  2.2 &  2.6 \\
Custom / unrecognized &  3.6 &  3.2 &  3.5 &  2.9 & 16.5 & 15.9 &  6.4 &  7.4 \\
\bottomrule
\end{tabular}
\end{table*}
 
\begin{table}[t]
\centering
\small
\caption{Added preprocessing among final pipelines, in percent of each backbone's finals.}
\label{tab:preproc-appendix}
\begin{tabular}{lrrrr}
\toprule
Category & DeepSeek & GPT & Qwen3 & Overall \\
\midrule
Any preprocessing   & 100.0 & 100.0 & 99.7 & 99.9 \\
Scaling             & 100.0 &  86.7 & 98.5 & 95.1 \\
Feature selection   &  77.4 &  73.5 & 55.3 & 68.7 \\
Dimensionality red. &  27.1 &   3.8 &  8.8 & 13.2 \\
Nonlinear transforms&  85.0 &  54.9 & 59.7 & 66.5 \\
Class imbalance     &  23.2 &  59.6 & 67.9 & 50.2 \\
\bottomrule
\end{tabular}
\end{table}

\paragraph{Selection pressure.}
Selection shifted the composition toward tree ensembles. Gradient boosting rose from \(59.6\%\) of valid candidates to \(69.4\%\) of final pipelines, and forests or extra trees rose from \(43.1\%\) to \(52.5\%\), while linear models fell from \(62.8\%\) to \(55.8\%\). The shift was strongest under GPT-5.4-mini, for which forests increased by \(18.3\) percentage points and linear models decreased by \(15.5\) points. These are composition shifts under selection, not causal effects of any single design choice.

\paragraph{Structural diversity.}
A pipeline's structural signature comprises its ordered preprocessing steps, its set of estimator families, its ensemble type, and the names of its exposed hyperparameters, with numerical values ignored. The \(70{,}998\) valid candidates produced \(65{,}298\) unique signatures. The most frequent signature appeared in 23 candidates, and the five most frequent together accounted for only \(0.1\%\). A median run produced 68 distinct structures among its 100 candidates, increasing to 86 under GPT-5.4-mini. At most eight candidate signatures were shared by any pair of backbones, none appeared under all three, and no final signature was produced by more than one backbone. The signature includes exposed hyperparameter names, so some of the measured uniqueness may reflect naming variation. The estimator-family distributions and within-run structure counts show the same pattern at coarser levels.
 
\paragraph{Ensembles.}
Of the 1{,}020 finals, 759 constructed an ensemble: 453 voting, 231 stacking, and 75 hand-built combinations (weighted or averaged predictions without the standard wrapper classes); 747 finals (73.3\%) combined at least two resolved base learners. The backbones differed sharply: voting made up 63.7\% of GPT-5.4-mini's finals but 24.7\% of Qwen3's, which preferred stacking (37.1\%) and hand-built combination (15.9\%). Among the 673 ensembles resolving to a literal estimator list of at least two, 353 used two base learners, 308 used three, and 12 used four (the ensemble-size flags of the main paper's audit section); the median size is two (interquartile range 2--3).
 
\paragraph{Beyond-library code.}
Manual reading of all 75 custom-family finals shows each implements hand-written prediction combination over its base learners, 43 over two learners, 31 over three, and one over four, computing weighted or averaged probabilities in NumPy in place of the corresponding wrapper class. A broad screening heuristic for hand-written transformation code (numeric array operations or element-wise mappings inside fitting and prediction methods) flagged 42.7\% of finals and 32.5\% of valid candidates; the heuristic is deliberately liberal, is not a validated prevalence estimate, and is reported only to bound the phenomenon.

\paragraph{Illustrative inspectability example.}
A source-level adjudication identified 89 of the \(1{,}020\) finals (\(8.7\%\), spanning 53 tasks) that combine predictions directly in NumPy or ordinary Python. This extends the reachability-based inventory of 75 hand-built combinations reported above: manual review found 14 additional active implementations the reachability pass missed, of which 82 of the 89 satisfy all audited constraints. Among these compliant finals, we selected the clearest compact source instance from the primary GPT-5.4-mini backbone for illustration. The selected program is the final pipeline for OpenML task 2073, repetition 4 (candidate \texttt{0ce14317-05e0-4fed-9eb4-3b7852b46cb2}). It obtained official-test score \(0.6206\) and final inner-validation fitness \(0.6263\); its traced lineage improved from \(0.5964\) to \(0.6203\) to \(0.6263\). The intermediate candidate introduced hand-written logistic-regression and ExtraTrees probability averaging with weights \(0.55/0.45\); the final candidate replaced these with an SVC and random forest combined at \(0.60/0.40\).
\begin{lstlisting}[language=Python,basicstyle=\ttfamily\footnotesize,
frame=single,columns=fixed,aboveskip=3pt,belowskip=3pt]
def __call__(self, X):
    Xp = self._transform(X)
    p_svc = self.svc.predict_proba(Xp)
    p_rf = self.rf.predict_proba(Xp)
    p = 0.60 * p_svc + 0.40 * p_rf
    return np.argmax(p, axis=1)
\end{lstlisting}
The returned predictions depend directly on this block, so the coefficients and members can be changed, removed, or reused without altering the search procedure. Both lineage transitions also changed other pipeline components, so the observed fitness increases are not attributed to the combination rule alone. The example demonstrates source-level visibility and editability, not the predictive effect of a particular weighting rule.

\paragraph{Compliance detail and manual classification.}
Table~\ref{tab:compliance-appendix} breaks the 24 audit flags down by type and backbone. No final pipeline triggered the internal hyperparameter-search, nested-AutoML, or estimator-switching detectors. Manual review classified two of the seven cross-validation flags as permitted uses of explicit splitter objects with \texttt{CalibratedClassifierCV}; the remaining five were candidate-authored out-of-fold constructions, comprising three DeepSeek and two Qwen3 finals. All thirteen four-learner ensembles and all four cases of fitting during prediction were confirmed violations, giving 22 in total: seven under DeepSeek, two under GPT-5.4-mini, and thirteen under Qwen3. Because the primary baseline comparison uses GPT-5.4-mini, only two confirmed violations enter that comparison; the remaining cases affect the alternative-backbone results. All flagged pipelines are retained in the reported results.

Cross-fitting internal to permitted components is recorded separately and is not treated as a violation. It occurred in \(21.9\%\) of finals through \texttt{StackingClassifier} and in \(35.2\%\) through probability calibration, with calibration use reaching \(70.0\%\) under Qwen3. The audit covers the stated detectors and subsequent manual classifications; it is not a proof that no other form of noncompliance exists.
 
\begin{table}[t]
\centering
\small
\caption{Audit flags among final pipelines by type and backbone, with the outcome of manual classification, and cross-fitting recorded descriptively. The internal hyperparameter search, nested AutoML, and estimator-switching detectors flagged no final and are omitted.}
\label{tab:compliance-appendix}
\begin{tabular}{lrrrr}
\toprule
& DeepSeek & GPT & Qwen3 & Overall \\
\midrule
Finals analyzed        & 340 & 340 & 340 & 1{,}020 \\
Internal CV constructs &   3 &   0 &   4 &   7 \\
Ensemble of four       &   4 &   2 &   7 &  13 \\
Fit at prediction time &   0 &   0 &   4 &   4 \\
Any flag               &   7 &   2 &  15 &  24 \\
Confirmed violations   &   7 &   2 &  13 &  22 \\
\midrule
Stacking cross-fit (\%)    & 25.6 &  3.5 & 36.5 & 21.9 \\
Calibration cross-fit (\%) & 18.8 & 16.8 & 70.0 & 35.2 \\
\bottomrule
\end{tabular}
\end{table}

\FloatBarrier
\section{Pairwise Statistical Comparisons}
\label{sec:statistical-results}
\begin{table*}[!t]
\centering
\small
\begin{tabular}{lrrrrrr}
\toprule
Method &
Full rank (68) \(\downarrow\) &
Full dev. (68) \(\downarrow\) &
Common rank (45) \(\downarrow\) &
Common dev. (45) \(\downarrow\) &
Frac. wins \(\uparrow\) &
Coverage \\
\midrule
TabICL                     & 3.28 & 18.7\% & 2.97 & 15.4\% & 14.75 & 67/68 \\
TabPFN v3                  & 3.91 & 26.1\% & 3.32 & 20.0\% & 18.25 & 62/68 \\
\sysname{} (GPT-5.4-mini)  & 4.50 & 32.3\% & 4.71 & 32.8\% & 3.0   & 68/68 \\
AutoGluon-Tabular          & 4.59 & 31.9\% & 5.49 & 38.8\% & 14.0  & 68/68 \\
\sysname{} (DeepSeek)      & 5.76 & 41.1\% & 5.76 & 38.8\% & 0.0   & 68/68 \\
\sysname{} (Qwen3)         & 5.99 & 45.5\% & 5.88 & 40.7\% & 2.75  & 68/68 \\
TabPFN v2.5                & 6.10 & 51.6\% & 4.19 & 26.9\% & 4.25  & 45/68 \\
auto-sklearn               & 6.29 & 43.9\% & 7.24 & 52.5\% & 1.0   & 68/68 \\
XGBoost                    & 6.49 & 53.4\% & 6.69 & 55.4\% & 9.0   & 68/68 \\
H2O AutoML                 & 8.09 & 76.8\% & 8.76 & 80.1\% & 1.0   & 68/68 \\
\bottomrule
\end{tabular}
\caption{Aggregate performance and coverage. Common statistics use the 45 tasks completed by all methods; full statistics use all 68 under the failure-aware convention of the method-comparison figure of the main paper, penalizing missing results with the tied average of the remaining worst ranks and normalized deviation 1. A task is valid when at least three of five repetitions return a finite score. Fractional wins split tied task wins equally among the tied methods. Methods are ordered by full-benchmark rank.}
\label{tab:aggregate-results-supp}
\end{table*}
Table~\ref{tab:aggregate-results-supp} reports the aggregate rank, deviation, fractional-win, and coverage values summarized by the method-comparison figure of the main paper.

For each task and method, we first average the task-specific evaluation score over the successful repetitions (a task enters a comparison only if both methods have a valid score under the validity rule of the main paper's results section). We then compare \sysname{} with GPT-5.4-mini against each baseline using a paired Wilcoxon signed-rank test over tasks for which both methods return a valid score. Missing tasks are excluded from the corresponding pairwise comparison rather than counted as losses. We apply Holm's correction separately within two pre-specified families, the seven baseline comparisons and the two backbone comparisons, controlling the family-wise error rate at \(\alpha=0.05\) within each family.

\begin{table*}[!t]
\centering
\small
\begin{tabular}{lrrrrrr}
\toprule
Comparison &
Shared tasks &
W--T--L &
Median diff. &
Mean diff. &
Raw \(p\) &
Holm \(p\) \\
\midrule
AutoGluon
    & 68 & 33--0--35 & \(-0.001\) & \(-0.268\)
    & \(0.66\) & \(0.66\) \\
auto-sklearn
    & 68 & 50--0--18 & \(+0.234\) & \(+0.381\)
    & \(0.000442\) & \(0.00133\) \\
XGBoost
    & 68 & 48--0--20 & \(+0.370\) & \(+1.233\)
    & \(4.97\times10^{-5}\) & \(0.000298\) \\
H2O AutoML
    & 68 & 61--0--7 & \(+1.024\) & \(+3.241\)
    & \(3.92\times10^{-10}\) & \(2.75\times10^{-9}\) \\
TabPFN v2.5
    & 45 & 16--0--29 & \(-0.144\) & \(-0.422\)
    & \(0.0997\) & \(0.199\) \\
TabPFN v3
    & 62 & 15--0--47 & \(-0.336\) & \(-1.019\)
    & \(9.02\times10^{-5}\) & \(0.000451\) \\
TabICL
    & 67 & 18--0--49 & \(-0.260\) & \(+0.221\)
    & \(0.000284\) & \(0.00114\) \\
DeepSeek
    & 68 & 48--0--20 & \(+0.056\) & \(+0.591\)
    & \(0.000453\) & \(0.000453\) \\
Qwen3-Coder-Next
    & 68 & 55--0--13 & \(+0.177\) & \(+0.690\)
    & \(6.39\times10^{-8}\) & \(1.28\times10^{-7}\) \\
\bottomrule
\end{tabular}
\caption{Paired Wilcoxon signed-rank comparisons between \sysname{} with GPT-5.4-mini and the external baselines and alternative \sysname{} backbones. W--T--L gives task-level wins, exact ties, and losses for GPT-5.4-mini. Differences are GPT-5.4-mini minus the comparison method, in percentage points of the metric designated for each task. \(p\)-values are Holm-adjusted within two families: the seven baseline comparisons and the two backbone comparisons.}
\label{tab:pairwise-comparisons}
\end{table*}

\begin{figure*}[!b]
\centering
\includegraphics[width=\textwidth]{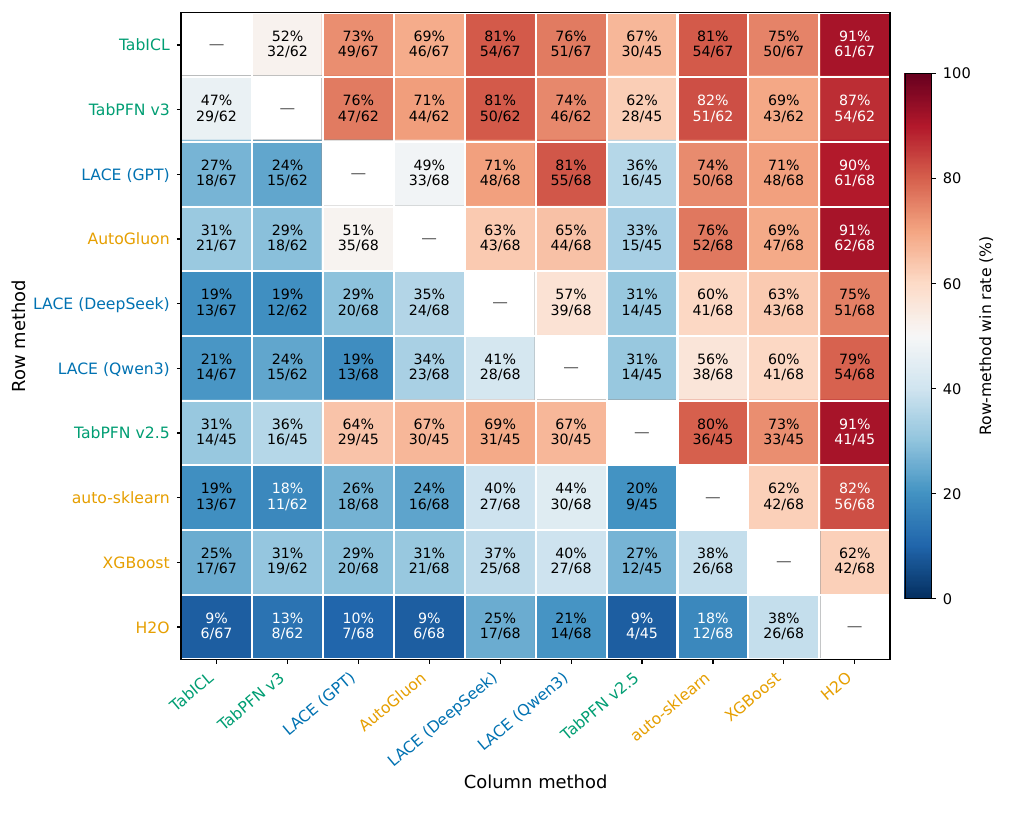}
\caption{Pairwise task-level win rates. Each cell gives the share of comparable tasks on which the row method achieves a higher score than the column method, with the underlying counts below it; a task is comparable when both methods return a valid result, and denominators therefore vary with coverage. Exact ties remain in the denominator, so complementary cells sum to 100\% only for pairs without ties; ties occur in ten pairs and never exceed two tasks. Methods are ordered by failure-aware mean rank, and tick-label colour groups them by family as in the main paper's method-comparison figure. These are raw win counts without significance testing; the tested comparisons involving \sysname{} (GPT-5.4-mini) are in Table~\ref{tab:pairwise-comparisons}.}
\label{fig:pairwise-heatmap}
\end{figure*}

Table~\ref{tab:pairwise-comparisons} contains the comparisons used for the paper's statistical claims. Figure~\ref{fig:pairwise-heatmap} additionally reports the task-level win rate for every ordered method pair, including baseline-to-baseline comparisons. Ordering the methods by failure-aware mean rank yields an almost monotone gradient, so the aggregate ranking is not an artifact of averaging over heterogeneous pairs, and the two departures from it are informative. \sysname{} with GPT-5.4-mini ranks marginally ahead of AutoGluon on both aggregate measures yet wins only 33 of their 68 shared tasks, consistent with the paired test that detects no difference between them. TabPFN~v2.5 wins the majority of its head-to-head comparisons against every full-coverage method, including 67\% against \sysname{} with Qwen3, while ranking behind them on the full benchmark: its comparisons are restricted to the 45 tasks it can run, which excludes precisely the tasks that its input limits rule out. The two summaries answer different questions: the pairwise view describes performance on shared supported tasks, whereas the failure-aware ranking also accounts for incomplete benchmark coverage.

\section{Performance by Task Size}
\label{sec:size-appendix}
\begin{figure*}[!t]
\centering
\includegraphics[width=\textwidth]{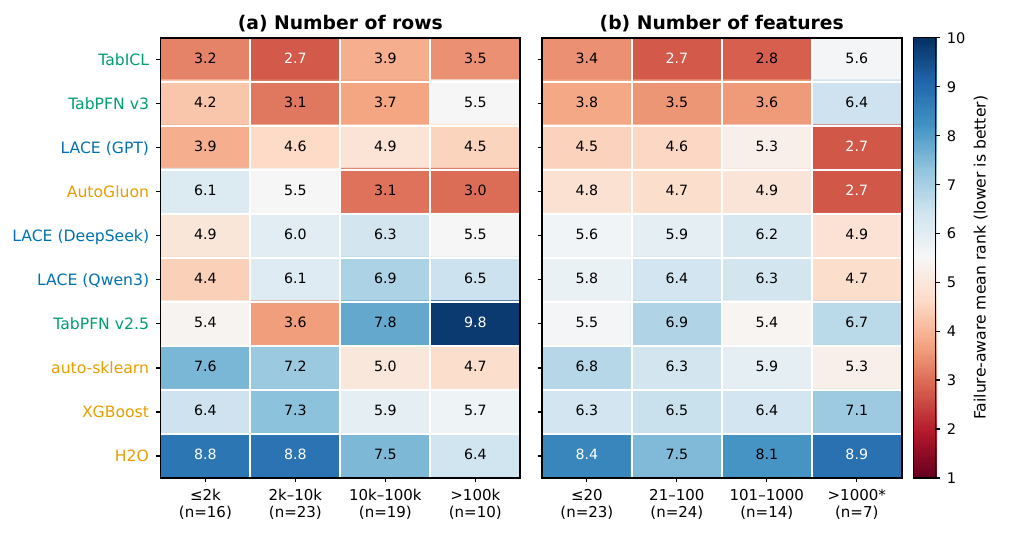}
\caption{Failure-aware mean task-level rank by (a)~number of rows and (b)~number of features. Lower is better; each column averages the within-task ranks of all ten methods over the tasks in that bin, with missing methods penalized as in the method-comparison figure of the main paper. Column labels give the number of tasks per bin; an asterisk marks bins with fewer than ten tasks, whose values should be read with caution. Row-label colour
groups methods by family.}
\label{fig:size-heatmap}
\end{figure*}

Figure~\ref{fig:size-heatmap} decomposes the failure-aware comparison of panel~(b) of the main paper's method-comparison figure by task size. Within each size bin, all ten methods are ranked on every task under the same convention as the main analysis: a method without a valid result receives the tied average of the remaining worst ranks, so incomplete coverage lowers a method's mean rank in exactly the bins containing the tasks it cannot run. The bins follow the row and feature counts of the task set (Appendix~\ref{sec:openml-task-appendix}); bin sizes are shown in the column labels, and the decomposition is descriptive, with no per-bin hypothesis tests.

\FloatBarrier
\section{Zero-Shot Validity and Improvement Details}
\label{sec:zero-shot-details}

For each \sysname{} repetition, the \emph{first-shot candidate} is the first pipeline evaluated from the initial population, before any execution feedback or evolutionary mutation is provided to the language model. The \emph{best initial candidate} is the highest-scoring valid pipeline among the four members of that population, and the \emph{final solution} is the highest-fitness pipeline returned at the end of the run. A candidate is valid if it completes evaluation and returns predictions with the required shape and type.

All three comparison points carry an inner-validation fitness. We report two complementary analyses: improvement on the inner-validation objective optimized during search, covering all three comparison points, and improvement on held-out official-test accuracy, obtained by independently refitting and re-evaluating the valid first programs after search (paragraph on the held-out replay below).

Survivor selection is elitist on this objective, so the final fitness is by construction at least as high as the fitness of either initial comparison point within the same repetition. Signed differences are consequently non-negative, and a signed-rank test would test a property of the selection rule rather than whether search improved its selected objective. For the inner-validation comparisons we therefore report medians, interquartile ranges, and percentile-bootstrap confidence intervals rather than hypothesis tests. This constraint does not apply to the held-out comparison below: official-test differences are not fixed in sign by the selection rule and can be negative, so signed-rank tests are meaningful there.

Differences are first computed within each repetition and then median-aggregated within each backbone--task pair. A pair is retained when at least three repetitions contribute to the corresponding comparison. The primary failure-aware analysis assigns an observed first-shot candidate with non-finite fitness a score of \(0\). This is a fixed failure penalty rather than an observed predictive accuracy. All first-candidate records are present, so no missing record is imputed. We additionally report a valid-only sensitivity analysis and a comparison against the best valid member of the four-candidate initial population. Over the 67 accuracy-scored tasks, the failure-aware comparison contains all \(201\) possible backbone--task units, the valid-only comparison contains \(132\), and the best-initial comparison contains \(190\).

Figure~\ref{fig:zero-shot-improvement-appendix} shows the failure-aware distribution.

% \begin{figure*}[!t]
% \centering
% \includegraphics[width=0.86\textwidth]
% {Figures/zero-shot/paired_final_vs_zero_shot_failure_aware.pdf}
% \caption{Failure-aware improvement in inner-validation fitness from the first-shot candidate to the returned \sysname{} pipeline over the 67 accuracy-scored tasks. Each point is one backbone--task unit; differences are computed within repetition and then median-aggregated over repetitions. An observed first-shot program with non-finite fitness is assigned a score of \(0\) before the difference is computed. This is a fixed failure penalty rather than an observed predictive accuracy. Diamonds denote medians, crosses denote means, and thick horizontal lines show percentile-bootstrap 95\% confidence intervals for the median. Task 360114 is scored by ROC--AUC and is excluded from this figure.}
% \label{fig:zero-shot-improvement-appendix}
% \end{figure*}
\begin{strip}
\begin{minipage}{\textwidth}
\centering

\includegraphics[width=0.86\textwidth]
{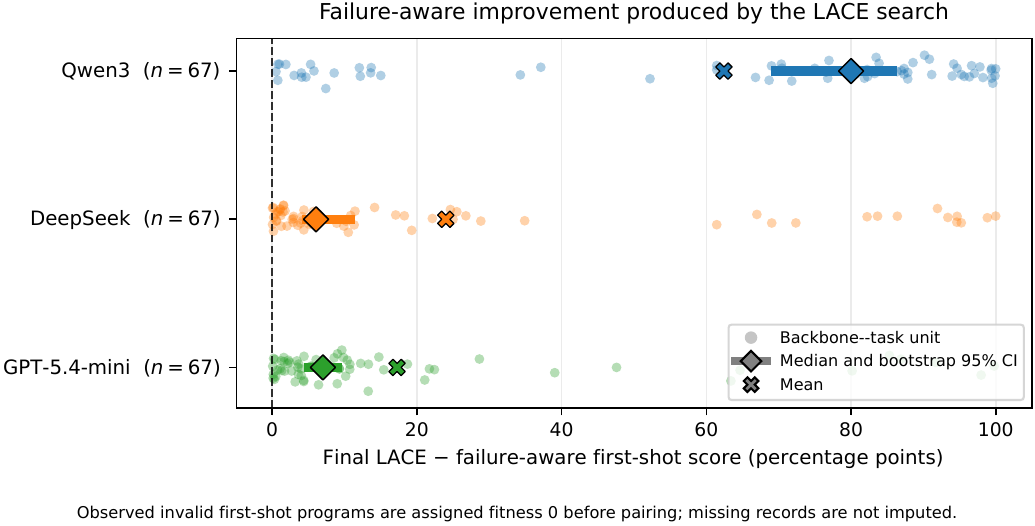}

\captionof{figure}{Failure-aware improvement in inner-validation fitness from the first-shot candidate to the returned \sysname{} pipeline over the 67 accuracy-scored tasks. Each point is one backbone--task unit; differences are computed within repetition and then median-aggregated over repetitions. An observed first-shot program with non-finite fitness is assigned a score of \(0\) before the difference is computed. This is a fixed failure penalty rather than an observed predictive accuracy. Diamonds denote medians, crosses denote means, and thick horizontal lines show percentile-bootstrap 95\% confidence intervals for the median. Task 360114 is scored by ROC--AUC and is excluded from this figure.}
\label{fig:zero-shot-improvement-appendix}

\end{minipage}
\end{strip}

\begin{table*}[!t]
\centering
\small
\begin{tabular}{lrrrrr}
\toprule
Backbone &
Planned runs &
First-shot valid &
Final valid &
Invalid recovered &
Recovery rate \\
\midrule
Qwen3
    & 340
    & \(128/340\) (\(37.6\%\))
    & \(340/340\) (\(100.0\%\))
    & \(212/212\)
    & \(100\%\) \\
DeepSeek
    & 340
    & \(226/340\) (\(66.5\%\))
    & \(340/340\) (\(100.0\%\))
    & \(114/114\)
    & \(100\%\) \\
GPT-5.4-mini
    & 340
    & \(271/340\) (\(79.7\%\))
    & \(340/340\) (\(100.0\%\))
    & \(69/69\)
    & \(100\%\) \\
\bottomrule
\end{tabular}
\caption{Validity of the first zero-shot candidate and the final selected pipeline. A recovered invalid first shot is a run whose first candidate has non-finite fitness but whose final selected pipeline is valid. A first-shot record is present for every planned run.}
\label{tab:zero-shot-validity}
\end{table*}

\begin{table}[!t]
\centering
\small
\begin{tabular}{lrrr}
\toprule
Backbone & \(n\) & Median [IQR] & 95\% CI \\
\midrule

\multicolumn{4}{l}{\emph{Final \( - \) first shot, invalid assigned \(0\)}} \\
Qwen3
    & 67 & 80.00 [14.35, 90.98] & [69.00, 86.38] \\
DeepSeek
    & 67 & 6.07 [1.68, 26.14] & [4.15, 11.44] \\
GPT-5.4-mini
    & 67 & 7.02 [2.55, 14.02] & [4.36, 9.63] \\
Overall
    & 201 & 11.33 [3.54, 76.94] & [8.99, 18.27] \\

\midrule
\multicolumn{4}{l}{\emph{Final \( - \) first shot, valid first shots only}} \\
Qwen3
    & 19 & 2.67 [0.83, 8.15] & [0.87, 7.92] \\
DeepSeek
    & 52 & 2.74 [0.84, 7.85] & [1.39, 3.58] \\
GPT-5.4-mini
    & 61 & 4.96 [1.71, 9.72] & [3.28, 6.58] \\
Overall
    & 132 & 3.46 [1.10, 8.59] & [2.70, 4.40] \\

\midrule
\multicolumn{4}{l}{\emph{Final \( - \) best valid initial candidate}} \\
Qwen3
    & 59 & 1.51 [0.64, 3.51] & [0.92, 1.86] \\
DeepSeek
    & 65 & 1.45 [0.52, 3.09] & [1.02, 2.50] \\
GPT-5.4-mini
    & 66 & 1.92 [0.49, 4.30] & [0.89, 2.72] \\
Overall
    & 190 & 1.58 [0.57, 3.73] & [1.16, 2.04] \\

\bottomrule
\end{tabular}
\caption{Task-level improvement in inner-validation fitness, in accuracy percentage points, over the 67 accuracy-scored tasks. Differences are computed within repetition and then median-aggregated within each backbone--task unit; \(n\) is the number of units with at least three paired repetitions. The first block assigns an observed invalid first shot a failure-aware score of \(0\); the second restricts the comparison to valid first shots; the third compares the final pipeline with the best valid member of the initial population. Confidence intervals are percentile-bootstrap intervals for the median over backbone--task units.}
\label{tab:zero-shot-performance}
\end{table}

The failure-aware estimate combines recovery from invalid first programs with predictive improvement among valid programs. It is especially large for Qwen3 because \(212\) of its \(340\) first-shot programs are invalid and therefore receive the score \(0\). The valid-only analysis separates this recovery effect from predictive improvement among executable programs; the held-out replay below confirms that this improvement is not confined to the search objective. The \(1.58\)-point gain over the best valid initial candidate further shows that the result is not explained only by selecting the best of four independent initial generations.

Task 360114 is scored by ROC--AUC and is excluded from all accuracy-pooled values above. Its differences are analyzed separately and are never combined with accuracy percentage points. 

\paragraph{Held-out replay of first programs.}
To test whether search improves generalization rather than only its selection objective, every valid literal first program (candidate index 1, generation 0, constructor defaults) was independently refit on the official training folds and scored once on the corresponding held-out folds, under the same protocol applied to the returned pipeline. Of the 625 runs with a valid first program, 618 returned a finite official aggregate (617 complete, one 8-of-10-fold partial, seven evaluation timeouts recorded as missing rather than zero); 611 finite pairs remain after separating task 360114 which is declared AUC. Runs whose first program was invalid are excluded, and no artificial worst score is assigned.
Table~\ref{tab:zero-shot-official} reports the paired differences. At run level 586 of 611 pairs improved, 6 tied, and 19 were worse. Restricting to first programs completing all ten folds changes the pooled median by less than \(0.01\) points.

% \begin{table}[!t]
% \centering
% \small
% \setlength{\tabcolsep}{4pt}
% \caption{Held-out official-test improvement from the valid first program to the returned pipeline, in accuracy percentage points. Differences are computed within repetition and median-aggregated per backbone--task unit, as in the main paper's zero-shot section. Confidence intervals are percentile bootstrap (10{,}000 resamples) for the median; \(p\)-values are two-sided Wilcoxon, Holm-adjusted within the three backbones.}
% \label{tab:zero-shot-official}
% \begin{tabular}{@{}lrrrr@{}}
% \toprule
% Backbone & Units & Median [IQR] & 95\% CI & Holm \(p\) \\
% \midrule
% DeepSeek     & 65/67  & 2.51 [0.70, 8.39] & [1.00, 3.63] & \(4.9\times10^{-12}\) \\
% GPT-5.4-mini & 66/67  & 4.53 [1.55, 9.60] & [2.74, 5.75] & \(4.9\times10^{-12}\) \\
% Qwen3        & 62/67  & 2.87 [0.72, 7.94] & [1.85, 4.51] & \(1.0\times10^{-11}\) \\
% \midrule
% Overall      & 193/201 & 2.99 [0.71, 8.73] & [2.42, 4.17] & --- \\
% \bottomrule
% \end{tabular}
% \end{table}

\par\medskip
\noindent
\begin{minipage}{\columnwidth}
\centering
\small
\setlength{\tabcolsep}{4pt}

\captionof{table}{Held-out official-test improvement from the valid first
program to the returned pipeline, in accuracy percentage points. Differences
are computed within repetition and median-aggregated per backbone--task unit,
as in the main paper's zero-shot section. Confidence intervals are percentile
bootstrap (10{,}000 resamples) for the median; \(p\)-values are two-sided
Wilcoxon, Holm-adjusted within the three backbones.}
\label{tab:zero-shot-official}

\begin{tabular}{@{}lrrrr@{}}
\toprule
Backbone & Units & Median [IQR] & 95\% CI & Holm \(p\) \\
\midrule
DeepSeek
  & 65/67 & 2.51 [0.70, 8.39] & [1.00, 3.63]
  & \(4.9\times10^{-12}\) \\
GPT-5.4-mini
  & 66/67 & 4.53 [1.55, 9.60] & [2.74, 5.75]
  & \(4.9\times10^{-12}\) \\
Qwen3
  & 62/67 & 2.87 [0.72, 7.94] & [1.85, 4.51]
  & \(1.0\times10^{-11}\) \\
\midrule
Overall
  & 193/201 & 2.99 [0.71, 8.73] & [2.42, 4.17] & --- \\
\bottomrule
\end{tabular}

\end{minipage}
\par\medskip
\paragraph{Inner objective versus held-out gain.}
Over the identical 611 finite pairs, the median inner-validation gain is \(3.21\) points and the median held-out gain \(2.95\) points; 609 of 611 pairs improved on the inner objective and 586 on held-out accuracy. The held-out gain is therefore slightly smaller than the objective optimized during search, consistent with mild selection overfitting, but remains positive on the large majority of runs.

\begin{figure*}[!t]
\centering

\includegraphics[width=0.80\textwidth]
{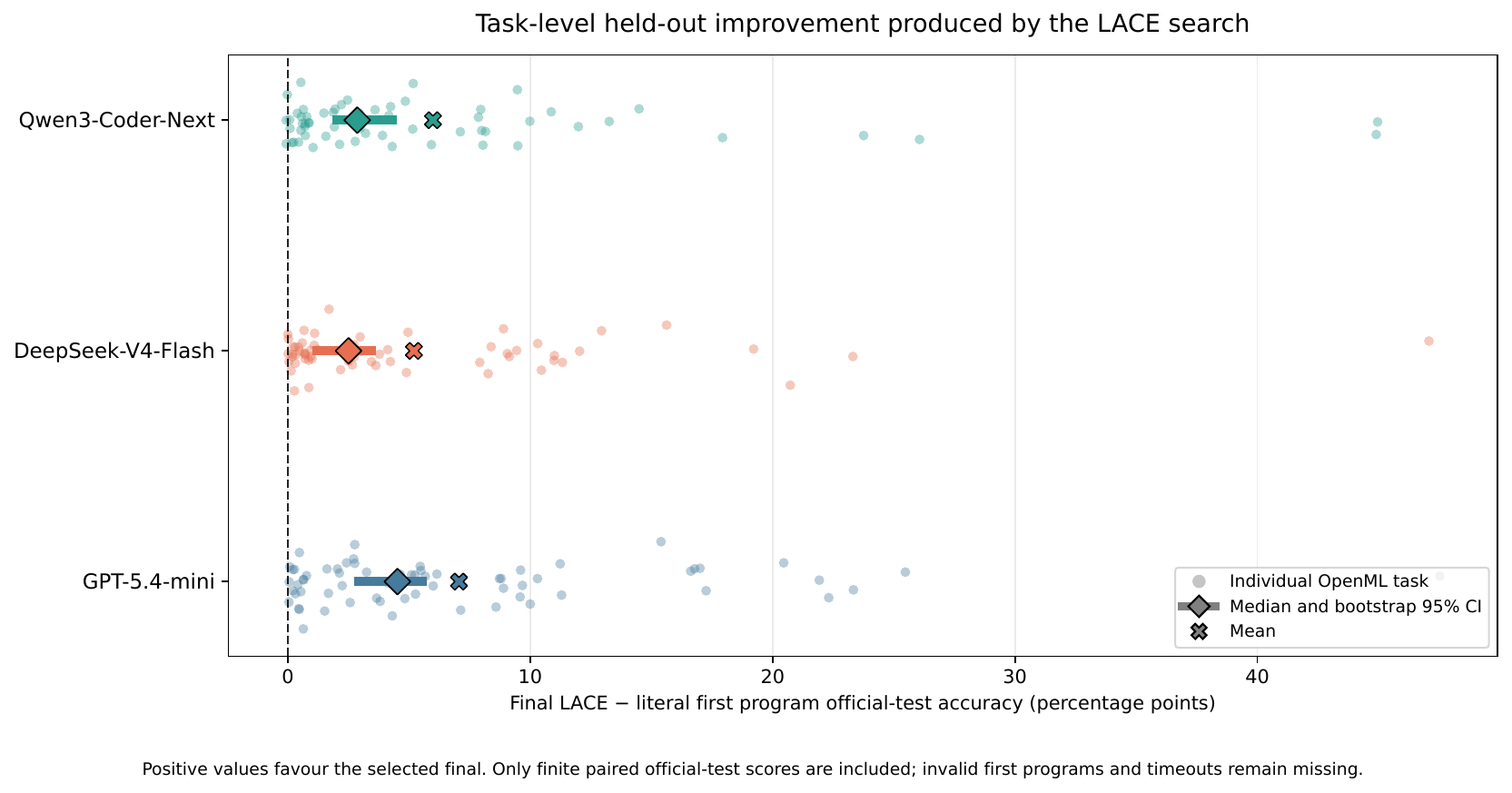}

\caption{Held-out official-test improvement from the valid literal first
program to the returned \sysname{} pipeline over the 67 accuracy-scored tasks.
Each point is one backbone--task unit with at least three finite paired official
replays; differences are computed within repetition and then median-aggregated
within each unit. Invalid first programs and official-evaluation timeouts remain
missing, and no artificial score of zero is assigned. Consequently, each
backbone is evaluated on a validity-conditioned subset that differs in size and
composition: 62/67 task units for Qwen3, 65/67 for DeepSeek, and 66/67 for
GPT-5.4-mini. The distributions therefore quantify within-backbone improvement
from first to final and should not be interpreted as a like-for-like comparison
between backbones. The axis extends below zero; three task units have negative
differences, demonstrating that held-out improvement is not sign-constrained by
elitist selection. Diamonds denote medians, crosses denote means, and thick
horizontal lines show percentile-bootstrap 95\% confidence intervals.
Task 360114 is excluded because its official evaluation recorded the AUC.}
\label{fig:zero-shot-official-improvement}

\end{figure*}

At the backbone--task level, the returned pipeline improves on 190 of the 193 retained units; the remaining three have negative differences. All 66 retained GPT-5.4-mini units improve. Across the 193 units, the two-sided Wilcoxon test gives \(p<10^{-32}\).
Figure~\ref{fig:zero-shot-official-improvement} visualizes the
task-level held-out distributions underlying
Table~\ref{tab:zero-shot-official}.

% \begin{figure*}[!b]
% \centering
% \includegraphics[width=0.86\textwidth]
% {Figures/zero-shot/official_test_improvement_all_successful_fold_mean.pdf}
% \caption{Held-out official-test improvement from the valid literal
% first program to the returned \sysname{} pipeline over the 67
% accuracy-scored tasks. Each point is one backbone--task unit with at
% least three finite paired official replays; differences are computed
% within repetition and then median-aggregated within each unit.
% Invalid first programs and official-evaluation timeouts remain
% missing, and no artificial score of zero is assigned. Consequently,
% each backbone is evaluated on a validity-conditioned subset that
% differs in size and composition: 62/67 task units for Qwen3, 65/67 for DeepSeek, and 66/67 for GPT-5.4-mini. The distributions therefore quantify within-backbone improvement from first to final and should not be interpreted as a like-for-like comparison between backbones. The axis extends below zero; three task units have negative differences, demonstrating that held-out improvement is not sign-constrained by elitist selection. Diamonds denote medians,
% crosses denote means, and thick horizontal lines show percentile-bootstrap 95\% confidence intervals. Task 360114 is
% excluded because its official evaluation recorded the AUC.}
% \label{fig:zero-shot-official-improvement}
% \end{figure*}

\FloatBarrier

\section{Search Dynamics Details}
\label{sec:dynamics-appendix}

The dynamics analysis uses the \(1{,}020\) complete runs: 340 each for deepseek-v4-flash, GPT-5.4-mini, and qwen3-coder-next. Thirteen runs (\(1.3\%\)) never improved on their initial population and are excluded from normalized progress statistics. A further 89 runs, comprising 21 DeepSeek, 4 GPT-5.4-mini, and 64 Qwen3 runs, had four invalid initial candidates and therefore have no generation-0 baseline. These runs remain in all non-normalized statistics. Twelve deepseek-v4-flash runs used an earlier search configuration; excluding them in a sensitivity analysis leaves the reported conclusions materially unchanged.

Twelve of the \(98{,}062\) declared parent pointers, all from one DeepSeek run, refer to candidates absent from the corresponding log and are excluded from parent--offspring denominators. The final pipeline of the affected run remains traceable.
 
\paragraph{Executability by generation.}
Table~\ref{tab:invalid-appendix} gives the run-level median invalid share per generation. The overall series returns to its generation-0 level by the end of a run; under qwen3-coder-next the median declines from 75\% at initialization to 58\% in the final generation while remaining the highest of the three backbones throughout.
 
% \begin{table*}[!t]
% \centering
% \small
% \setlength{\tabcolsep}{5pt}
% \caption{Run-level median invalid-candidate share per generation, in percent.}
% \label{tab:invalid-appendix}
% \begin{tabular}{@{}lrrrrrrrrr@{}}
% \toprule
% Backbone & 0 & 1 & 2 & 3 & 4 & 5 & 6 & 7 & 8 \\
% \midrule
% DeepSeek      & 25.0 & 25.0 & 16.7 & 16.7 & 16.7 & 25.0 & 25.0 & 25.0 & 25.0 \\
% GPT-5.4-mini  &  0.0 &  8.3 &  8.3 &  8.3 &  8.3 &  8.3 &  8.3 &  8.3 &  8.3 \\
% Qwen3         & 75.0 & 58.3 & 50.0 & 50.0 & 50.0 & 50.0 & 50.0 & 50.0 & 58.3 \\
% \midrule
% Overall       & 25.0 & 33.3 & 25.0 & 25.0 & 25.0 & 25.0 & 25.0 & 25.0 & 25.0 \\
% \bottomrule
% \end{tabular}
% \end{table*}

\begin{table*}[t]
\centering
\small
\setlength{\tabcolsep}{5pt}

\caption{Run-level median invalid-candidate share per generation, in percent.}
\label{tab:invalid-appendix}

\begin{tabular}{@{}lrrrrrrrrr@{}}
\toprule
Backbone & 0 & 1 & 2 & 3 & 4 & 5 & 6 & 7 & 8 \\
\midrule
DeepSeek
  & 25.0 & 25.0 & 16.7 & 16.7 & 16.7 & 25.0 & 25.0 & 25.0 & 25.0 \\
GPT-5.4-mini
  & 0.0 & 8.3 & 8.3 & 8.3 & 8.3 & 8.3 & 8.3 & 8.3 & 8.3 \\
Qwen3
  & 75.0 & 58.3 & 50.0 & 50.0 & 50.0 & 50.0 & 50.0 & 50.0 & 58.3 \\
\midrule
Overall
  & 25.0 & 33.3 & 25.0 & 25.0 & 25.0 & 25.0 & 25.0 & 25.0 & 25.0 \\
\bottomrule
\end{tabular}

\end{table*}
 
\paragraph{Attainment and progress.}
Table~\ref{tab:attainment-appendix} reports the fraction of runs at final quality by generations 1, 2, and 4 under the exact criterion (final best reached within \(10^{-12}\)) and the practical criterion (at least 99\% of the run's total improvement). The median share of total improvement reached by candidate 50 was 87.6\% for deepseek-v4-flash, 93.7\% for GPT-5.4-mini, and 90.7\% for qwen3-coder-next.
 
% \begin{table}[!ht]
% \centering
% \small
% \setlength{\tabcolsep}{4pt}
% \caption{Runs at final quality by generation, exact / practical criterion, in percent. Practical denominators exclude runs without a defined normalized trajectory.}
% \label{tab:attainment-appendix}
% \begin{tabular}{@{}lrrr@{}}
% \toprule
% Backbone & Gen.\ 1 & Gen.\ 2 & Gen.\ 4 \\
% \midrule
% DeepSeek      & 2.1 / 1.6 & 2.9 / 2.9   & 19.4 / 21.7 \\
% GPT-5.4-mini  & 2.9 / 4.2 & 7.7 / 9.3   & 19.5 / 24.9 \\
% Qwen3         & 5.6 / 7.0 & 11.5 / 13.7 & 25.6 / 30.4 \\
% \midrule
% Overall       & 3.5 / 4.1 & 7.4 / 8.4   & 21.5 / 25.4 \\
% \bottomrule
% \end{tabular}
% \end{table}

\par\medskip
\noindent
\begin{minipage}{\columnwidth}
\centering
\small
\setlength{\tabcolsep}{4pt}

\captionof{table}{Runs at final quality by generation, exact / practical
criterion, in percent. Practical denominators exclude runs without a defined
normalized trajectory.}
\label{tab:attainment-appendix}

\begin{tabular}{@{}lrrr@{}}
\toprule
Backbone & Gen.\ 1 & Gen.\ 2 & Gen.\ 4 \\
\midrule
DeepSeek
  & 2.1 / 1.6 & 2.9 / 2.9 & 19.4 / 21.7 \\
GPT-5.4-mini
  & 2.9 / 4.2 & 7.7 / 9.3 & 19.5 / 24.9 \\
Qwen3
  & 5.6 / 7.0 & 11.5 / 13.7 & 25.6 / 30.4 \\
\midrule
Overall
  & 3.5 / 4.1 & 7.4 / 8.4 & 21.5 / 25.4 \\
\bottomrule
\end{tabular}

\end{minipage}
\par\medskip
 
\paragraph{Lineage of finals.}
Table~\ref{tab:lineage-appendix} reports, per backbone, the generation in which the final candidate was created and its ancestry depth, the number of mutation edges to its nearest generation-0 ancestor.
 
% \begin{table}[!ht]
% \centering
% \small
% \setlength{\tabcolsep}{4pt}
% \caption{Lineage of the final pipelines: creation generation, share created in generation 0 (unmodified initial candidates), share created in the final two generations, and ancestry depth in mutation edges.}
% \label{tab:lineage-appendix}
% \begin{tabular}{@{}lrrrr@{}}
% \toprule
% Backbone & Creation gen. & In g0 & Final two & Depth \\
% \midrule
% DeepSeek      & 7 [5, 8]    & 1.5\% & 56.8\% & 4 [3, 5] \\
% GPT-5.4-mini  & 7 [5, 8]    & 0.6\% & 55.8\% & 4 [3, 5] \\
% Qwen3         & 6 [4.75, 8] & 1.8\% & 45.3\% & 3 [2, 4] \\
% \midrule
% Overall       & 7 [5, 8]    & 1.3\% & 52.6\% & 4 [3, 5] \\
% \bottomrule
% \end{tabular}
% \end{table}
\par\medskip
\noindent
\begin{minipage}{\columnwidth}
\centering
\small
\setlength{\tabcolsep}{4pt}

\captionof{table}{Lineage of the final pipelines: creation generation,
share created in generation 0 (unmodified initial candidates), share created
in the final two generations, and ancestry depth in mutation edges.}
\label{tab:lineage-appendix}

\begin{tabular}{@{}lrrrr@{}}
\toprule
Backbone & Creation gen. & In g0 & Final two & Depth \\
\midrule
DeepSeek
  & 7 [5, 8] & 1.5\% & 56.8\% & 4 [3, 5] \\
GPT-5.4-mini
  & 7 [5, 8] & 0.6\% & 55.8\% & 4 [3, 5] \\
Qwen3
  & 6 [4.75, 8] & 1.8\% & 45.3\% & 3 [2, 4] \\
\midrule
Overall
  & 7 [5, 8] & 1.3\% & 52.6\% & 4 [3, 5] \\
\bottomrule
\end{tabular}

\end{minipage}
\par\medskip
 
\paragraph{Parent--offspring outcomes.}
Table~\ref{tab:pairs-appendix} reports pooled pair counts and run-level median rates. Fitness differences are reported only within a metric: on the accuracy-scored runs, the median difference over all valid pairs was \(-0.2\) percentage points and the median improving step \(+0.23\); the 15 runs of the single AUC-scored task (360114) show the same pattern (\(-0.2\) and \(+0.10\) AUC points) and are never pooled with the accuracy runs. Mutation-instruction labels were logged only for crossover offspring, so a per-instruction breakdown is unavailable.
 
% \begin{table*}[!t]
% \centering
% \small
% \caption{Parent--offspring outcomes among valid parent--offspring pairs. Improving and degrading columns give pooled percentages, with run-level medians in parentheses; the final two columns give pooled offspring-validity rates conditional on parent validity.}
% \label{tab:pairs-appendix}
% \begin{tabular}{@{}lrrrcc@{}}
% \toprule
% Backbone & Pairs & Improving (\%) & Degrading (\%)
% & $P(\text{valid}\mid\text{inv.})$
% & $P(\text{valid}\mid\text{val.})$ \\
% \midrule
% DeepSeek      & 23{,}993 & 21.0 (21.0) & 71.3 (72.3) & 48.7 & 76.6 \\
% GPT-5.4-mini  & 28{,}150 & 23.0 (22.6) & 73.2 (73.8) & 60.1 & 88.4 \\
% Qwen3         & 14{,}557 & 22.1 (21.6) & 73.7 (74.5) & 23.9 & 49.9 \\
% \midrule
% Overall       & 66{,}700 & 22.1 (21.9) & 72.6 (73.6) & 34.9 & 72.2 \\
% \bottomrule
% \end{tabular}
% \end{table*}

\begin{table*}[t]
\centering
\small

\caption{Parent--offspring outcomes among valid parent--offspring
pairs. Improving and degrading columns give pooled percentages, with run-level
medians in parentheses; the final two columns give pooled offspring-validity
rates conditional on parent validity.}
\label{tab:pairs-appendix}

\begin{tabular}{@{}lrrrcc@{}}
\toprule
Backbone
& Pairs
& Improving (\%)
& Degrading (\%)
& $P(\text{valid}\mid\text{inv.})$
& $P(\text{valid}\mid\text{val.})$ \\
\midrule
DeepSeek
  & 23{,}993 & 21.0 (21.0) & 71.3 (72.3) & 48.7 & 76.6 \\
GPT-5.4-mini
  & 28{,}150 & 23.0 (22.6) & 73.2 (73.8) & 60.1 & 88.4 \\
Qwen3
  & 14{,}557 & 22.1 (21.6) & 73.7 (74.5) & 23.9 & 49.9 \\
\midrule
Overall
  & 66{,}700 & 22.1 (21.9) & 72.6 (73.6) & 34.9 & 72.2 \\
\bottomrule
\end{tabular}

\end{table*}

\paragraph{Representative run.}
Total inner-validation improvement varies widely across runs. Over the 335 GPT-5.4-mini runs with a valid initial candidate its median is 1.7 accuracy percentage points (interquartile range 0.5--5.3), rising to 10.8 points at the ninetieth percentile and 69.5 points at the maximum; the largest gains occur on tasks whose initial pipelines fall far short of competent accuracy. The run traced below is therefore typical rather than favourable. It was selected by a fixed rule: the task whose median total improvement across repetitions lies closest to the backbone-wide median, then the repetition at the task median, giving task 168757, repetition 4 (best initial-population fitness 0.747, final fitness 0.764). Table~\ref{tab:lineage-run-appendix} traces the ancestry of the final pipeline: the run began from a voting ensemble of a forest and a linear model behind imputation, scaling, and PCA, moved to a single gradient-boosted model behind a per-feature-type \texttt{ColumnTransformer}, reintroduced a linear voting partner, and settled on a three-family voting ensemble of a forest, gradient boosting, and a linear model. The best initial candidate need not be an ancestor of the final, so the ancestry begins below the generation-0 population best. The first and final programs of this run are included in the accompanying Zenodo artifact.

\par\medskip
\noindent
\begin{minipage}{\columnwidth}
\centering
\small
\setlength{\tabcolsep}{4pt}

\captionof{table}{Parent ancestry of the final pipeline in the representative
GPT-5.4-mini run (task 168757, repetition 4). Fitness is mean
inner-validation accuracy.}
\label{tab:lineage-run-appendix}

\begin{tabular}{@{}rr>{\raggedright\arraybackslash}p{4.2cm}r@{}}
\toprule
Cand. & Gen. & Structure & Fitness \\
\midrule
4
  & 0
  & impute, scale, PCA; voting (forest + linear)
  & 0.724 \\
14
  & 1
  & per-type encoding; single gradient boosting
  & 0.750 \\
21
  & 2
  & per-type encoding; voting (boosting + linear)
  & 0.754 \\
40
  & 3
  & per-type encoding; voting (forest + boosting + linear)
  & 0.764 \\
\bottomrule
\end{tabular}

\end{minipage}
\par\medskip

\paragraph{Final pipeline interface.}
Listing~\ref{lst:representative} shows the interface of the final pipeline
selected in this representative run.

\par\medskip
\noindent
\begin{minipage}{\columnwidth}
\begin{lstlisting}[
language=Python,
basicstyle=\ttfamily\footnotesize,
frame=single,
columns=fullflexible,
keepspaces=true,
breaklines=true,
breakatwhitespace=true,
aboveskip=3pt,
belowskip=3pt,
caption={Interface of the final pipeline in the representative run
(GPT-5.4-mini, task 168757, repetition 4): a soft-voting ensemble of
logistic regression, extra trees, and histogram gradient boosting behind a
per-feature-type \texttt{ColumnTransformer}. The complete source is available
in the accompanying Zenodo artifact.},
label={lst:representative}
]
class TabularEnsemblePipeline(BaseEstimator, ClassifierMixin):
    def __init__(self, X, y, **hp):
        # fit once, no CV
        self.pre_ = self._build_preprocessor()
        Xt = self.pre_.fit_transform(self._to_df(X))

        self.model = VotingClassifier(
            voting="soft",
            weights=(1.0, 1.0, 1.15),
            estimators=[
                ("lr", LogisticRegression(...)),
                ("et", ExtraTreesClassifier(...)),
                ("hgb", HistGradientBoostingClassifier(...)),
            ],
        )
        self.model.fit(Xt, y)

    def __call__(self, X):
        # transform, then predict
        Xt = self.pre_.transform(self._to_df(X))
        return self.model.predict(Xt)
\end{lstlisting}
\end{minipage}
\par\medskip
\FloatBarrier
\section{Prompts}
\label{sec:prompts}
This appendix reproduces the prompts used in all main experiments. The static prompt text is identical across the three backbones; the only difference is a trailing newline added by the adapter for the locally served
model. Leading indentation from the source templates has been removed and long lines wrapped to the column width; placeholders appear in braces.

\paragraph{Initialization prompt.}
Listing~\ref{lst:init-prompt} generates each of the $\mu$ initial candidates. It carries no parent program and no evaluation feedback, so the initial population reflects the model's prior over pipelines for the described task. The example class in the initialization prompt makes the fit-on-construction, predict-on-call interface explicit and is the complete example referred to in the main paper. The task summary is constructed from the dataset alone: the task type, the number of samples (rounded to the nearest thousand for datasets of at least 1{,}000 instances), the number of features, and how many of those features are boolean, integer, and real-valued. For task 359955 the resulting summary reads \texttt{classification task with 748 samples and \textasciitilde 4 features: 0 boolean, 3 integer, 1 real}. The dataset name, OpenML task identifier, feature names, target semantics, class distribution, and evaluation metric are never inserted.

\begin{lstlisting}[caption={Initialization prompt. The role text and the task,
example, and output-format blocks are concatenated in this order and sent as a
single user message.}, label={lst:init-prompt}]
You are an excellent Python programmer.

You can use the following Python packages: scikit-learn, numpy, scipy, pandas.
Design an ML pipeline for a {task_type} task with {samples_desc} and
{feats_desc}.
Write a single Python class:
- __init__(self, X, y, **hyperparameters)  -> fit exactly once (no CV here)
- __call__(self, X)
IMPORTANT CONSTRAINTS (must follow):
- Do NOT use any internal HPO or CV search (no GridSearchCV,
  RandomizedSearchCV, BayesSearchCV, Optuna, Hyperopt, skopt, etc.).
- Do NOT implement your own tuning loops (no KFold/StratifiedKFold loops, no
  parameter sweeps).
- Pick ONE primary estimator OBJECT inside the class. This can be either (a) a
  single sklearn estimator, OR (b) a simple sklearn ensemble wrapper (e.g.,
  VotingClassifier or StackingClassifier) that combines up to 2-3 base
  estimators.
- Do NOT do estimator-switching in a param grid.
- If you use an ensemble wrapper, keep it small (2-3 models max), and keep
  preprocessing shared (apply preprocessing once, then feed the same
  transformed X to all base estimators).
- If you do preprocessing (e.g., StandardScaler, PCA), assign them to self.*
  and reuse them in __call__.
- Expose tunable hyperparameters via __init__ kwargs with sensible defaults;

Here is a minimal template (for reference only; you must output your own
improved class). An example code structure is as follows:
```python
import numpy as np
import sklearn
from sklearn.preprocessing import StandardScaler
from sklearn.linear_model import LogisticRegression

class MyPipeline:
    "Simple, single-estimator pipeline without internal HPO."
    def __init__(self, X, y, C=1.0):
        # keep references for use in __call__
        self.scaler = StandardScaler()
        Xs = self.scaler.fit_transform(X)
        # choose ONE estimator; expose its hyperparameters
        self.model = LogisticRegression(C=C, max_iter=1000, n_jobs=1,
                                        random_state=42)
        self.model.fit(Xs, y)

    def __call__(self, X):
        Xs = self.scaler.transform(X)
        return self.model.predict(Xs)
```

Give an excellent and novel ML pipeline to solve this task and also give it a
one-line description, describing the main idea. Give the response in the
format:
# Description: <short-description>
# Code:
```python
<code>
```
# Space:
```python
{
    # e.g. "C": (1e-3, 10.0), "max_depth": (3, 15), "alpha": (1e-4, 1.0)
}
```
\end{lstlisting}

\paragraph{Mutation prompt.}
Listing~\ref{lst:mut-prompt} generates each offspring. It repeats the role and task blocks of the initialization prompt, then adds the population summary, the selected parent with its description and code, the parent's evaluation
feedback and any error text, one mutation instruction drawn uniformly at random, and the same output-format block. Population members are summarized as one line each, giving the class name, the model's own one-line description, and
the inner-validation score; invalid members appear with a score of $-\infty$.

\begin{lstlisting}[caption={Mutation prompt template.},
label={lst:mut-prompt}]
{role_prompt}{task_prompt}
The current population of algorithms already evaluated (name, description,
score) is:
{population_summary}
The selected solution to update is:
{description}
With code:
```python
{solution}
```
Feedback:
{feedback}
{error_message}
{mutation_operator}
{output_format_prompt}
\end{lstlisting}

\paragraph{Mutation instructions.}
One of the following three instructions is drawn uniformly at random for each offspring and inserted as \texttt{\{mutation\_operator\}}.

\begin{lstlisting}[caption={The three mutation instructions.},
label={lst:mut-ops}]
Change the model family used (e.g., linear <-> tree ensemble <-> kernel
method), while keeping the same class interface and avoiding internal CV/HPO
loops.

Refine the preprocessing part of the pipeline: add/remove/adjust scaling,
imputation handling, or simple feature selection, but keep the estimator
family unchanged.

Add or remove ONE base model to a small ensemble (e.g., VotingClassifier or
StackingClassifier) while keeping everything simple and within constraints.
\end{lstlisting}

\paragraph{Instructed versus audited constraints.}
The prompt states the package set, the fit-once interface, the prohibition on internal search and tuning loops, the ensemble size limit, and the requirement to expose hyperparameters as constructor arguments. The audit described in the main paper is broader in three respects: it tests for nested AutoML frameworks, which the prompt excludes only by naming the available packages; it tests for hyperparameter-driven estimator switching in any form, whereas the prompt forbids it only within a parameter grid; and it tests for fitting at prediction time, which the prompt implies through ''fit exactly once'' without stating separately. The absence of the first two in every final pipeline is therefore a property of the generated programs rather than compliance with an explicit instruction.

\FloatBarrier
\section{Foundation-Model Ablation}
\label{sec:tabpfn-appendix}
The main experiments give \sysname{} and the tabular foundation models separate component sets, so that they are compared as distinct systems rather than one containing the other. This ablation asks what happens when a foundation model becomes available to the generator. We use tasks 2073, 146818, 168350, 359983, and 359990, chosen to span the size range of the benchmark, with five repetitions and a budget of 100 candidates per run. The \emph{default} condition reuses the corresponding main-experiment runs. The \emph{passive} condition additionally installs TabPFN without mentioning it in the prompt, while the \emph{explicit} condition announces it as an available estimator. All other settings are unchanged.

We conduct the ablation for two backbone and TabPFN-version pairs: deepseek-v4-flash with TabPFN~v2.5 and GPT-5.4-mini with TabPFN~v3. In the explicit condition, the prompt states that \texttt{TabPFNClassifier} is available, that it should be treated as a single estimator family rather than as an AutoML system, and gives its detected constructor signature. One explicit run on task 168350 in the DeepSeek experiment is missing, so that cell averages four repetitions.

Availability alone had no observable effect on the generated search space. None of the 5{,}000 candidates produced while TabPFN was installed but unannounced imported it, and none of the resulting 50 selected pipelines used it. Announcing the component changed generation immediately: \(74.5\%\) of candidates used TabPFN in the DeepSeek experiment and \(42.0\%\) in the GPT-5.4-mini experiment. In total, 39 of the 50 selected pipelines contained TabPFN, usually as part of an ensemble rather than as the sole estimator.

Table~\ref{tab:tabpfn-ablation} reports predictive performance. We do not report hypothesis tests: with five tasks, the smallest attainable two-sided signed-rank \(p\)-value is \(0.0625\).

\begin{table}[!ht]
\centering
\footnotesize
\setlength{\tabcolsep}{4pt}
\caption{Accuracy in percent, averaged over five repetitions, with standard deviation across repetitions in parentheses. \emph{Default} is the environment of the main experiments; \emph{passive} installs TabPFN without mentioning it to the generator; and \emph{explicit} announces it in the prompt. TabPFN~v2.5 returns no valid score on task 359990, whose size exceeds its input limit.}
\label{tab:tabpfn-ablation}
\begin{tabular}{@{}lrrrr@{}}
\toprule
Task & Default & Passive & Explicit & TabPFN \\
\midrule
\multicolumn{5}{l}{\emph{deepseek-v4-flash with TabPFN~v2.5}} \\
2073   & 62.28 (0.58) & 62.53 (0.61) & 62.61 (0.50) & 61.31 (0.24) \\
146818 & 87.36 (0.38) & 86.84 (0.50) & 87.48 (0.22) & 86.75 (0.22) \\
168350 & 90.68 (0.48) & 90.64 (0.56) & 92.27 (0.18) & 91.77 (0.13) \\
359983 & 87.43 (0.02) & 87.40 (0.09) & 87.40 (0.08) & 86.68 (0.04) \\
359990 & 94.73 (0.15) & 94.85 (0.20) & 94.68 (0.40) & --- \\
\midrule
\multicolumn{5}{l}{\emph{GPT-5.4-mini with TabPFN~v3}} \\
2073   & 62.21 (0.73) & 62.44 (0.48) & 63.01 (0.44) & 62.30 (0.28) \\
146818 & 87.39 (0.49) & 87.51 (0.64) & 87.32 (0.99) & 86.49 (0.28) \\
168350 & 91.54 (0.53) & 91.28 (0.81) & 93.29 (0.13) & 93.19 (0.07) \\
359983 & 87.45 (0.06) & 87.49 (0.05) & 87.46 (0.04) & 86.29 (0.05) \\
359990 & 94.63 (0.05) & 94.59 (0.14) & 95.53 (0.18) & 95.61 (0.01) \\
\bottomrule
\end{tabular}
\end{table}

\paragraph{How TabPFN was used.}
Among candidates that used TabPFN, soft-voting ensembles pairing it with a tree or boosting model were the most common architecture in the DeepSeek experiment (38.9\% of TabPFN users) and voting ensembles of other compositions in the GPT-5.4-mini experiment (31.3\%); TabPFN as the sole estimator accounted for 28.1\% and 24.3\%. The most frequent companions were random forests and gradient boosting in the DeepSeek experiment and logistic regression and extra trees in the GPT-5.4-mini experiment. Among the selected pipelines, 20 of 25 in the DeepSeek experiment and 19 of 25 in the GPT-5.4-mini experiment used TabPFN, of which 31 of 39 embedded it in an ensemble. On task 359990, whose size exceeds the input limit of TabPFN~v2.5, generated pipelines subsampled the training data for TabPFN and combined its predictions with a model fitted on the full data, a construction the generator adopted without being told about the limit.

\paragraph{Performance effect.}
The effect of announcing TabPFN was concentrated on the tasks where the standalone foundation model was already stronger than \sysname{} in the default environment. On task 168350, the explicit condition improved accuracy by \(1.6\) points under DeepSeek and \(1.8\) points under GPT-5.4-mini. The remaining tasks were unchanged within run-to-run variation, with the largest observed loss equal to \(0.07\) points. The generated pipelines matched or exceeded standalone TabPFN on eight of the nine task--version combinations where it returned a valid score. On task 359990, which exceeds the input limit of TabPFN~v2.5, the generator subsampled the data for TabPFN and combined it with a model trained on the full dataset, reaching \(94.7\%\) accuracy where standalone TabPFN returned no score.

These five tasks do not support significance claims, and \sysname{} evaluates 100 candidates whereas standalone TabPFN is fitted once. Given the five-task scope and unequal evaluation budgets, we interpret this ablation only as evidence about how the available components shape the generated search space: a component present only in the environment was never used, while one announced in the prompt was incorporated where it was useful.

\FloatBarrier
\section{Per-Task Wall-Clock Time}
\label{sec:runtime-appendix}
Table~\ref{tab:runtime-appendix} reports the mean total wall-clock time per repetition for a complete \sysname{} run of 100 candidates, averaged over repetitions and backbones on a shared heterogeneous cluster. These are end-to-end run times, dominated by candidate fitting and evaluation, and they are not hardware-normalized; the per-candidate caps of one hour (light tasks) and four hours (heavy tasks) bound individual candidate evaluations, not the per-run totals reported here. Times range from roughly \(0.7\) hours on the smallest tasks to over \(240\) hours on the largest, with the longest-running tasks concentrated in the heavy tier. Because the cluster is heterogeneous and prior program-evolution studies use different hardware and protocols, we report these values to document the cost of the campaign, not to compare
runtime across methods or hardware platforms.

\par\medskip
\begin{strip}
\centering
\small
\setlength{\tabcolsep}{5pt}
\renewcommand{\arraystretch}{0.98}

\captionof{table}{Mean wall-clock time per repetition (hours) for a full
100-candidate \sysname{} run, by OpenML task. Heavy-tier tasks
(four-hour per-candidate cap) are marked $\dagger$. Entries are ordered
by increasing runtime down each task--time block.}
\label{tab:runtime-appendix}

\begin{tabular}{@{}lr@{\hspace{1.5em}}
                    lr@{\hspace{1.5em}}
                    lr@{\hspace{1.5em}}
                    lr@{}}
\toprule
Task & Time (h) &
Task & Time (h) &
Task & Time (h) &
Task & Time (h) \\
\midrule
168757           & 0.72 & 190411           & 1.71 & 190410           & 6.62  & 3945$^\dagger$   & 24.94 \\
359955           & 0.72 & 359968           & 1.91 & 359981           & 6.64  & 359994$^\dagger$ & 28.47 \\
146820           & 0.78 & 359971           & 2.14 & 359957           & 6.97  & 360113$^\dagger$ & 31.11 \\
359959           & 0.79 & 359965           & 2.27 & 189922           & 7.70  & 168910           & 34.26 \\
146818           & 0.87 & 359975           & 2.56 & 359980           & 7.73  & 359967$^\dagger$ & 37.13 \\
190146           & 0.93 & 359983           & 2.69 & 359991           & 8.01  & 189356$^\dagger$ & 40.28 \\
359956           & 0.96 & 168911           & 3.19 & 359953$^\dagger$ & 9.37  & 168868$^\dagger$ & 40.69 \\
2073             & 1.01 & 359982           & 3.22 & 190412$^\dagger$ & 9.52  & 359984           & 46.61 \\
359960           & 1.06 & 359961           & 3.28 & 359992           & 10.79 & 360114$^\dagger$ & 54.02 \\
359962           & 1.10 & 359964           & 3.45 & 211986           & 12.20 & 168909$^\dagger$ & 85.54 \\
359954           & 1.13 & 168784           & 3.66 & 359977           & 12.94 & 359985$^\dagger$ & 99.44 \\
190137           & 1.23 & 359993           & 4.85 & 359990           & 14.40 & 7593$^\dagger$   & 118.18 \\
359963           & 1.35 & 359979           & 4.96 & 167120           & 15.30 & 10090$^\dagger$  & 118.43 \\
359972           & 1.37 & 190392           & 5.40 & 359966$^\dagger$ & 19.82 & 359989$^\dagger$ & 132.79 \\
359958           & 1.42 & 359970           & 5.62 & 211979           & 20.43 & 359976$^\dagger$ & 149.62 \\
168350           & 1.53 & 359969           & 5.87 & 189354$^\dagger$ & 20.75 & 360112$^\dagger$ & 228.02 \\
359974           & 1.60 & 359987           & 6.41 & 359973$^\dagger$ & 24.52 & 189355$^\dagger$ & 241.39 \\
\bottomrule
\end{tabular}
\end{strip}
\medskip

\FloatBarrier
\section{Per-Task Performance}
\label{sec:per-task}

\begin{figure*}[p]
\centering
\includegraphics[width=\textwidth]
{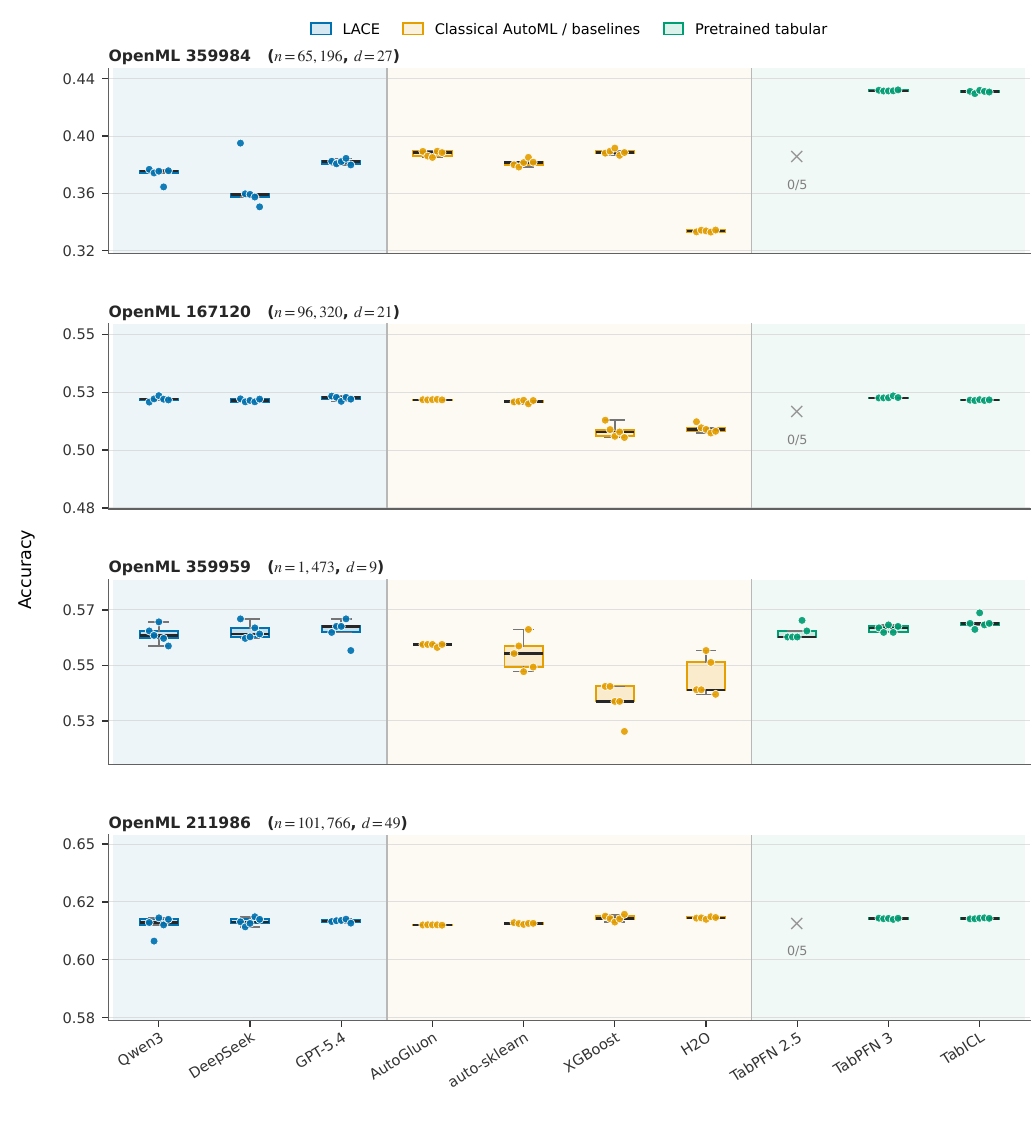}
\caption{Official-test accuracy distributions for OpenML tasks
359984, 167120, 359959, and 211986.}
\label{fig:per-task-01}
\end{figure*}

\begin{figure*}[p]
\centering
\includegraphics[width=\textwidth]
{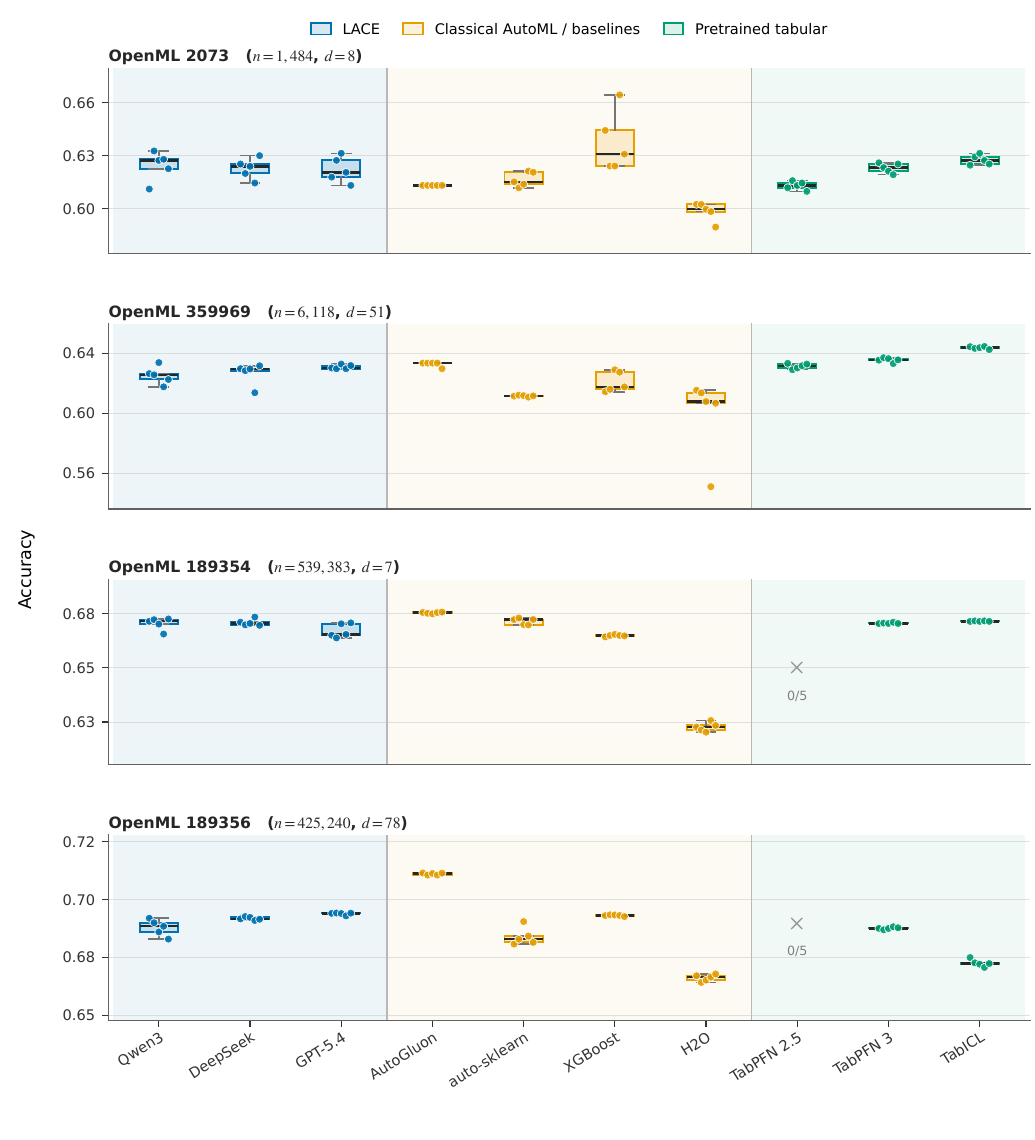}
\caption{Official-test accuracy distributions for OpenML tasks
2073, 359969, 189354, and 189356.}
\label{fig:per-task-02}
\end{figure*}

\begin{figure*}[p]
\centering
\includegraphics[width=\textwidth]
{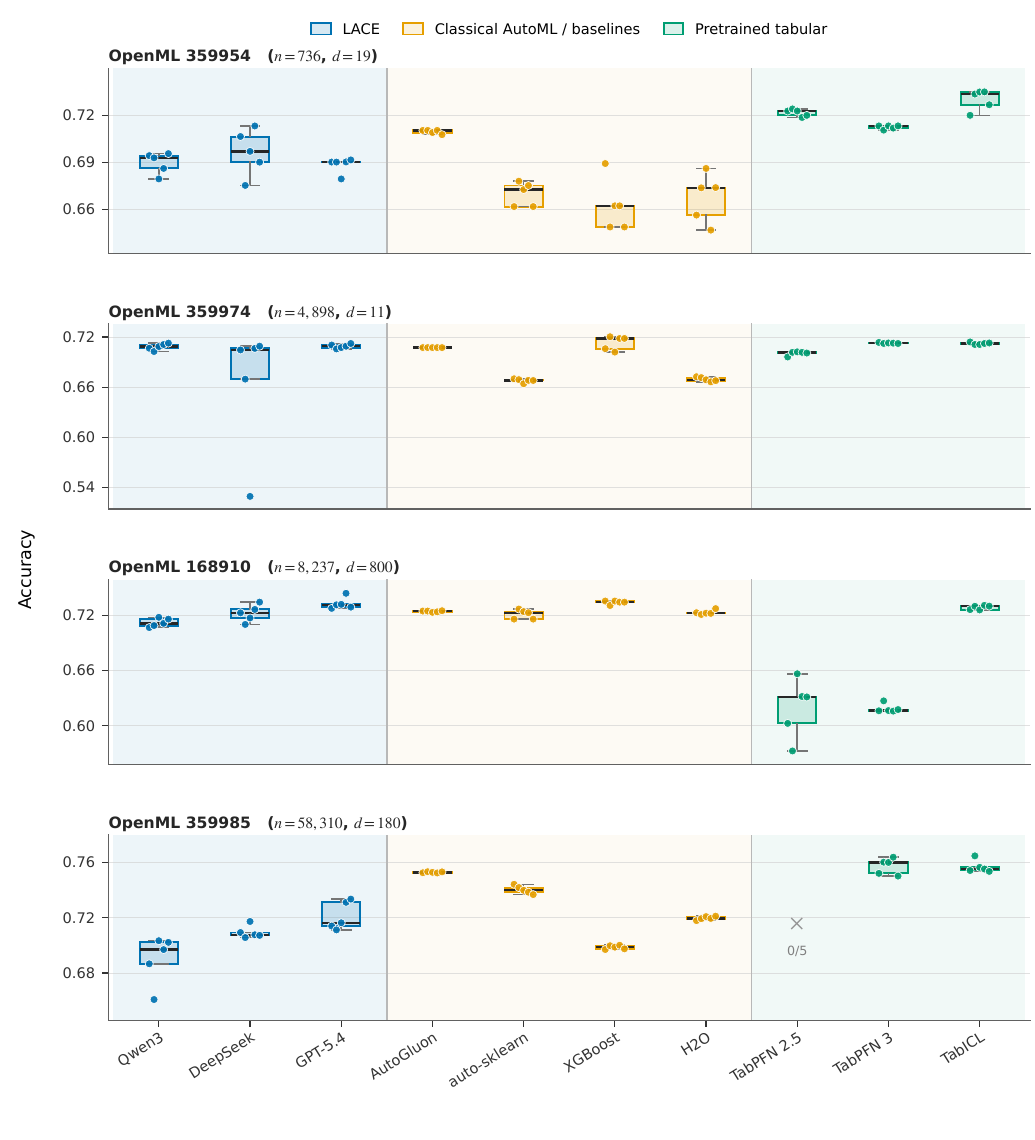}
\caption{Official-test accuracy distributions for OpenML tasks
359954, 359974, 168910, and 359985.}
\label{fig:per-task-03}
\end{figure*}

\begin{figure*}[p]
\centering
\includegraphics[width=\textwidth]
{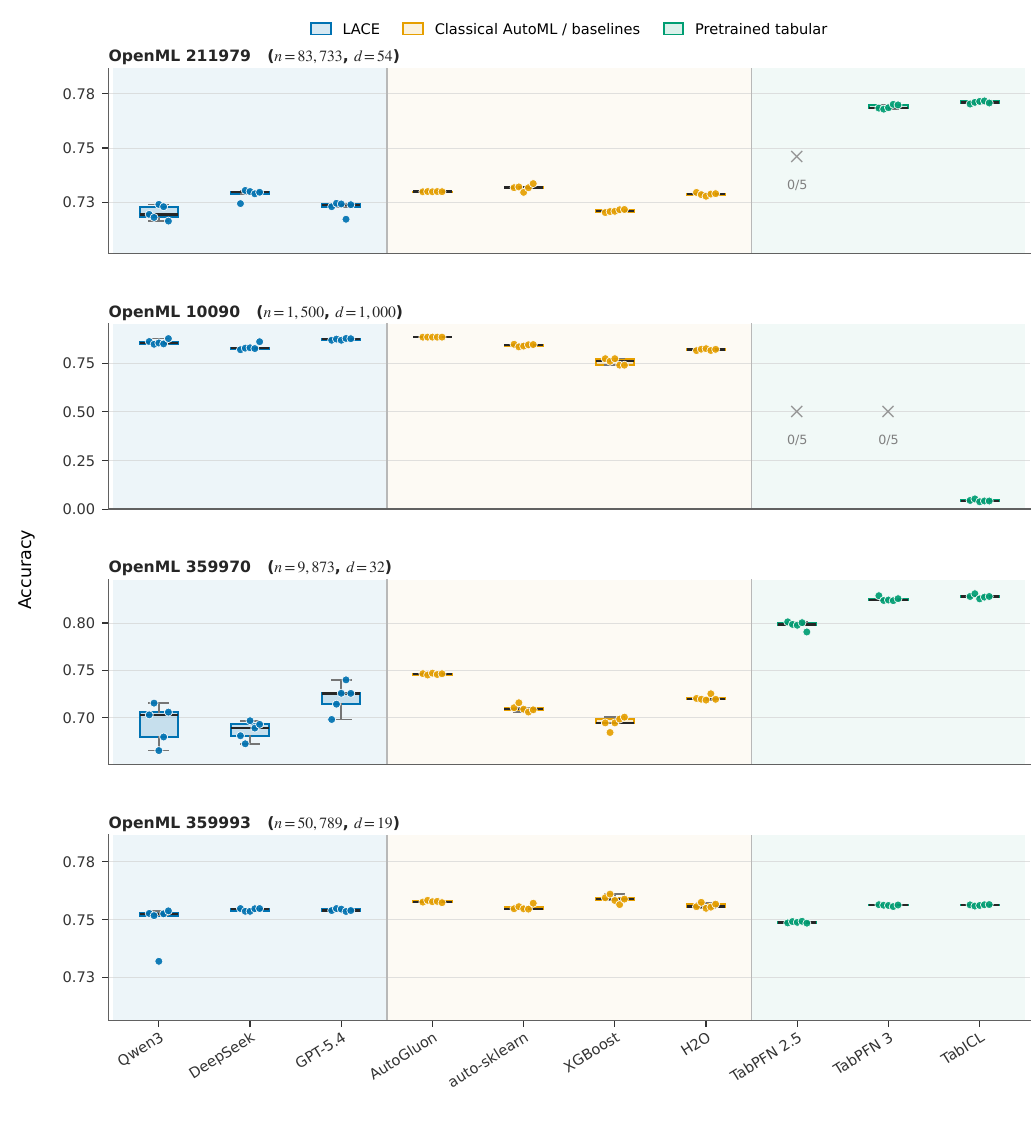}
\caption{Official-test accuracy distributions for OpenML tasks
211979, 10090, 359970, and 359993.}
\label{fig:per-task-04}
\end{figure*}

\begin{figure*}[p]
\centering
\includegraphics[width=\textwidth]
{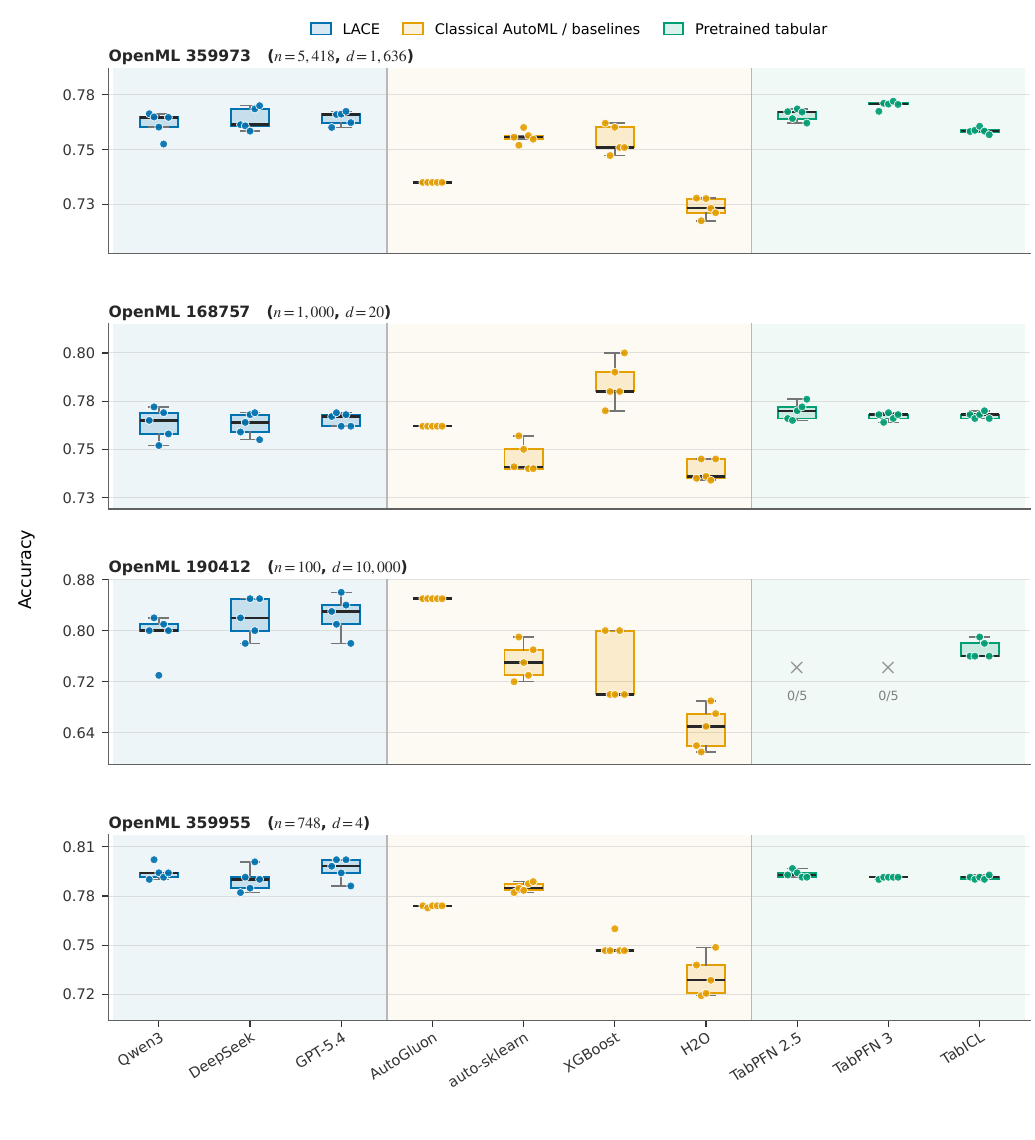}
\caption{Official-test accuracy distributions for OpenML tasks
359973, 168757, 190412, and 359955.}
\label{fig:per-task-05}
\end{figure*}

\begin{figure*}[p]
\centering
\includegraphics[width=\textwidth]
{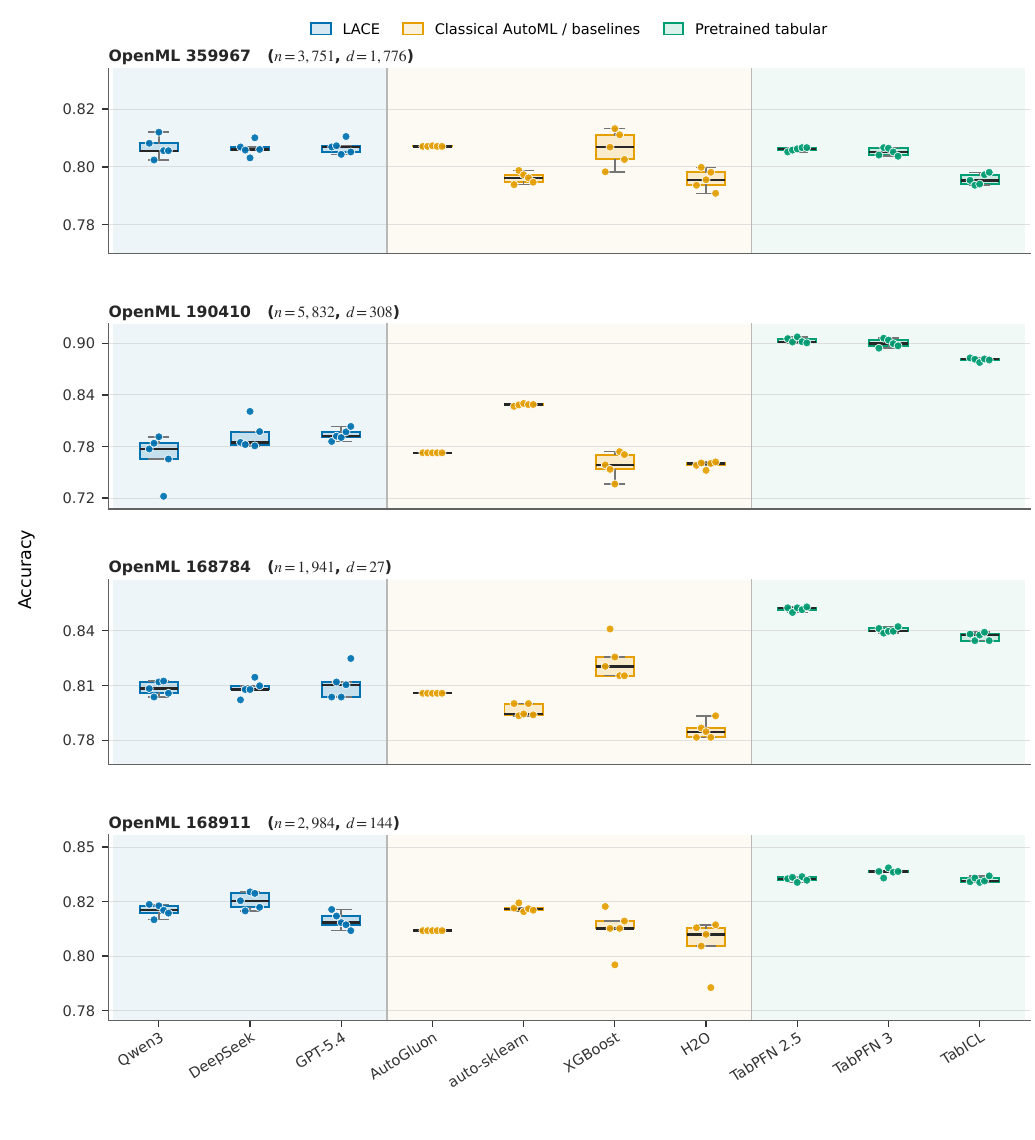}
\caption{Official-test accuracy distributions for OpenML tasks
359967, 190410, 168784, and 168911.}
\label{fig:per-task-06}
\end{figure*}

\begin{figure*}[p]
\centering
\includegraphics[width=\textwidth]
{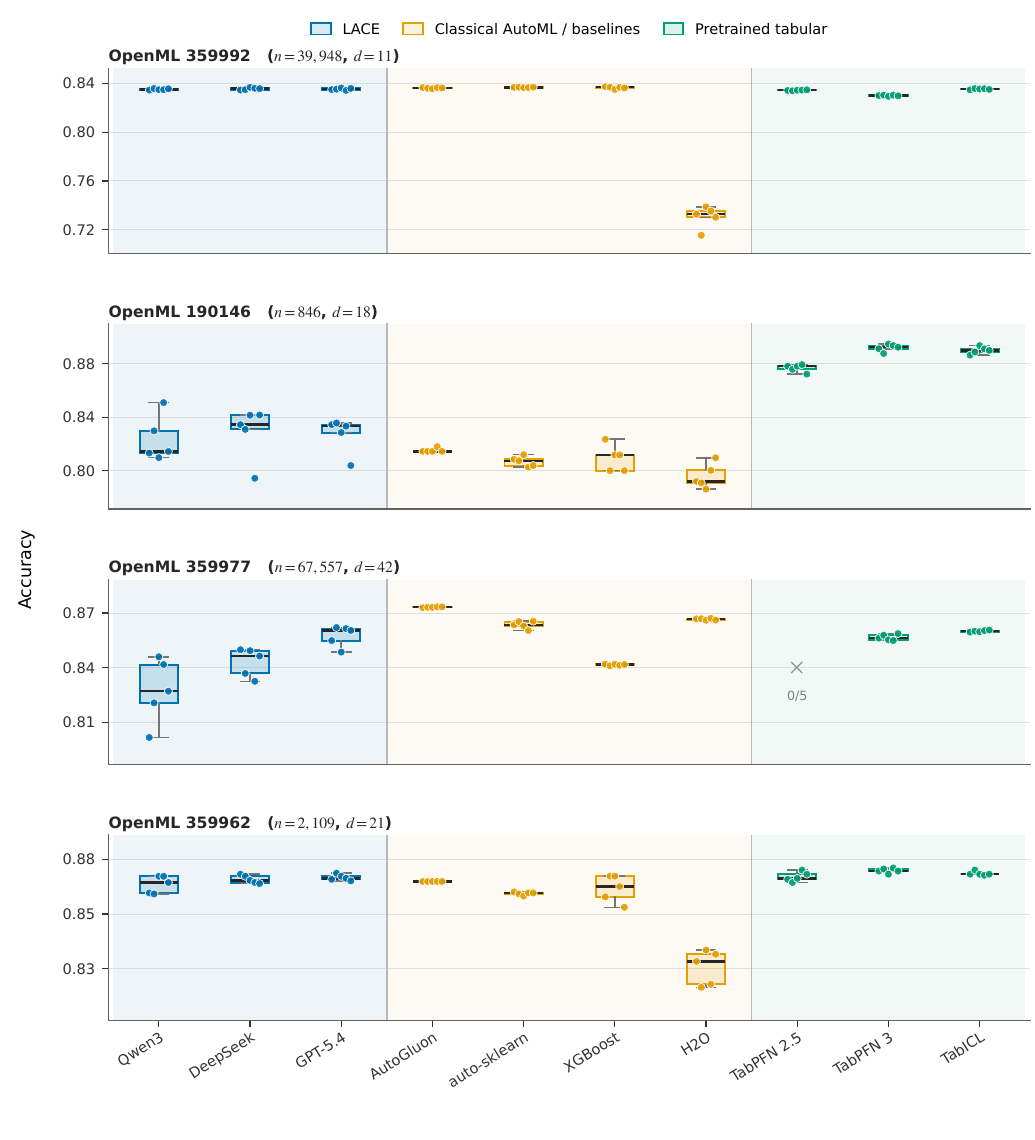}
\caption{Official-test accuracy distributions for OpenML tasks
359992, 190146, 359977, and 359962.}
\label{fig:per-task-07}
\end{figure*}

\begin{figure*}[p]
\centering
\includegraphics[width=\textwidth]
{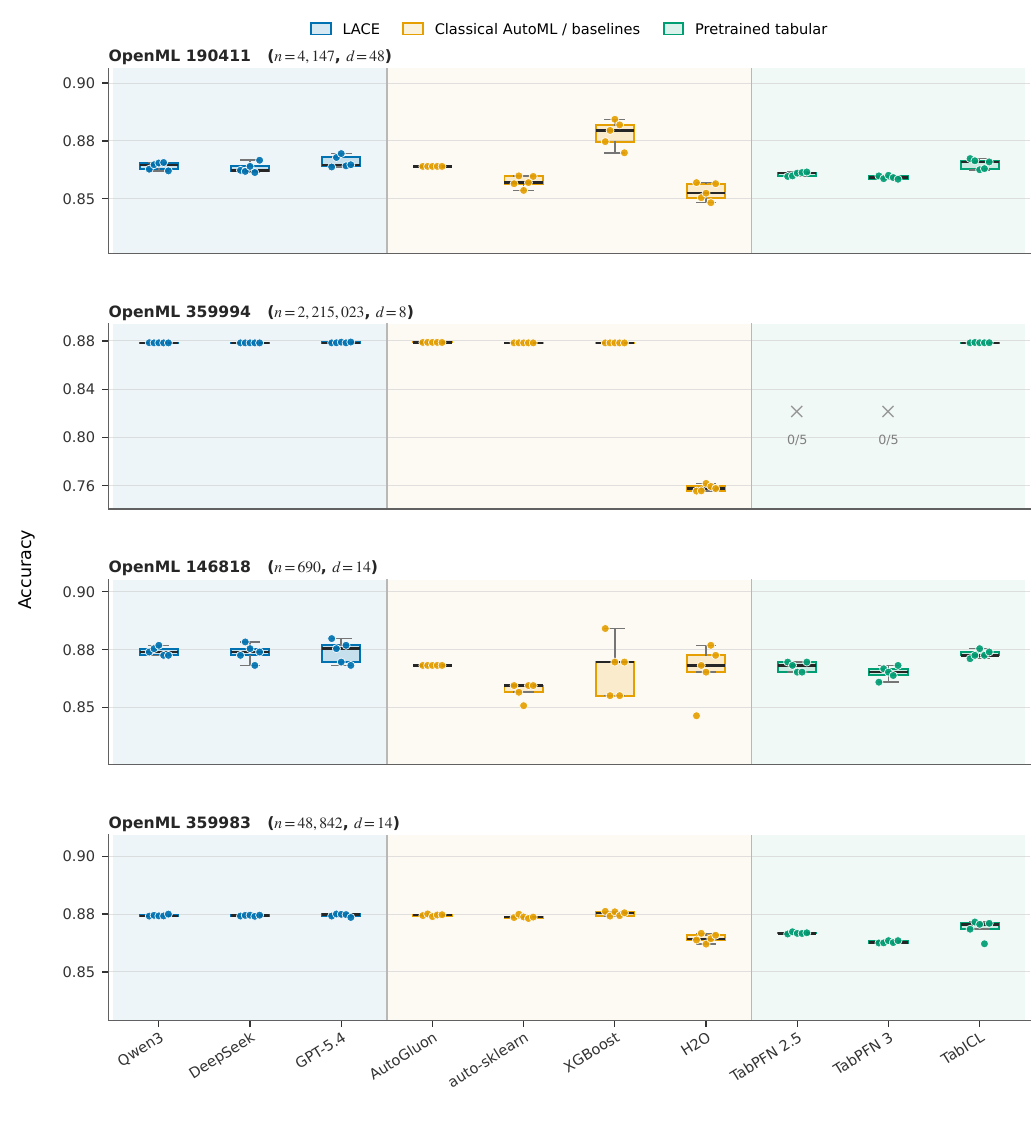}
\caption{Official-test accuracy distributions for OpenML tasks
190411, 359994, 146818, and 359983.}
\label{fig:per-task-08}
\end{figure*}

\begin{figure*}[p]
\centering
\includegraphics[width=\textwidth]
{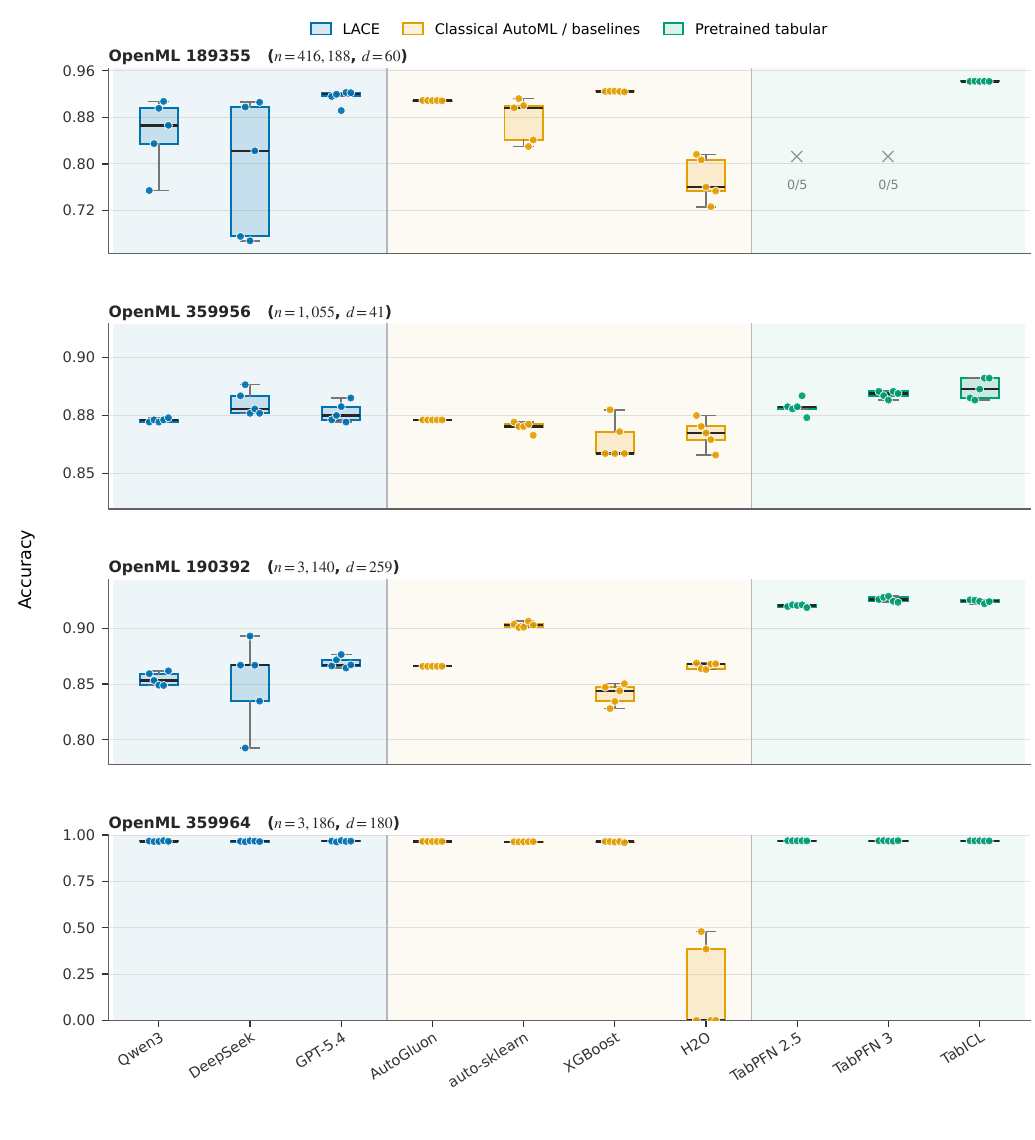}
\caption{Official-test accuracy distributions for OpenML tasks
189355, 359956, 190392, and 359964.}
\label{fig:per-task-09}
\end{figure*}

\begin{figure*}[p]
\centering
\includegraphics[width=\textwidth]
{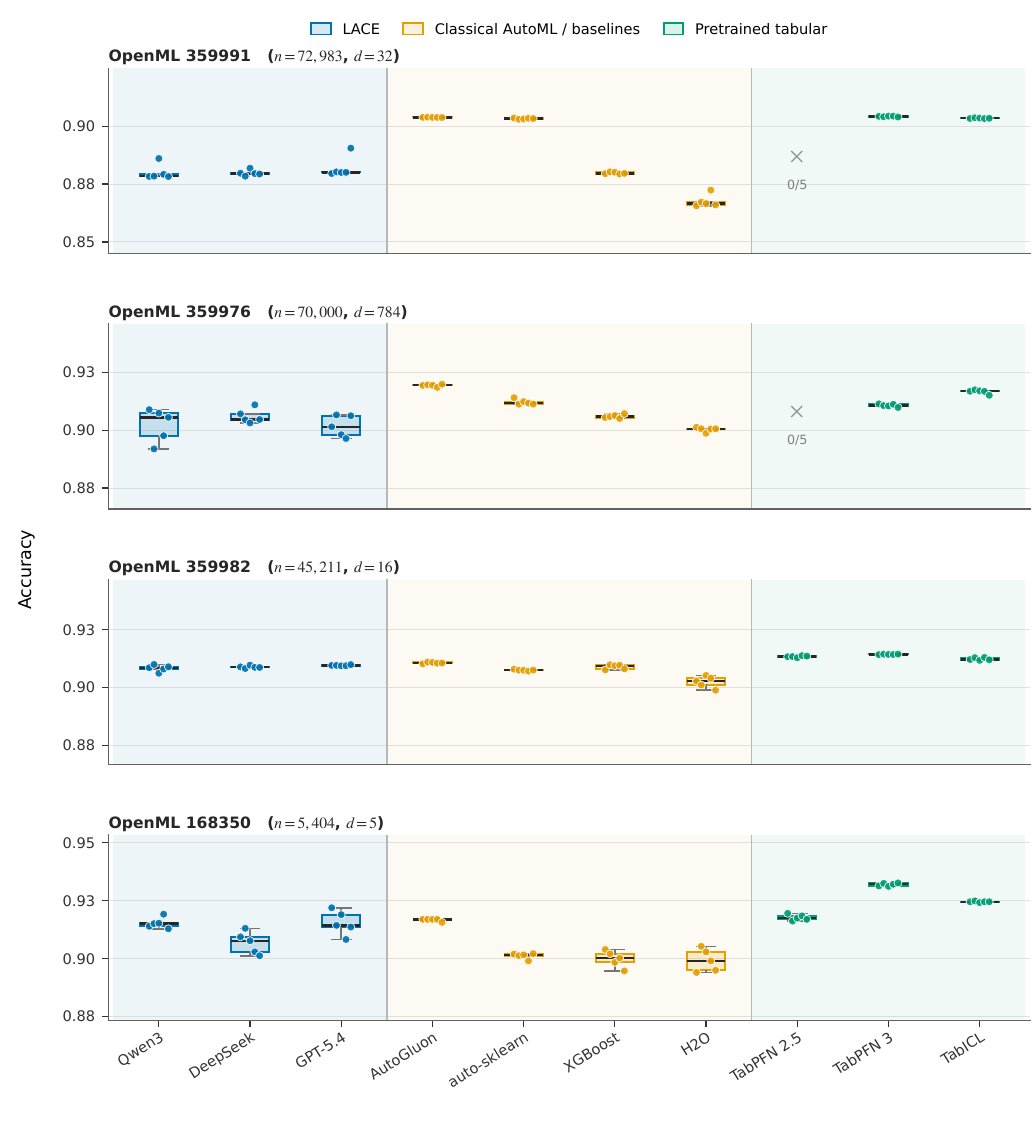}
\caption{Official-test accuracy distributions for OpenML tasks
359991, 359976, 359982, and 168350.}
\label{fig:per-task-10}
\end{figure*}

\begin{figure*}[p]
\centering
\includegraphics[width=\textwidth]
{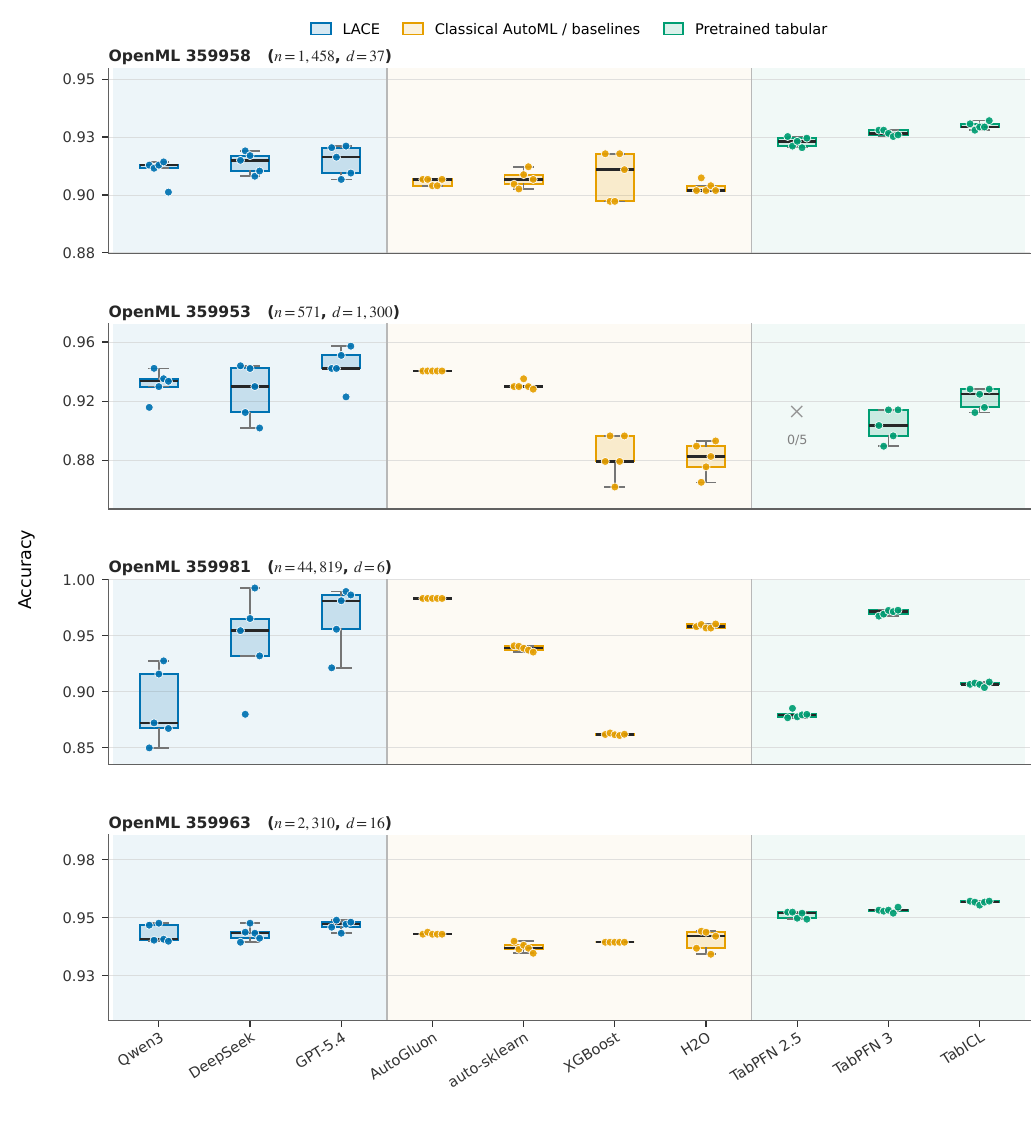}
\caption{Official-test accuracy distributions for OpenML tasks
359958, 359953, 359981, and 359963.}
\label{fig:per-task-11}
\end{figure*}

\begin{figure*}[p]
\centering
\includegraphics[width=\textwidth]
{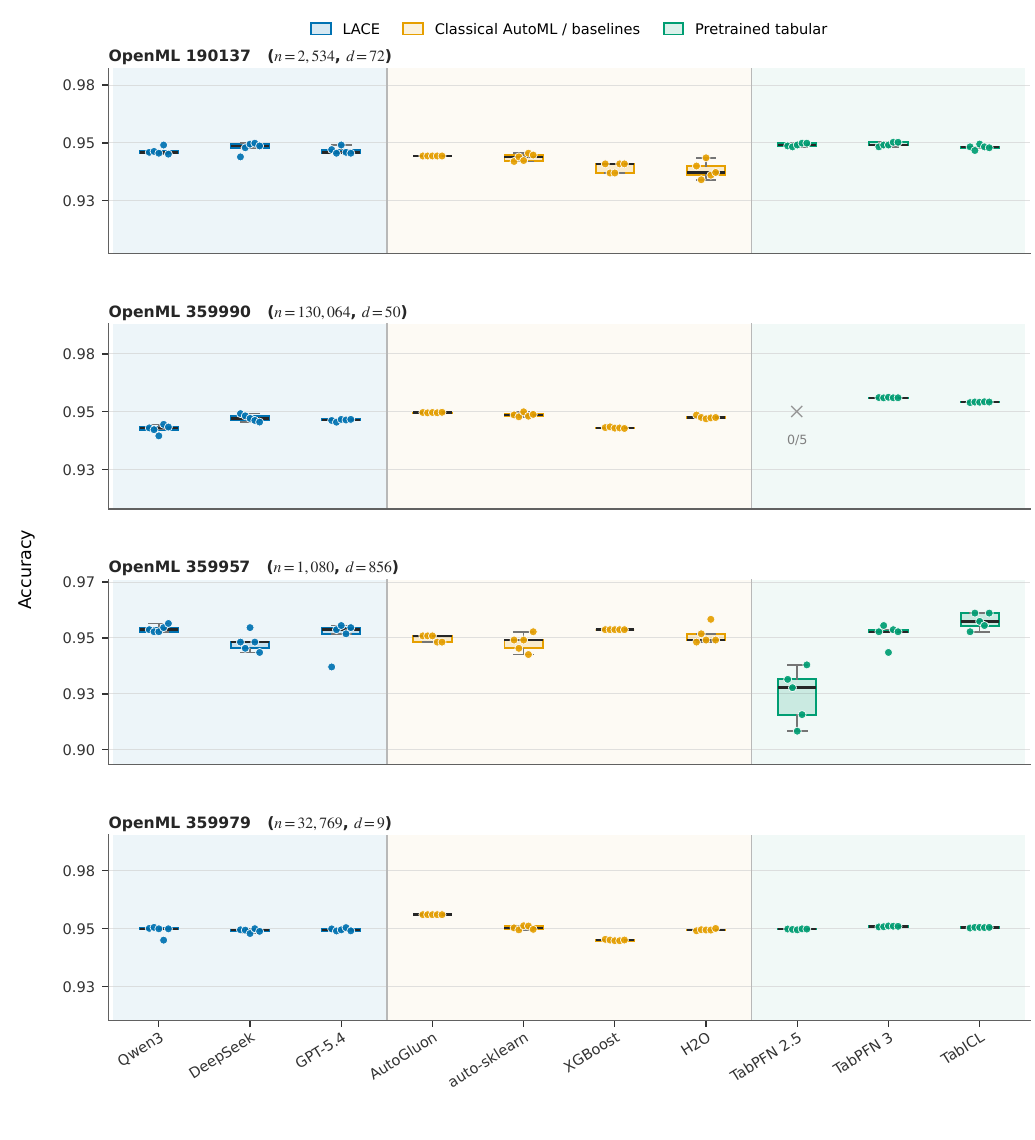}
\caption{Official-test accuracy distributions for OpenML tasks
190137, 359990, 359957, and 359979.}
\label{fig:per-task-12}
\end{figure*}

\begin{figure*}[p]
\centering
\includegraphics[width=\textwidth]
{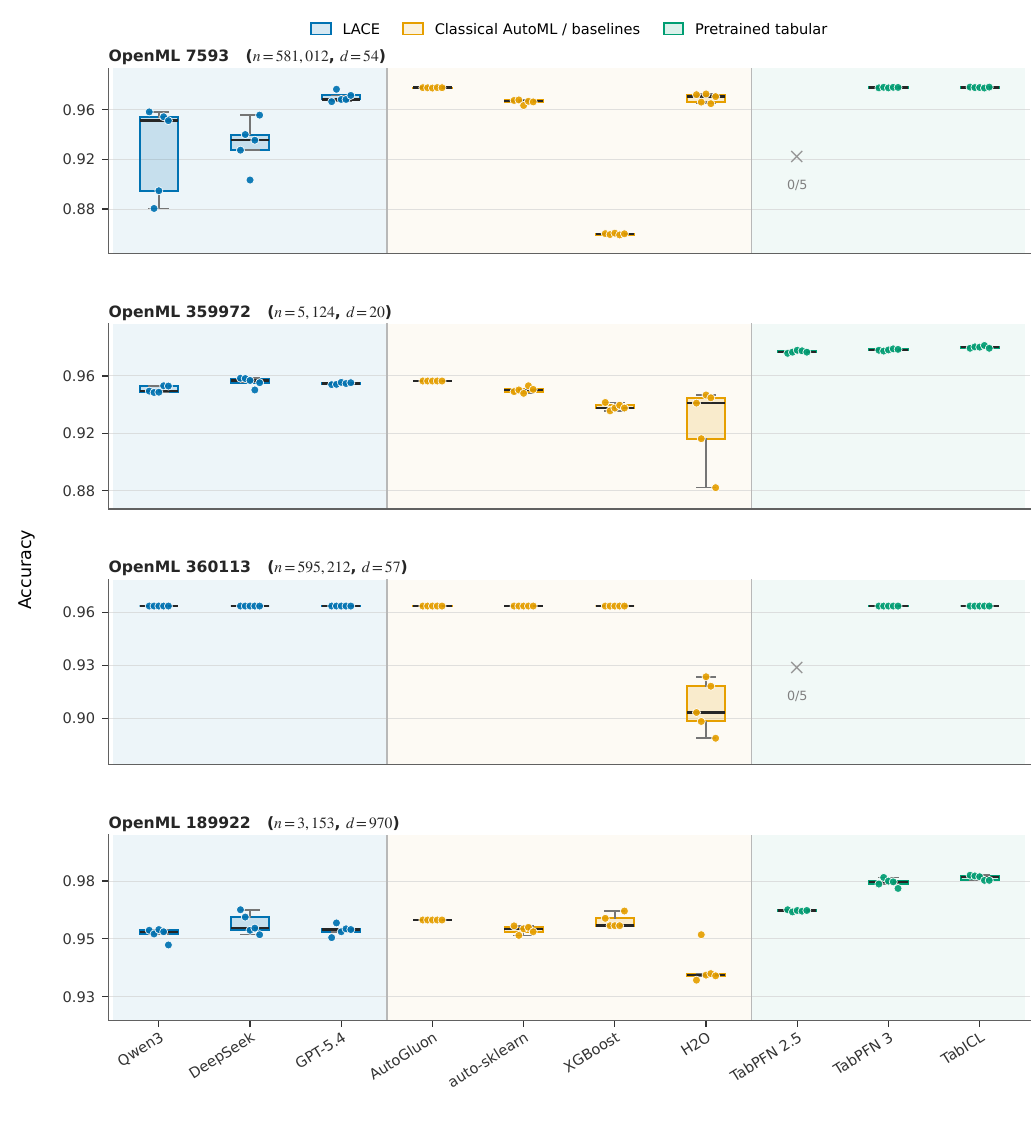}
\caption{Official-test accuracy distributions for OpenML tasks
7593, 359972, 360113, and 189922.}
\label{fig:per-task-13}
\end{figure*}

\begin{figure*}[p]
\centering
\includegraphics[width=\textwidth]
{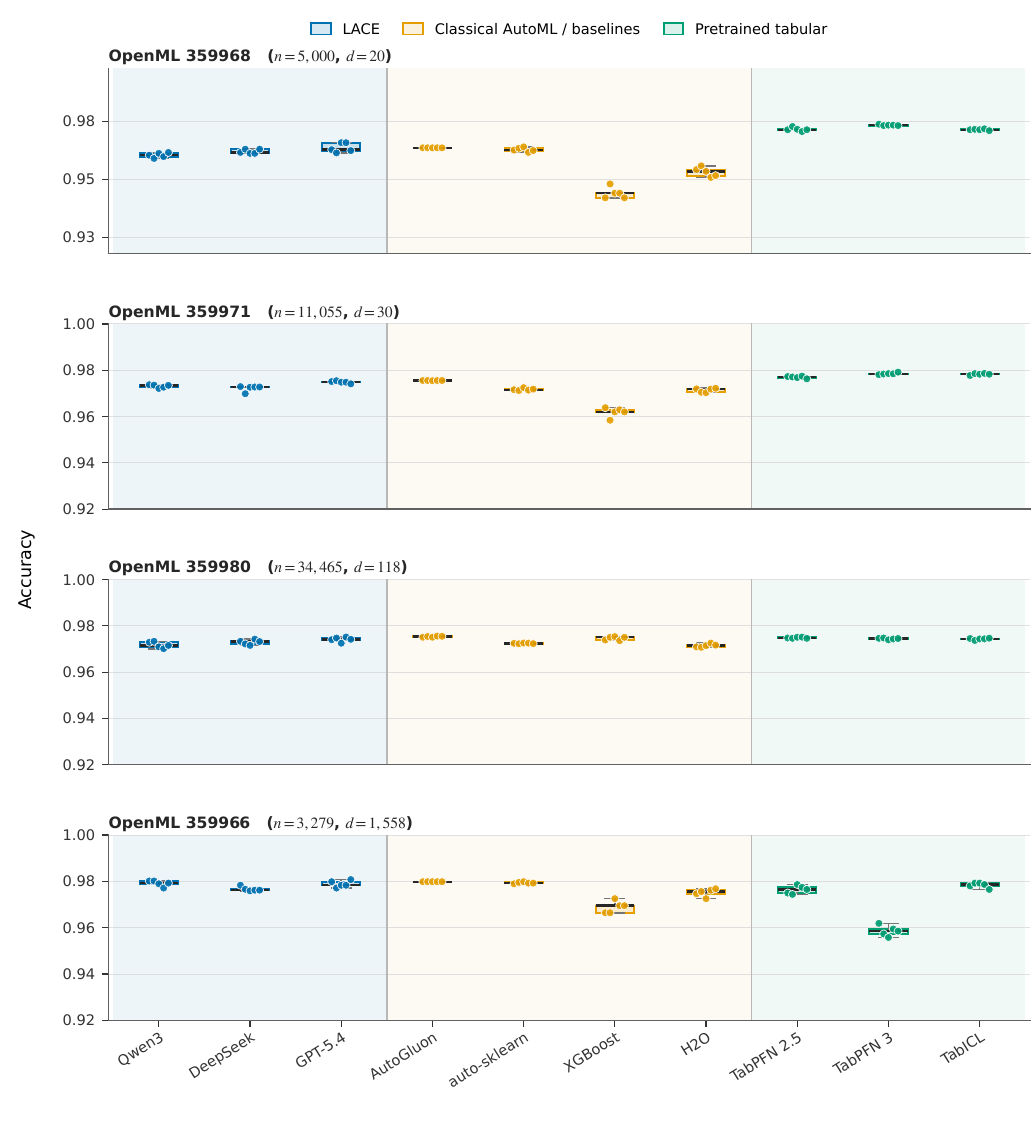}
\caption{Official-test accuracy distributions for OpenML tasks
359968, 359971, 359980, and 359966.}
\label{fig:per-task-14}
\end{figure*}

\begin{figure*}[p]
\centering
\includegraphics[width=\textwidth]
{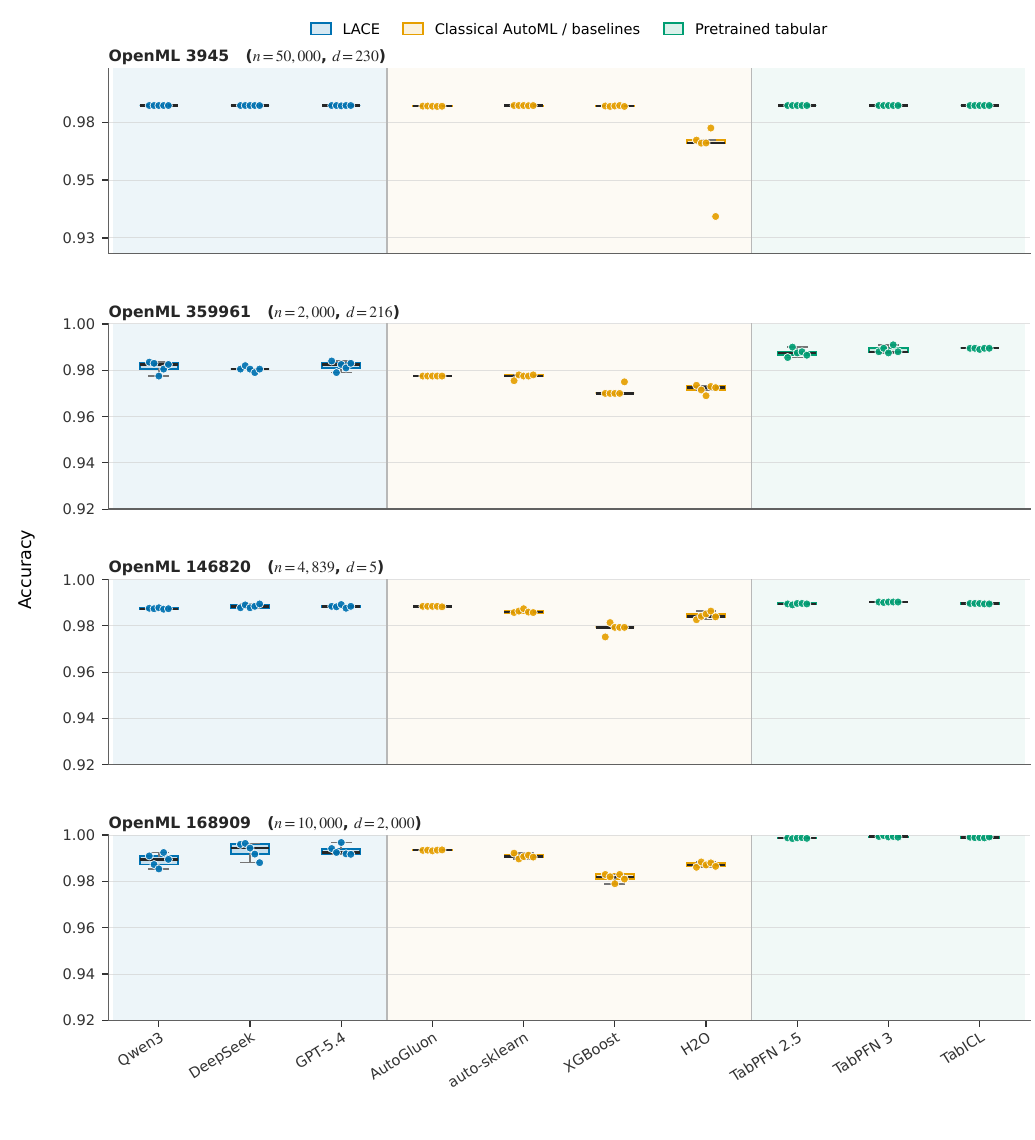}
\caption{Official-test accuracy distributions for OpenML tasks
3945, 359961, 146820, and 168909.}
\label{fig:per-task-15}
\end{figure*}

\begin{figure*}[p]
\centering
\includegraphics[width=\textwidth]
{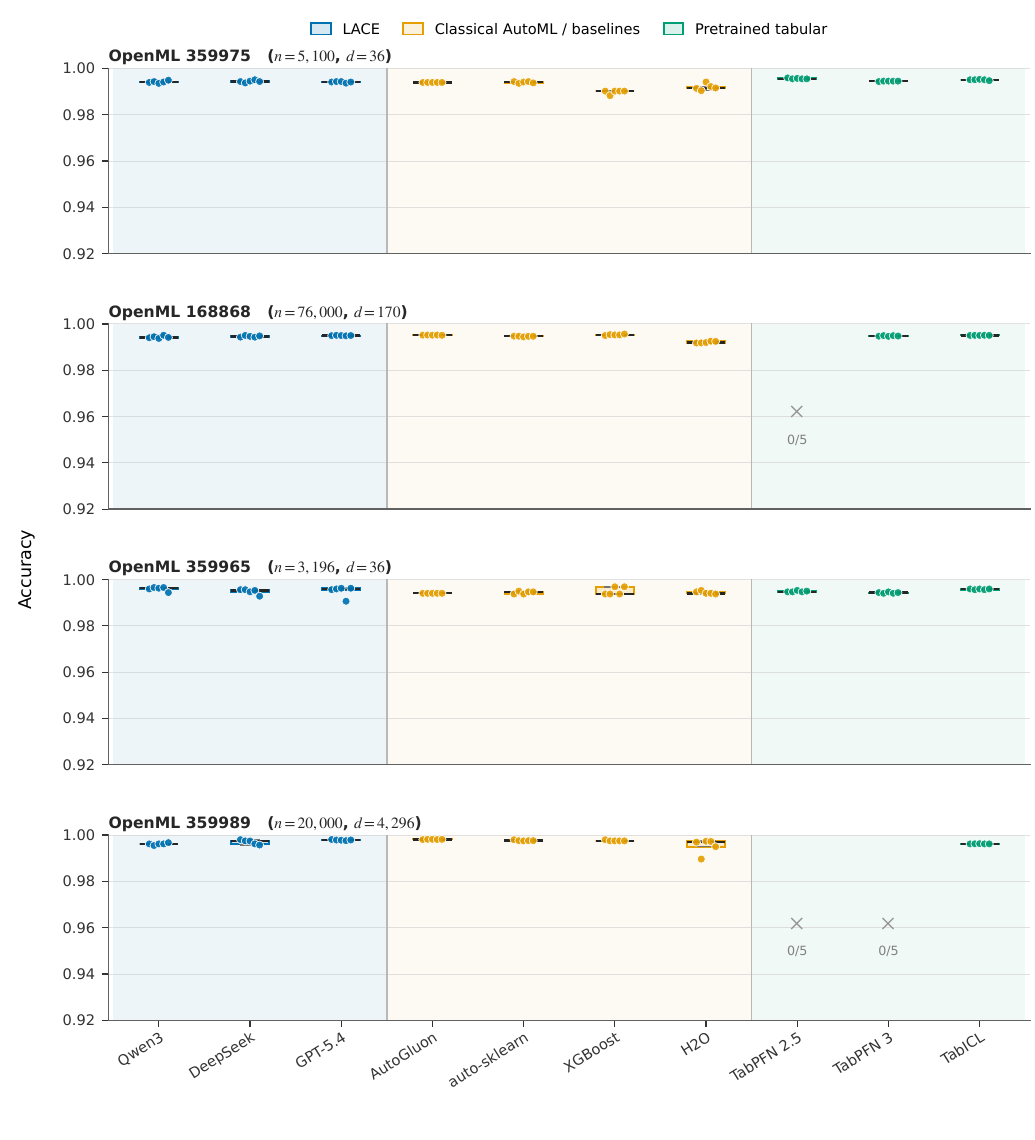}
\caption{Official-test accuracy distributions for OpenML tasks
359975, 168868, 359965, and 359989.}
\label{fig:per-task-16}
\end{figure*}

\begin{figure*}[p]
\centering
\includegraphics[width=\textwidth]
{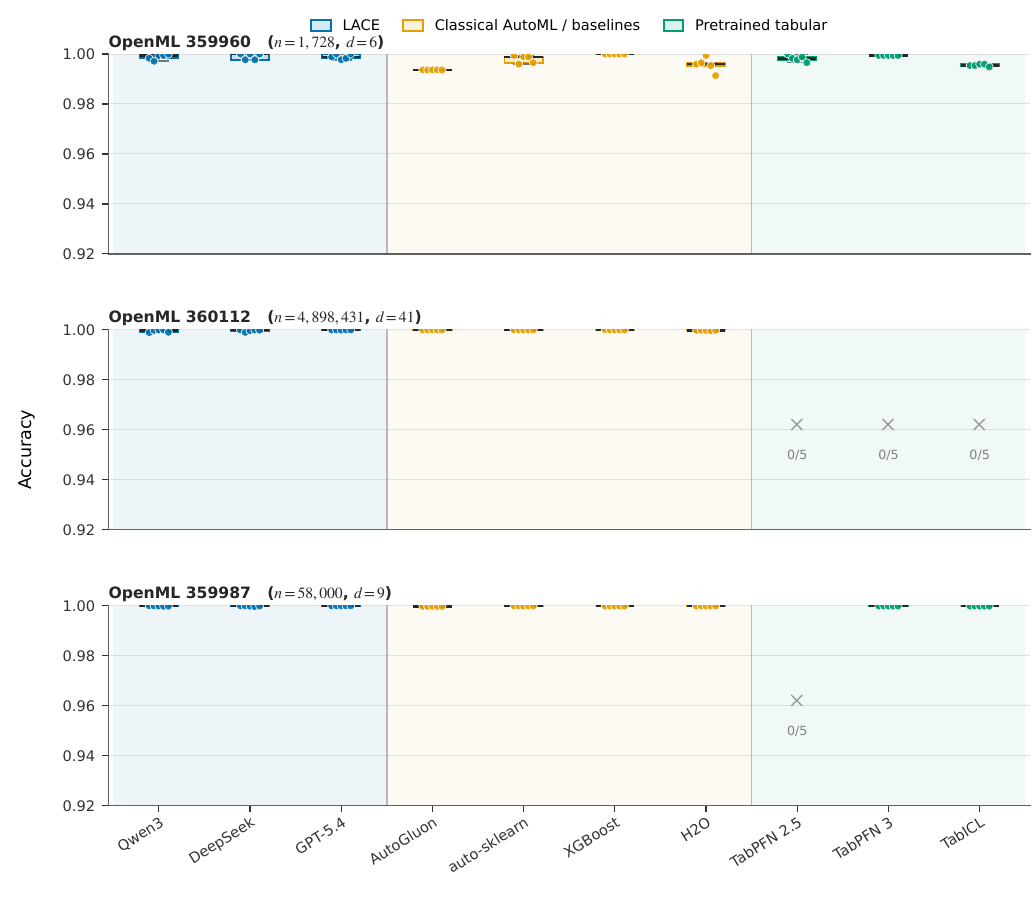}
\caption{Official-test accuracy distributions for OpenML tasks
359960, 360112, and 359987.}
\label{fig:per-task-17}
\end{figure*}

\begin{figure*}[p]
\centering
\includegraphics[width=\textwidth]
{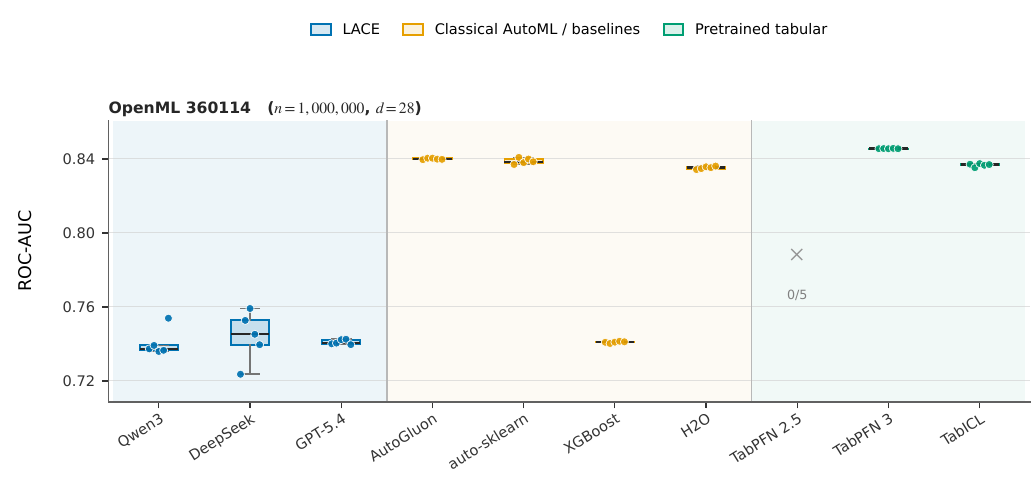}
\caption{Official-test ROC-AUC distributions for OpenML task 360114.}
\label{fig:per-task-18}
\end{figure*}

Figures~\ref{fig:per-task-01}--\ref{fig:per-task-18} show the full
per-task, per-method distributions of official-test scores underlying the aggregate results in the main paper. Accuracy tasks are ordered by the mean performance across eligible methods, from lower to higher average score. Most figures contain four tasks; the final accuracy figure contains three tasks, and the final figure contains the single ROC-AUC task.

Within each panel, methods are grouped into LACE variants, classical AutoML and baseline methods, and pretrained tabular methods. A boxplot is shown only when at least three of the five expected repetitions returned a valid score, following the validity rule used throughout the paper. Boxes show the interquartile range and median, while points show the individual repetitions. Where three or four valid repetitions are available, the corresponding coverage is indicated. A grey \(\times\) denotes fewer than three valid repetitions, for which no distribution is reported. Note that each task uses its own vertical scale to make within-task differences visible.

% ============================================================================
% Accuracy figures
% ============================================================================

% ============================================================================
% ROC-AUC figure
% ============================================================================

\clearpage
\end{document}

%% file: Figures/method_figure/lace_overview.tex
\begin{tikzpicture}[
  x=1cm, y=1cm,
  base/.style={draw=black!60, fill=white, line width=0.5pt, inner sep=3pt,
               align=left, rounded corners=1pt, font=\scriptsize},
  proc/.style={base, fill=blue!12},
  data/.style={base, align=center},
  code/.style={base, fill=orange!15},
  stepbadge/.style={circle, draw=none, fill=black!75, text=white, inner sep=0pt, minimum size=3.4mm, font=\tiny\bfseries},
  lbl/.style={draw=none, fill=none, inner sep=1pt, font=\tiny},
  bandlbl/.style={draw=none, inner xsep=4pt, inner ysep=2pt,
                rounded corners=1pt,
                font=\scriptsize\itshape, text=black!65},
  ar/.style={draw=black!70, line width=0.5pt, -{Stealth[length=4pt]}},
  arl/.style={draw=black!70, line width=0.5pt},
  ard/.style={draw=black!45, line width=0.5pt, dashed, -{Stealth[length=4pt]}},
  ardl/.style={draw=black!45, line width=0.5pt, dashed}
]
\path[use as bounding box] (0,-0.3) rectangle (17.6,6.6);

% ---- bands -------------------------------------------------------------
\fill[blue!4]  (0,3.35) rectangle (17.6,6.6);
\fill[black!6] (0,-0.3) rectangle (17.6,3.25);
\draw[dashed, black!55] (0,3.3) -- (17.6,3.3);
\node[bandlbl, fill=blue!10] at (8.8,3.55)
  {\textbf{Generator layer:} coarse summary only};
\node[bandlbl, fill=black!10] at (8.8,3.05)
  {\textbf{Evaluator layer:} full task access};

% ---- top band ----------------------------------------------------------
\node[base, text width=2.6cm] (N1) at (1.6,5.0)
 {\textbf{Task summary}\\
  binary classification\\
  about \(5{,}000\) samples\\
  mostly numeric\\
  some categorical\\[2pt]
  {\tiny\itshape\color{black!60} no names or raw values}};

\node[proc, text width=3.0cm] (N2) at (5.3,5.0)
 {\textbf{Prompt}\\
  {\tiny\bfseries Initial generation}\\[-1pt]
  task summary + class format\\[2pt]
  {\tiny\bfseries Later generations}\\[-1pt]
  parent code + feedback\\
  population summary\\
  mutation instruction};

\node[proc, text width=1.4cm, align=center] (N3) at (8.5,5.0)
 {\textbf{LLM}\\ variation\\ operator};

\node[code, text width=3.8cm] (N4) at (11.7,5.0)
 {{\tiny candidate program $p$}\\[1pt]
  {\tiny\ttfamily class Pipeline:\\
   \hspace*{0.7em}\_\_init\_\_(self, X, y)\hfill\# fit once\\
   \hspace*{0.7em}\_\_call\_\_(self, X)\hfill\# predict}};

\node[proc, text width=2.4cm] (N5) at (15.5,5.0)
 {\textbf{Population}\\
  $\mu=4$ parents\\
  $\lambda=12$ offspring\\
  keep best $\mu$ of $\mu+\lambda$};

% ---- bottom band -------------------------------------------------------
\node[data, text width=2.0cm] (N6) at (1.6,2.55)
 {OpenML task\\ $\mathcal{T}=(\mathcal{D},\mathcal{S},m)$};
\node[data, text width=1.2cm] (N7) at (0.9,1.5)
 {$\mathcal{D}_{\mathrm{train}}$};
\node[data, text width=1.1cm] (N10) at (3.3,1.5)
 {$\mathcal{D}_{\mathrm{test}}$};
\node[data, text width=0.85cm] (N8) at (0.6,0.5)
 {$\mathcal{D}_{\mathrm{tr}}$};
\node[data, text width=0.9cm] (N9) at (1.95,0.5)
 {$\mathcal{D}_{\mathrm{val}}$};

\node[base, text width=3.8cm] (N11) at (11.7,1.2)
 {\textbf{Evaluator}\\
  isolated subprocess\\
  fit on $\mathcal{D}_{\mathrm{tr}}$, score on $\mathcal{D}_{\mathrm{val}}$\\
  failure $\rightarrow$ worst fitness + error};

\node[base, text width=2.6cm] (N12) at (15.5,1.2)
 {\textbf{Final output}\\
  refit $\hat{p}$ on $\mathcal{D}_{\mathrm{train}}$\\
  score once on $\mathcal{D}_{\mathrm{test}}$\\
  return code + score};
  
% ---- data hierarchy ----------------------------------------------------
\draw[arl] (N6.south) -- (1.6,1.95);
\draw[arl] (0.9,1.95) -- (3.3,1.95);
\draw[ar]  (0.9,1.95) -- (N7.north);
\draw[ar]  (3.3,1.95) -- (N10.north);
\draw[arl] (N7.south) -- (0.9,1.0);
\draw[arl] (0.6,1.0)  -- (1.95,1.0);
\draw[ar]  (0.6,1.0)  -- (N8.north);
\draw[ar]  (1.95,1.0) -- (N9.north);

% ---- data feed to evaluator -------------------------------------------
\draw[arl] (N8.south) -- (0.6,0.0);
\draw[arl] (N9.south) -- (1.95,0.0);
\draw[arl] (0.6,0.0) -- (8.0,0.0) -- (8.0,1.2);
\draw[ar]  (8.0,1.2) -- (N11.west);
\node[lbl] at (8.85,1.4)
  {$\mathcal{D}_{\mathrm{tr}}$, $\mathcal{D}_{\mathrm{val}}$};

\draw[ardl] (N10.east) -- (4.4,1.5) -- (4.4,2.5) -- (15.5,2.5);
\draw[ard]  (15.5,2.5) -- (N12.north);
\node[lbl, anchor=west, text=black!55] at (4.6,2.35)
  {held out; used once, at the end};

% main loop 
\draw[ar] (N6.north) -- (N1.south);
\node[lbl, anchor=east] at (3.1,3.6) {coarse summary};
\draw[ar] (N1.east) -- (N2.west);
\draw[ar] (N2.east) -- (N3.west);
\draw[ar] (N3.east) -- (N4.west);

\draw[ar] (N4.south) -- (N11.north);
\node[lbl, anchor=east] at (12.45,3.6) {execute};

\draw[arl] (N11.east) -- (13.95,1.2) -- (13.95,4.0) -- (15.5,4.0);
\draw[ar]  (15.5,4.0) -- (N5.south);
\node[lbl, anchor=east] at (15.08,3.6) {score $f(p)$};

\draw[ar] ([xshift=0.7cm]N5.south) -- (16.2,1.85);
\node[lbl, anchor=west] at (16.24,3.6) {best-so-far $\hat{p}$};

\draw[arl] (N5.north) -- (15.5,6.47) -- (5.3,6.47);
\draw[ar]  (5.3,6.47) -- (N2.north);
\node[lbl, fill=blue!4, inner sep=2pt, align=center] at (10.4,6.47)
  {repeat until budget $B$: selected parent, its feedback, population summary};

% step badges 
\node[stepbadge] at ([xshift=-2pt,yshift=2pt]N2.north west)  {1};
\node[stepbadge] at ([xshift=-2pt,yshift=2pt]N3.north west)  {2};
\node[stepbadge] at ([xshift=-2pt,yshift=2pt]N4.north west)  {3};
\node[stepbadge] at ([xshift=-2pt,yshift=2pt]N11.north west) {4};
\node[stepbadge] at ([xshift=-2pt,yshift=2pt]N5.north west)  {5};
\node[stepbadge] at ([xshift=2pt,yshift=2pt]N12.north east) {6};

\end{tikzpicture}